# A Unified Framework for Low-Rank plus Sparse Matrix Recovery

Xiao Zhang*† and Lingxiao Wang*‡ and Quanquan Gu§

## Abstract

We propose a unified framework to solve general low-rank plus sparse matrix recovery problems based on matrix factorization, which covers a broad family of objective functions satisfying the restricted strong convexity and smoothness conditions. Based on projected gradient descent and the double thresholding operator, our proposed generic algorithm is guaranteed to converge to the unknown low-rank and sparse matrices at a locally linear rate, while matching the best-known robustness guarantee (i.e., tolerance for sparsity). At the core of our theory is a novel structural Lipschitz gradient condition for low-rank plus sparse matrices, which is essential for proving the linear convergence rate of our algorithm, and we believe is of independent interest to prove fast rates for general superposition-structured models. We illustrate the application of our framework through two concrete examples: robust matrix sensing and robust PCA. Experiments on both synthetic and real datasets corroborate our theory.

## 1 Introduction

Low-rank matrix recovery has received considerable attention in machine learning and high-dimensional statistical inference in the past decades (Candès and Recht, 2009; Candès and Tao, 2010; Recht et al., 2010; Jain et al., 2010; Hsu et al., 2011; Negahban and Wainwright, 2011; Agarwal et al., 2012a; Negahban and Wainwright, 2012; Yang and Ravikumar, 2013; Chen et al., 2013; Jain and Netrapalli, 2014; Hardt and Wootters, 2014; Hardt et al., 2014; Chen and Wainwright, 2015; Gui and Gu, 2015; Bhojanapalli et al., 2016; Wang et al., 2016, 2017; Xu et al., 2017). One important question is whether low-rank matrix estimation algorithms are robust to arbitrarily sparse corruptions, which motivates the problem of low-rank plus sparse matrix recovery, such as robust matrix sensing (Waters et al., 2011; Kyrillidis and Cevher, 2012), robust PCA (Candès et al., 2011; Xu et al., 2010; Yi et al., 2016), robust covariance matrix estimation (Agarwal et al., 2012b) and robust multi-task regression (Negahban and Wainwright, 2011; Xu and Leng, 2012). Following this line of research, we consider the general problem of low-rank plus sparse matrix recovery, where the objective is to recover an unknown model parameter matrix that can be decomposed as the sum of a low-rank matrix $\mathbf{X}^* \in \mathbb{R}^{d_1 \times d_2}$ and a sparse matrix $\mathbf{S}^* \in \mathbb{R}^{d_1 \times d_2}$, from a set of $n$ observations generated from the model. More specifically, let $\mathcal{L}_n : \mathbb{R}^{d_1 \times d_2} \to \mathbb{R}$ be the sample loss function derived from some statistical model, which measures the goodness of fit to the observations with respect to any given low-rank matrix $\mathbf{X}$ and sparse matrix $\mathbf{S}$. Then the general low-rank plus sparse matrix recovery problem can be cast into the following nonconvex optimization problem

$$\min_{\mathbf{X},\mathbf{S}\in\mathbb{R}^{d_1 \times d_2}} \mathcal{L}_n(\mathbf{X} + \mathbf{S}), \quad \text{subject to } \mathbf{X} \in \mathcal{C}, \text{ rank}(\mathbf{X}) \leq r, \|\mathbf{S}\|_0 \leq s, \tag{1.1}$$

where $\mathcal{C}$ is a constraint set such that $\mathbf{X}^* \in \mathcal{C}$ (see Section 3 for more details), $r$ denotes the rank of $\mathbf{X}^*$, $\|\mathbf{S}\|_0$ denotes the number of nonzero entries in $\mathbf{S}$, and $s$ denotes the number of nonzero entries in $\mathbf{S}^*$.

---

*Equal Contribution

†Department of Computer Science, University of Virginia, Charlottesville, VA 22904, USA; e-mail:xz7bc@virginia.edu

‡Department of Computer Science, University of Virginia, Charlottesville, VA 22904, USA; e-mail: lw4wr@virginia.edu

§Department of Computer Science, University of Virginia, Charlottesville, VA 22904, USA; e-mail: qg5w@virginia.edu

A long line of research has been proposed to study how to recover the unknown decomposition via convex relaxation (Xu et al., 2010; Candès et al., 2011; Chandrasekaran et al., 2011; Hsu et al., 2011; Agarwal et al., 2012b; Chen et al., 2013; Yang and Ravikumar, 2013; Klopp et al., 2014). However, convex relaxation based algorithms usually involve a time-consuming singular value decomposition (SVD) step in each iteration, which is computationally very expensive for large scale high-dimensional data. In order to solve low-rank plus sparse matrix recovery problems more efficiently, recent studies (Kyrillidis and Cevher, 2012; Netrapalli et al., 2014; Chen and Wainwright, 2015; Gu et al., 2016; Yi et al., 2016) proposed to use nonconvex optimization algorithms such as alternating minimization and gradient descent. Although these nonconvex optimization based approaches improve the computational efficiency upon convex relaxation based methods, they still suffer from either unsatisfied robustness guarantee and/or limitations to specific models.

In this paper, we aim to develop a unified framework to recover both the low-rank and the sparse matrices from generic statistical models. Following Burer and Monteiro (2003), we reparameterize the low-rank matrix as the product of two small factor matrices, i.e., $\mathbf{X} = \mathbf{U}\mathbf{V}^\top$ where $\mathbf{U} \in \mathbb{R}^{d_1 \times r}$ and $\mathbf{V} \in \mathbb{R}^{d_2 \times r}$, and propose to solve the following nonconvex optimization problem

$$\min_{\substack{\mathbf{U}\in\mathcal{C}_1,\mathbf{V}\in\mathcal{C}_2 \\ \mathbf{S}\in\mathbb{R}^{d_1 \times d_2}}} \mathcal{L}_n(\mathbf{U}\mathbf{V}^\top + \mathbf{S}), \quad \text{subject to } \|\mathbf{S}\|_0 \leq s, \tag{1.2}$$

where $\mathcal{C}_1 \subseteq \mathbb{R}^{d_1 \times r}, \mathcal{C}_2 \subseteq \mathbb{R}^{d_2 \times r}$ are the corresponding rotation invariant constraint sets induced by $\mathcal{C}$ (see Section 3 for more details). Due to Burer-Monteiro factorization (Burer and Monteiro, 2003), i.e., the reformulation $\mathbf{X} = \mathbf{U}\mathbf{V}^\top$, the rank constraint is automatically satisfied in (1.2), which gets rid of the computationally inefficient SVD step. In order to solve (1.2), we propose a projected gradient descent algorithm, along with a unified theory that integrates both optimization-theoretic and statistical analyses. We further summarize our main contributions as follows:

- Compared with existing work, our generic framework can be applied to a larger family of loss functions satisfying the restricted strong convexity and smoothness conditions (Negahban et al., 2009; Negahban and Wainwright, 2011; Klopp et al., 2014). We demonstrate the superiority of our framework through two concrete examples: robust matrix sensing and robust PCA.

- The gradient descent phase of our proposed algorithm matches the best-known robustness guarantee $O(1/r)$ (Hsu et al., 2011; Chen et al., 2013). Compared with existing robust PCA algorithms (Yi et al., 2016; Cherapanamjeri et al., 2016), our algorithm achieves improved computational complexity $O\bigl(r^3 d \log d \log(1/\epsilon)\bigr)$, while matching the optimal sample complexity $O(r^2 d \log d)$ for Burer-Monteiro factorization-based low-rank matrix recovery (Zheng and Lafferty, 2016) under the incoherence condition.

- To ensure the linear convergence rate, from the algorithmic perspective, we construct a double thresholding operator, which integrates both hard thresholding (Blumensath and Davies, 2009) and truncation operators (Yi et al., 2016); in terms of technical proof, we propose a novel *structural Lipschitz gradient condition* for low-rank plus sparse matrices. We believe both the double thresholding operator and the structural Lipschitz gradient condition are of independent interest for other superposition-structured models to prove faster convergence rates.

**Notation.** Denote $[d]$ to be the index set $\{1, \ldots, d\}$. For any matrix $\mathbf{A} \in \mathbb{R}^{d_1 \times d_2}$, let $\mathbf{A}_{i,*}$, $\mathbf{A}_{*,j}$ be the $i$-th row and the $j$-th column of $\mathbf{A}$ respectively, and let $A_{ij}$ be its $(i,j)$-th entry. Let the $k$-th largest singular value of $\mathbf{A}$ be $\sigma_k(\mathbf{A})$, and let $\text{SVD}_r(\mathbf{A})$ be the rank-$r$ SVD of matrix $\mathbf{A}$. For any $d$-dimensional vector $\mathbf{x}$, the $\ell_q$ vector norm of $\mathbf{x}$ is denoted by $\|\mathbf{x}\|_q = (\Sigma_{i=1}^d |x_i|^q)^{1/q}$, where $1 \leq q < \infty$, and we use $\|\mathbf{x}\|_0$ to represent the number of nonzero entries of $\mathbf{x}$. For any $d_1$-by-$d_2$ matrix $\mathbf{A}$, we use $\|\mathbf{A}\|_2$ and $\|\mathbf{A}\|_F$ to denote the spectral norm and Frobenius norm respectively. And we use $\|\mathbf{A}\|_{\infty,\infty}$ to denote the elementwise infinity norm. In addition, we denote the number of nonzero entries in $\mathbf{A}$ by $\|\mathbf{A}\|_0$, and use $\|\mathbf{A}\|_{2,\infty}$ to represent the largest $\ell_2$-norm of its rows. For any two sequences $\{a_n\}$ and $\{b_n\}$, if there exists a constant $0 < C < \infty$ such that $a_n \leq C b_n$, then we write $a_n = O(b_n)$.

## 2 Related Work

In recent years, there has been a large body of literature (Chandrasekaran et al., 2010; Waters et al., 2011; Chen et al., 2011; Candès et al., 2011; Chandrasekaran et al., 2011; Hsu et al., 2011; Kyrillidis and Cevher, 2012; Agarwal et al., 2012b; Chen et al., 2013; Yang and Ravikumar, 2013; Klopp et al., 2014) focusing on the matrix recovery problems with low-rank plus sparse structures. For instance, Waters et al. (2011); Kyrillidis and Cevher (2012) studied the problem of robust matrix sensing, where they aim to recover both the low-rank matrix and the sparse matrix from compressive measurements. Chen et al. (2011) analyzed the robust multi-task learning, where they characterize the task relationships using a low-rank structure, and simultaneously identify the outlier tasks using a sparse structure. The most widely studied low-rank plus sparse matrix recovery problem is robust PCA (Candès et al., 2011; Chandrasekaran et al., 2011; Hsu et al., 2011; Chen et al., 2013; Klopp et al., 2014), where the goal is to recover the unknown low-rank matrix from corrupted observations. In the context of robust PCA, Candès et al. (2011) proved that under random corruption model, their algorithm enables exact recovery with constant fraction of corruptions. Meanwhile, Chandrasekaran et al. (2011) considered the deterministic corruption model and showed that the tolerance of row/column sparsity for exact recovery is in the order of $O(1/r\sqrt{d})$, which was further improved to $O(1/r)$ (Hsu et al., 2011; Chen et al., 2013). Instead of considering specific models, unified analysis framework was proposed to cover more general low-rank plus sparse matrix recovery problems. In particular, Agarwal et al. (2012b) proposed to analyze a class of estimators for noisy matrix decomposition based on convex optimization with decomposable regualrizer. Yang and Ravikumar (2013) considered a general class of $M$-estimators and provided a unified framework for superposition-structured statistical models.

However, most of the aforementioned work are based on convex relaxation, which involves a computationally expensive SVD step in each iteration. To address such computational barrier, various nonconvex optimization algorithms (Netrapalli et al., 2014; Chen and Wainwright, 2015; Gu et al., 2016; Yi et al., 2016; Cherapanamjeri et al., 2016) have been carried out to solve low-rank plus sparse matrix recovery with provable guarantees. For example, Netrapalli et al. (2014) proposed alternating projection to simultaneously estimate the low-rank and sparse structure, while Chen and Wainwright (2015) showed that projected gradient descent based algorithm will linearly converge to the unknown matrix decomposition under suitable initialization procedure. The most related work to ours is Yi et al. (2016), which proposed a fast gradient descent algorithm based on a novel truncation operator to recover the unknown low-rank matrix for robust PCA. Their approach allows for $O(1/r^{1.5})$ sparsity with improved computational efficiency upon previous work. Most recently, Cherapanamjeri et al. (2016) further improved the existing work in terms of robustness guarantee from $O(1/r^{1.5})$ to $O(1/r)$. It is worth noting that these several pieces of work are limited to robust PCA, thus unable to deal with more general problem settings, such as robust matrix sensing.

## 3 The Proposed Algorithm

Recall that our objective is to recover both unknown low-rank matrix $\mathbf{X}^* \in \mathbb{R}^{d_1 \times d_2}$ with rank-$r$ and unknown sparse matrix $\mathbf{S}^* \in \mathbb{R}^{d_1 \times d_2}$ with $s$ nonzero entries simultaneously. Let $\overline{\mathbf{U}}^* \mathbf{\Sigma}^* \overline{\mathbf{V}}^{*\top}$ be the SVD of $\mathbf{X}^*$, where $\overline{\mathbf{U}}^*, \overline{\mathbf{V}}^*$ are the left and right singular matrices respectively, and $\mathbf{\Sigma}^*$ denotes a $r$-by-$r$ diagonal matrix with elements $\sigma_1 \geq \sigma_2 \geq \ldots \geq \sigma_r > 0$. Denote the condition number of $\mathbf{X}^*$ by $\kappa = \sigma_1/\sigma_r$.

Intuitively speaking, in order to distinguish between low-rank and sparse structures, the unknown low-rank matrix $\mathbf{X}^*$ cannot be too sparse. For instance, if $\mathbf{X}^*$ is equal to zero in nearly all elements, the recovery task is impossible unless all of the entries are sampled (Gross, 2011). Therefore, we impose the following incoherence condition (Candès and Recht, 2009) on $\mathbf{X}^*$ to avoid such identifiability issue. More specifically, let the SVD of $\mathbf{X}^*$ be $\mathbf{X}^* = \overline{\mathbf{U}}^* \mathbf{\Sigma}^* \overline{\mathbf{V}}^{*\top}$, then we assume $\mathbf{X}^*$ is $\alpha$-incoherent

$$\|\overline{\mathbf{U}}^*\|_{2,\infty} \leq \sqrt{\frac{\alpha r}{d_1}} \quad \text{and} \quad \|\overline{\mathbf{V}}^*\|_{2,\infty} \leq \sqrt{\frac{\alpha r}{d_2}}, \tag{3.1}$$

where $\alpha \geq 1$ denotes the incoherence parameter. Thus, we let the constraint set $\mathcal{C}$ in (1.1) be the set of $\alpha$-incoherent matrices. In addition, suppose $\mathbf{S}^*$ has at most $\beta$-fraction nonzero entries for each row and column (Chandrasekaran et al., 2011), or in other words $\mathbf{S}^* \in \mathcal{K}$, where $\mathcal{K}$ is defined as follows

$$\mathcal{K} = \big\{ \mathbf{S} \in \mathbb{R}^{d_1 \times d_2} \,\big|\, \|\mathbf{S}\|_0 \leq s, \, \|\mathbf{S}_{i,*}\|_0 \leq \beta d_2, \, \forall i \in [d_1]; \, \|\mathbf{S}_{*,j}\|_0 \leq \beta d_1, \, \forall j \in [d_2] \big\}.$$

Here, $\beta \in (0,1)$ represents the sparsity tolerance parameter. Recall (1.2), we define two constraint sets $\mathcal{C}_1, \mathcal{C}_2$ for $\mathbf{U}, \mathbf{V}$ respectively. Here, we provide the definitions of $\mathcal{C}_1, \mathcal{C}_2$ as follows

$$\mathcal{C}_1 = \left\{ \mathbf{U} \in \mathbb{R}^{d_1 \times r} \Big| \|\mathbf{U}\|_{2,\infty} \leq \sqrt{\frac{\alpha r}{d_1}} \|\mathbf{Z}^0\|_2 \right\}, \mathcal{C}_2 = \left\{ \mathbf{V} \in \mathbb{R}^{d_2 \times r} \Big| \|\mathbf{V}\|_{2,\infty} \leq \sqrt{\frac{\alpha r}{d_2}} \|\mathbf{Z}^0\|_2 \right\}, \quad (3.2)$$

where $\mathbf{Z}^0 = [\mathbf{U}^0; \mathbf{V}^0]$ represents the initial solution of Algorithm 1, and we will further demonstrate in the theoretical analysis that $\mathbf{U}^* \in \mathcal{C}_1, \mathbf{V}^* \in \mathcal{C}_2$. Furthermore, in order to guarantee the uniqueness of the optimal solution to optimization problem (1.2), following Tu et al. (2015); Zheng and Lafferty (2016); Park et al. (2016), we impose an additional regularizer to penalize the scale difference between $\mathbf{U}$ and $\mathbf{V}$. In other words, we aim to estimate the unknown parameter set $(\mathbf{U}^*, \mathbf{V}^*, \mathbf{S}^*)$ by solving the following optimization problem

$$\min_{\mathbf{U} \in \mathcal{C}_1, \mathbf{V} \in \mathcal{C}_2, \mathbf{S} \in \mathcal{K}} F_n(\mathbf{U}, \mathbf{V}, \mathbf{S}) := \mathcal{L}_n(\mathbf{U}\mathbf{V}^\top + \mathbf{S}) + \frac{1}{8}\|\mathbf{U}^\top \mathbf{U} - \mathbf{V}^\top \mathbf{V}\|_F^2. \quad (3.3)$$

Next, we present our proposed generic gradient descent algorithm for solving (3.3), as displayed in Algorithm 1. For low-rank structure, we perform gradient descent on $\mathbf{U}$ and $\mathbf{V}$ respectively, followed by projection onto the corresponding constraint sets $\mathcal{C}_1$ and $\mathcal{C}_2$. For sparse structure, we perform double thresholding, which integrates both the hard thresholding operator in Blumensath and Davies (2009) and the truncation operator in Yi et al. (2016), to ensure the output estimator $\mathbf{S}^t$ is sparse and has at most $\beta$-fraction nonzero entries per row and column as well.

---

**Algorithm 1** Gradient Descent Phase

**Input:** Sample loss function $\mathcal{L}_n$; step size $\tau, \eta$; total number of iterations $T$; parameters $\gamma, \gamma'$; initial solution $(\mathbf{U}^0, \mathbf{V}^0, \mathbf{S}^0)$.

$\mathbf{Z}^0 = [\mathbf{U}^0; \mathbf{V}^0]$; Let $\mathcal{C}_1, \mathcal{C}_2$ be defined in (3.2).

**for:** $t = 0, 1, 2, \ldots, T-1$ **do**

$\mathbf{S}^{t+1} = \mathcal{T}_{\gamma\beta} \circ \mathcal{H}_{\gamma's}\big(\mathbf{S}^t - \tau \nabla_{\mathbf{S}} \mathcal{L}_n(\mathbf{U}^t \mathbf{V}^{t\top} + \mathbf{S}^t)\big)$

$\mathbf{U}^{t+1} = \mathcal{P}_{\mathcal{C}_1}\big(\mathbf{U}^t - \eta \nabla_{\mathbf{U}} \mathcal{L}_n(\mathbf{U}^t \mathbf{V}^{t\top} + \mathbf{S}^t) - \frac{1}{2}\eta \mathbf{U}^t(\mathbf{U}^{t\top}\mathbf{U}^t - \mathbf{V}^{t\top}\mathbf{V}^t)\big)$

$\mathbf{V}^{t+1} = \mathcal{P}_{\mathcal{C}_2}\big(\mathbf{V}^t - \eta \nabla_{\mathbf{V}} \mathcal{L}_n(\mathbf{U}^t \mathbf{V}^{t\top} + \mathbf{S}^t) - \frac{1}{2}\eta \mathbf{V}^t(\mathbf{V}^{t\top}\mathbf{V}^t - \mathbf{U}^{t\top}\mathbf{U}^t)\big)$

**end for**

**Output:** $(\mathbf{U}^T, \mathbf{V}^T, \mathbf{S}^T)$

---

In Algorithm 1, we let $\mathcal{P}_{\mathcal{C}_i}$ be the projection operator onto the constraint set $\mathcal{C}_i$, where $i = 1, 2$. We define $\mathcal{H}_k : \mathbb{R}^{d_1 \times d_2} \to \mathbb{R}^{d_1 \times d_2}$ as the hard thresholding operator, which keeps the largest $k$ elements in terms of absolute value (i.e., magnitude) and sets the remaining entries as 0. In addition, we define $\mathcal{T}_\theta : \mathbb{R}^{d_1 \times d_2} \to \mathbb{R}^{d_1 \times d_2}$ as the truncation operator with parameter $\theta \in (0, 1)$ as follows: for all $(i, j) \in [d_1] \times [d_2]$, we have

$$[\mathcal{T}_\theta(\mathbf{S})]_{ij} := \begin{cases} S_{ij}, & \text{if } |S_{ij}| \geq |S_{i,*}^{(\theta d_2)}| \text{ and } |S_{ij}| \geq |S_{*,j}^{(\theta d_1)}|, \\ 0, & \text{otherwise}, \end{cases}$$

where $S_{i,*}^{(k)}$ and $S_{*,j}^{(k)}$ denote the $k$-th largest magnitude entries of $\mathbf{S}_{i,*}$ and $\mathbf{S}_{*,j}$ respectively.

It will be shown in later analysis that Algorithm 1 is guaranteed to converge to the unknown true parameters $(\mathbf{U}^*, \mathbf{V}^*, \mathbf{S}^*)$, as long as the initial solution $(\mathbf{U}^0, \mathbf{V}^0, \mathbf{S}^0)$ is close enough to $(\mathbf{U}^*, \mathbf{V}^*, \mathbf{S}^*)$.

Therefore, motivated by gradient hard thresholding (Blumensath and Davies, 2009) and singular value projection (Jain et al., 2010), we propose a novel initialization algorithm in Algorithm 2 to ensure the condition on the initial solutions. Based on singular value projection operator, we add an additional infinity norm constraint for low-rank structure. Specifically, we use $\mathcal{P}_{\lambda', \zeta^*} : \mathbb{R}^{d_1 \times d_2} \to \mathbb{R}^{d_1 \times d_2}$ to denote the constrained projection operator such that

$$\mathcal{P}_{\lambda', \zeta^*}(\mathbf{X}) = \mathrm{argmin}_{\mathrm{rank}(\mathbf{Y}) \leq r, \|\mathbf{Y}\|_{\infty,\infty} \leq \zeta^*} \|\mathbf{Y} - \mathbf{X}\|_F,$$

where $\zeta^*$ is defined as $\zeta^* = c_0 \alpha r \kappa / \sqrt{d_1 d_2}$, with $c_0$ as a predetermined upper bound of $\sigma_r(\mathbf{X}^*)$. According to (3.1), we have $\|\mathbf{X}^*\|_{\infty,\infty} \leq \|\overline{\mathbf{U}}^*\|_{2,\infty} \cdot \sigma_1(\mathbf{X}^*) \cdot \|\overline{\mathbf{V}}^*\|_{2,\infty} \leq \zeta^*$. In practice, we can use Dykstra's alternating projection algorithm (Bauschke and Borwein, 1994) to solve the projection operator $\mathcal{P}_{\lambda', \zeta^*}$. According to Lewis and Malick (2008) and Lewis et al. (2009), the alternating projection achieves a local R-linear convergence rate. In our experiments, we only perform one step alternating projection, which is sufficient to derive the fast convergence rate of Algorithm 1. We believe this alternating projection step is efficient, and will further investigate it theoretically.

---

**Algorithm 2** Initialization Phase

**Input:** Sample loss function $\mathcal{L}_n$; step size $\tau', \eta'$; total number of iterations $L$; parameters $\lambda, \lambda'$.
$\mathbf{X}_0 = \mathbf{S}_0 = \mathbf{0}$
**for:** $\ell = 0, 1, 2, \ldots, L-1$ **do**
  $\mathbf{S}_{\ell+1} = \mathcal{H}_{\lambda s}(\mathbf{S}_\ell - \tau' \nabla_{\mathbf{S}} \mathcal{L}_n(\mathbf{X}_\ell + \mathbf{S}_\ell))$
  $\mathbf{X}_{\ell+1} = \mathcal{P}_{\lambda', \zeta^*}(\mathbf{X}_\ell - \eta' \nabla_{\mathbf{X}} \mathcal{L}_n(\mathbf{X}_\ell + \mathbf{S}_\ell))$
**end for**
$[\overline{\mathbf{U}}^0, \mathbf{\Sigma}^0, \overline{\mathbf{V}}^0] = \mathrm{SVD}_r(\mathbf{X}_L)$
$\mathbf{U}^0 = \overline{\mathbf{U}}^0 (\mathbf{\Sigma}^0)^{1/2}, \mathbf{V}^0 = \overline{\mathbf{V}}^0 (\mathbf{\Sigma}^0)^{1/2}, \mathbf{S}^0 = \mathbf{S}_L$
**Output:** $(\mathbf{U}^0, \mathbf{V}^0, \mathbf{S}^0)$

---

## 4 Main Theory

Let $\mathbf{U}^* = \overline{\mathbf{U}}^*(\mathbf{\Sigma}^*)^{1/2}$, $\mathbf{V}^* = \overline{\mathbf{V}}^*(\mathbf{\Sigma}^*)^{1/2}$ and $\mathbf{Z}^* = [\mathbf{U}^*; \mathbf{V}^*]$ be the unknown matrix parameters we aim to estimate. Following Jain et al. (2013); Tu et al. (2015); Zheng and Lafferty (2016), we introduce the following distance metric.

**Definition 4.1.** For any $\mathbf{Z} \in \mathbb{R}^{(d_1+d_2) \times r}$, define the distance metric between $\mathbf{Z}$ and $\mathbf{Z}^*$ with respect to the optimal rotation as $d(\mathbf{Z}, \mathbf{Z}^*)$ such that $d(\mathbf{Z}, \mathbf{Z}^*) = \min_{\mathbf{R} \in \mathbb{Q}_r} \|\mathbf{Z} - \mathbf{Z}^* \mathbf{R}\|_F$, where $\mathbb{Q}_r$ denotes the set of $r$-by-$r$ orthonormal matrices.

Next, we lay out the restricted strong convexity (RSC) and restricted strong smoothness (RSS) conditions (Negahban et al., 2009; Loh and Wainwright, 2013) regarding $\mathcal{L}_n$. Note that our problem includes both low-rank and sparse structures, thus we assume the restricted strong smoothness and convexity conditions hold for one structure given the other.

**Condition 4.2** (Low Rank Structure). For all fixed sparse matrix $\mathbf{S} \in \mathbb{R}^{d_1 \times d_2}$ with at most $\widetilde{s}$ nonzero entries, we assume $\mathcal{L}_n$ is restricted strongly convex with parameter $\mu_1$ and restricted strongly smooth with parameter $L_1$ with respect to the low-rank structure, such that for all matrices $\mathbf{X}_1, \mathbf{X}_2 \in \mathbb{R}^{d_1 \times d_2}$ with rank at most $\widetilde{r}$, we have

$$\mathcal{L}_n(\mathbf{X}_2 + \mathbf{S}) \geq \mathcal{L}_n(\mathbf{X}_1 + \mathbf{S}) + \langle \nabla_{\mathbf{X}}\mathcal{L}_n(\mathbf{X}_1 + \mathbf{S}), \mathbf{X}_2 - \mathbf{X}_1 \rangle + \frac{\mu_1}{2}\|\mathbf{X}_2 - \mathbf{X}_1\|_F^2,$$

$$\mathcal{L}_n(\mathbf{X}_2 + \mathbf{S}) \leq \mathcal{L}_n(\mathbf{X}_1 + \mathbf{S}) + \langle \nabla_{\mathbf{X}}\mathcal{L}_n(\mathbf{X}_1 + \mathbf{S}), \mathbf{X}_2 - \mathbf{X}_1 \rangle + \frac{L_1}{2}\|\mathbf{X}_2 - \mathbf{X}_1\|_F^2.$$

Here, $\widetilde{r}, \widetilde{s}$ satisfy $r \leq \widetilde{r} \leq Cr$ and $s \leq \widetilde{s} \leq Cs$, where $C \geq 1$ is a universal constant to be determined.

**Condition 4.3** (Sparse Structure). For all fixed rank-$\widetilde{r}$ matrix $\mathbf{X} \in \mathbb{R}^{d_1 \times d_2}$, we assume $\mathcal{L}_n$ is restricted strongly convex with parameter $\mu_2$ and restricted strongly smooth with parameter $L_2$ in terms of the sparse structure, such that for all matrices $\mathbf{S}_1, \mathbf{S}_2 \in \mathbb{R}^{d_1 \times d_2}$ with at most $\widetilde{s}$ nonzero entries, we have

$$\mathcal{L}_n(\mathbf{X} + \mathbf{S}_2) \geq \mathcal{L}_n(\mathbf{X} + \mathbf{S}_1) + \langle \nabla_{\mathbf{S}}\mathcal{L}_n(\mathbf{X} + \mathbf{S}_1), \mathbf{S}_2 - \mathbf{S}_1 \rangle + \frac{\mu_2}{2}\|\mathbf{S}_2 - \mathbf{S}_1\|_F^2$$

$$\mathcal{L}_n(\mathbf{X} + \mathbf{S}_2) \leq \mathcal{L}_n(\mathbf{X} + \mathbf{S}_1) + \langle \nabla_{\mathbf{S}}\mathcal{L}_n(\mathbf{X} + \mathbf{S}_1), \mathbf{S}_2 - \mathbf{S}_1 \rangle + \frac{L_2}{2}\|\mathbf{S}_2 - \mathbf{S}_1\|_F^2.$$

Moreover, we propose the following novel structural Lipschitz gradient condition on the interaction term between low-rank and sparse structures.

**Condition 4.4** (Structural Lipschitz Gradient). Let $\mathbf{X}^*, \mathbf{S}^*$ be the unknown low-rank and sparse matrices respectively. For all low-rank matrices $\mathbf{X} \in \mathbb{R}^{d_1 \times d_2}$ with rank at most $\widetilde{r}$ and sparse matrices $\mathbf{S}$ with at most $\widetilde{s}$ nonzero entries, we assume

$$|\langle \nabla_{\mathbf{X}}\mathcal{L}_n(\mathbf{X}^* + \mathbf{S}) - \nabla_{\mathbf{X}}\mathcal{L}_n(\mathbf{X}^* + \mathbf{S}^*), \mathbf{X}\rangle - \langle \mathbf{S} - \mathbf{S}^*, \mathbf{X}\rangle| \leq K\|\mathbf{X}\|_F \cdot \|\mathbf{S} - \mathbf{S}^*\|_F,$$

$$|\langle \nabla_{\mathbf{S}}\mathcal{L}_n(\mathbf{X} + \mathbf{S}^*) - \nabla_{\mathbf{S}}\mathcal{L}_n(\mathbf{X}^* + \mathbf{S}^*), \mathbf{S}\rangle - \langle \mathbf{X} - \mathbf{X}^*, \mathbf{S}\rangle| \leq K\|\mathbf{X} - \mathbf{X}^*\|_F \cdot \|\mathbf{S}\|_F,$$

where $K \in (0, 1)$ is the structural Lipschitz gradient parameter depending on $r, s, d_1, d_2$ and $n$, which can be a sufficiently small constant, as long as sample size $n$ is large enough.

Roughly speaking, Condition 4.4 defines a variant of Lipschitz continuity on $\nabla \mathcal{L}_n$. Take the first inequality for example, the gradient is taken with respect to the low-rank structure, while the Lipschitz continuity is with respect to any $\widetilde{s}$-sparse matrix $\mathbf{S}$ and $\mathbf{S}^*$.

Finally, we assume that at $\mathbf{X}^* + \mathbf{S}^*$, the gradient of the sample loss function $\nabla \mathcal{L}_n$ is upper bounded in terms of both matrix spectral and infinity norms.

**Condition 4.5.** For a given sample size $n$ and tolerance parameter $\delta \in (0, 1)$, we let $\epsilon_1(n, \delta)$ and $\epsilon_2(n, \delta)$ be the smallest scalars such that

$$\|\nabla_{\mathbf{X}}\mathcal{L}_n(\mathbf{X}^* + \mathbf{S}^*)\|_2 \leq \epsilon_1(n, \delta) \quad \text{and} \quad \|\nabla_{\mathbf{S}}\mathcal{L}_n(\mathbf{X}^* + \mathbf{S}^*)\|_{\infty,\infty} \leq \epsilon_2(n, \delta)$$

hold with probability at least $1 - \delta$. Here $\epsilon_1(n, \delta)$ and $\epsilon_2(n, \delta)$ depend on $n$ and $\delta$.

## 4.1 Results for the Generic Model

Now we provide main results for our proposed algorithms. The following theorem guarantees the linear convergence rate of Algorithm 1 under proper conditions. We introduce the following distance metric to measure the estimation error of the output

$$D(\mathbf{Z}, \mathbf{S}) = d^2(\mathbf{Z}, \mathbf{Z}^*) + \|\mathbf{S} - \mathbf{S}^*\|_F^2 / \sigma_1. \tag{4.1}$$

The parameter $1/\sigma_1$ comes from the scale difference between $\mathbf{X} = \mathbf{U}\mathbf{V}^\top$ and $\mathbf{Z} = [\mathbf{U}; \mathbf{V}]$, or specifically, $\|\mathbf{X} - \mathbf{X}^*\|_F^2 \leq c\sigma_1 d^2(\mathbf{Z}, \mathbf{Z}^*)$ for some constant $c$.

**Theorem 4.6.** Let $\mathbf{X}^* = \mathbf{U}^* \mathbf{V}^{*\top}$ be the unknown rank-$r$ matrix that satisfies (3.1) and $\mathbf{S}^*$ be the unknown $s$-sparse matrix with at most $\beta$-fraction nonzero entries per row/column. Suppose the sample loss function $\mathcal{L}_n$ satisfies Conditions 4.2 - 4.5. There exist constants $c_1, c_2, c_3, c_4$ such that if set step size $\eta = c_1/\sigma_1, \tau = c_2/L_2$ and $\gamma, \gamma'$ large enough, under condition $\beta \leq c_3/(\alpha r \kappa)$, for any initial estimator $(\mathbf{Z}^0, \mathbf{S}^0)$ satisfying $D(\mathbf{Z}^0, \mathbf{S}^0) \leq c_4^2 \sigma_r$, with probability at least $1 - \delta$, the $t$-th iterate of Algorithm 1 satisfies

$$D(\mathbf{Z}^t, \mathbf{S}^t) \leq \rho^t D(\mathbf{Z}^0, \mathbf{S}^0) + \frac{\Gamma_1' r \epsilon_1^2(n, \delta) + \Gamma_2' s \epsilon_2^2(\epsilon, \delta)}{(1-\rho)\sigma_1},$$

where $D(\mathbf{Z}, \mathbf{S})$ is defined in (4.1), and $\rho = \max\{1 - \eta\mu_1\sigma_r/80, 1 - \mu_2\tau/32\} \in (0, 1)$ denotes the contraction parameter, provided that the sample size $n$ is large enough such that the structural Lipschitz parameter $K$ is sufficiently small and $\Gamma_1' r \epsilon_1^2(n, \delta) + \Gamma_2' s \epsilon_2^2(n, \delta) \leq (1-\rho) c_2^2 \sigma_1 \sigma_r$. Here, $\Gamma_1', \Gamma_2'$ are absolute constants depending on $\mu_1, \mu_2, L_1, L_2, \gamma$ and $\gamma'$.

**Remark 4.7.** Theorem 4.6 establishes the linear convergence rate of Algorithm 1. The right hand side of the contraction inequality consists of two terms: The first term corresponds to the optimization error, while the other term represents the statistical error. When considering the noiseless case, only the optimization error term exists. It is worth noting that our robustness guarantee required for the gradient descent phase matches the best-known results $O(1/r)$ in Hsu et al. (2011); Chen et al. (2013); Cherapanamjeri et al. (2016).

The next theorem provides the theoretical guarantee of Algorithm 2 regarding the initialization.

**Theorem 4.8** (Initialization)**.** Under the same condition as in Theorem 4.6, suppose $L_1/\mu_1 \in (1, 6)$, $L_2/\mu_2 \in (1, 4/3)$, $\mu_1 \geq 1/3$ and $K \leq c \cdot \min\{\mu_1, \mu_2\}$, where $c$ is a small constant. For any $\ell \geq 0$, with step size $\eta' = 1/(6\mu_1)$, $\tau' = 3/(4\mu_2)$ and $\lambda, \lambda'$ sufficient large, the $\ell$-th iterate of Algorithm 2 satisfies

$$\|\mathbf{X}_\ell - \mathbf{X}^*\|_F + \|\mathbf{S}_\ell - \mathbf{S}^*\|_F \leq \rho'^\ell(\|\mathbf{X}^*\|_F + \|\mathbf{S}^*\|_F) + \Gamma_1\sqrt{r}\epsilon_1(n,\delta) + \Gamma_2\sqrt{s}\epsilon_2(n,\delta) + \Gamma_3 \frac{c_0\alpha r\kappa\sqrt{s}}{\sqrt{d_1 d_2}} \tag{4.2}$$

with probability at least $1 - \delta$, where $\rho' = \max\{\rho_1', \rho_2'\} \in (0, 19/20)$ with $\rho_1' = (1 + 2/\sqrt{\lambda' - 1}) \cdot (\sqrt{1 - \mu_1\eta'} + \tau'K)$ and $\rho_2' = (1 + 2/\sqrt{\lambda - 1}) \cdot (\sqrt{1 - \mu_2\tau'} + \eta'(1 + K))$. Here, $\Gamma_1, \Gamma_2$ and $\Gamma_3$ are absolute constants depending on $\mu_1, \mu_2, \lambda, \lambda', r$ and $s$.

Combined both Theorem 4.6 and Theorem 4.8, we arrive at the following main result regarding our method.

**Theorem 4.9.** Suppose the rank-$r$ matrix $\mathbf{X}^*$ satisfies (3.1) and the $s$-sparse matrix $\mathbf{S}^*$ has at most $\beta$-fraction nonzero entries per row/column. Assume the sample loss function $\mathcal{L}_n$ satisfies Conditions 4.2 - 4.5. There exist constants $c_1, c_2, c_3, c_4, c_5$, provided that $\beta \leq c_1/(\alpha r \kappa)$, $s \leq c_2 d_1 d_2/(\alpha^2 r^2 \kappa^2)$ and the sample size $n$ large enough, if perform $L = O(1)$ iterations in Algorithm 2 with step size $\eta' = 1/(6\mu_1), \tau' = 3/(4\mu_2)$ and parameters $\lambda, \lambda'$ large enough, the output of Algorithm 1, with step size $\eta = c_3/\sigma_1, \tau = c_4/L_2$ and parameters $\gamma, \gamma'$ large enough, satisfies

$$D(\mathbf{Z}^T, \mathbf{S}^T) \leq \rho^T \cdot c_5 \sigma_r + \Gamma \cdot \frac{r\epsilon_1^2(n,\delta) + s\epsilon_2^2(\epsilon,\delta)}{(1-\rho)\sigma_1}$$

with probability at least $1 - \delta$, where $\mathbf{Z}^T = [\mathbf{U}^T; \mathbf{V}^T]$, $\rho$ denotes the contraction parameter defined in Theorem 4.6, and $\Gamma$ is an absolute constant depending on $\mu_1$, $\mu_2$, $L_1$, $L_2$, $\gamma$ and $\gamma'$.

**Remark 4.10.** In Theorem 4.9, we require the tolerance of overall sparsity for $\mathbf{S}^*$ is in the order of $O(d_1 d_2/r^2)$, which is near optimal compared with existing work regarding robust PCA. This suboptimality is due to the more general settings we considered in this work. Specifically, we aim to derive the recovery results for both low-rank and sparse structures, which is applicable for more general loss function beyond robust PCA, such as robust matrix sensing.

## 4.2 Results for Specific Models

Our main result for the generic model can be readily applied to specific models. In the following discussions, we assume $d_1 = d_2 = d$ for simplicity.

**Robust Matrix Sensing.** The problem of robust matrix sensing (Waters et al., 2011; Kyrillidis and Cevher, 2012) has a broad range of applications in video recovery (Cevher et al., 2008) and hyperspectral imaging (Chakrabarti and Zickler, 2011). Specifically, we observe $\boldsymbol{y} = \mathcal{A}(\mathbf{X}^* + \mathbf{S}^*) + \boldsymbol{\epsilon}$, where $\mathbf{X}^*, \mathbf{S}^*$ are the unknown low-rank and sparse matrices respectively, and $\boldsymbol{\epsilon}$ denotes the noise vector. Let $\mathcal{A} : \mathbb{R}^{d_1 \times d_2} \to \mathbb{R}^n$ be a linear measurement operator such that $\mathcal{A}(\mathbf{X}^* + \mathbf{S}^*) = (\langle \mathbf{A}_1, \mathbf{X}^* + \mathbf{S}^* \rangle, \ldots, \langle \mathbf{A}_n, \mathbf{X}^* + \mathbf{S}^* \rangle)^\top$, where each random matrix $\mathbf{A}_i \in \mathbb{R}^{d_1 \times d_2}$ is called sensing matrix, whose entries follow i.i.d. standard normal distribution. In the following discussions, we call $\mathcal{A}$ the standard normal linear operator for simplicity. Thus the sample loss function derived from robust matrix sensing is

$$\mathcal{L}_n(\mathbf{U}\mathbf{V}^\top + \mathbf{S}) := (2n)^{-1}\|\mathbf{y} - \mathcal{A}_n(\mathbf{U}\mathbf{V}^\top + \mathbf{S})\|_2^2.$$

Next, we present the theoretical guarantee of our proposed algorithm for robust matrix sensing.

**Corollary 4.11.** Suppose $\mathbf{X}^*$, $\mathbf{S}^*$ and $\mathcal{L}_n$ satisfy the same conditions as in Theorem 4.9. Consider robust matrix sensing with standard normal linear operator $\mathcal{A}$ and noise vector $\boldsymbol{\epsilon}$, whose entries follow i.i.d. sub-Gaussian distribution with parameter $\nu$. There exist constants $\{c_i\}_{i=1}^{10}$ such that under condition that sample size $n \geq c_1(rd + s)\log d$, robustness guarantee $\beta \leq 1/(c_2 r \kappa)$ and $s \leq c_3 d_1 d_2/(\alpha^2 r^2 \kappa^2)$, if we perform $L = O(1)$ iterations in Algorithm 2 with appropriate step size $\eta', \tau'$ and parameters $\lambda, \lambda'$ large enough, then with probability at least $1 - c_4/d$, the output of Algorithm 1, with $\eta = c_5/\sigma_1$, $\tau = c_6$ and $\gamma, \gamma'$ large enough, satisfies

$$D(\mathbf{Z}^T, \mathbf{S}^T) \leq \rho^T D(\mathbf{Z}^0, \mathbf{S}^0) + c_7 \nu^2 \frac{rd}{n} + c_8 \nu^2 \frac{s \log d}{n},$$

where $\rho = \max\{1 - c_9 \eta \sigma_r, 1 - c_{10} \tau\}$.

**Remark 4.12.** According to Corollary 4.11, in the noiseless setting, our algorithm can achieve exactly recovery for both low-rank and sparse matrices. In addition, to establish the structural Lipschitz gradient condition, we require the sample size $n = O((rd + s)\log d)$. If $s \leq rd$, it achieves the optimal sample complexity as that of standard matrix sensing (Recht et al., 2010; Tu et al., 2015; Wang et al., 2016) up to a logarithmic term. In the noisy setting, after $O(\kappa \log(n/(rd + s \log d)))$ number of iterations, our estimator achieves $O((rd + s \log d)/n)$ statistical error. The term $O(rd/n)$ corresponds to the statistical error for the low-rank matrix recovery, which matches the minimax lower bound of standard noisy matrix sensing (Negahban and Wainwright, 2011). The other term $O(s \log d/n)$ corresponds to the statistical error for the sparse matrix recovery, which also matches the minimax lower bound of sparse matrix regression (Raskutti et al., 2011). We notice that Waters et al. (2011) studied the same problem using a greedy algorithm. However, there is no theoretical guarantee of their algorithm.

Table 1: Complexity comparisons among different algorithms for robust PCA under partially observed model.

| Algorithm | Sample Complexity | Computational Complexity |
|---|---|---|
| Fast RPCA (Yi et al., 2016) | $O(r^2 d \log d)$ | $O(r^4 d \log d \log(1/\epsilon))$ |
| PG-RMC (Cherapanamjeri et al., 2016) | $O(r^2 d \log^2 d \log^2(\sigma_1/\epsilon))$ | $O(r^3 d \log^2 d \log^2(\sigma_1/\epsilon))$ |
| This paper | $O(r^2 d \log d)$ | $O(r^3 d \log d \log(1/\epsilon))$ |

**Robust PCA.** We proceed to consider robust PCA. More specifically, we observe a data matrix $\mathbf{Y} \in \mathbb{R}^{d_1 \times d_2}$ such that $\mathbf{Y} = \mathbf{X}^* + \mathbf{S}^*$, where $\mathbf{X}^*, \mathbf{S}^* \in \mathbb{R}^{d_1 \times d_2}$ are the unknown low-rank and sparse matrices. We consider the uniform observation model

$$Y_{jk} := \begin{cases} X_{jk}^* + S_{jk}^* + E_{jk}, & \text{for any } (j,k) \in \Omega \\ *, & \text{otherwise,} \end{cases}$$

where $\Omega \subseteq [d_1] \times [d_2]$ denotes the observed index set such that for any $(j,k) \in \Omega$, $j \sim \text{uniform}([d_1])$ and $k \sim \text{uniform}([d_2])$. Here $\mathbf{E} \in \mathbb{R}^{d_1 \times d_2}$ is the noise matrix, where each entry of $\mathbf{E}$ follows i.i.d.

normal distribution with variance $\nu^2/(d_1 d_2)$ such that each observation has a dimension-free signal-to-noise ratio. In addition, we assume that $\mathbf{S}^*$ is not restrictive to $\Omega$. Therefore, for robust PCA, we have the following objective loss function

$$\mathcal{L}_\Omega(\mathbf{U}\mathbf{V}^\top + \mathbf{S}) := (2p)^{-1}\sum_{(j,k)\in\Omega}(\mathbf{U}_{j*}\mathbf{V}_{k*}^\top + S_{jk} - Y_{jk})^2.$$

In the following discussions, we are going to consider both full observation model ($p = 1$) and partial observation model ($0 < p < 1$) for robust PCA.

**Corollary 4.13** (Fully Observed RPCA). Suppose $\mathbf{X}^*$, $\mathbf{S}^*$ and $\mathcal{L}_n$ satisfy the same conditions as in Theorem 4.9. There exist constants $\{c_i\}_{i=1}^9$ such that under the robustness guarantee $\beta \leq 1/(c_1 r\kappa)$ and $s \leq c_2 d_1 d_2/(\alpha^2 r^2 \kappa^2)$, if we perform $L = O(1)$ iterations in Algorithm 2 with appropriate step size $\eta', \tau'$ and $\lambda, \lambda'$ large enough, then with probability at least $1 - c_3/d$, the output of Algorithm 1, with step size $\eta = c_4/\sigma_1, \tau = c_5$ and $\gamma, \gamma'$ large enough, satisfies

$$D(\mathbf{Z}^T, \mathbf{S}^T) \leq \rho^T D(\mathbf{Z}^0, \mathbf{S}^0) + c_6 \nu^2 \frac{rd}{d_1 d_2} + c_7 \nu^2 \frac{s \log d}{d_1 d_2},$$

where $\rho = \max\{1 - c_8 \eta \sigma_r, 1 - c_9 \tau\}$.

**Remark 4.14.** Corollary 4.13 suggests that in the noiseless setting, the statistical error terms equal to zero. Therefore, our algorithm can exactly recover both low-rank and sparse matrices. Note that Agarwal et al. (2012b) also analyzed this model using M-estimators. However, their results include an additional standardized error term $\widetilde{\alpha}^2 s/(d_1 d_2)$, where $\widetilde{\alpha}$ is the maximum magnitude among entries of $\mathbf{X}^*$.

For the partially observed robust PCA, in order to provide the statistical guarantee for the sparse structure, we further impose an infinity norm constraint for $\mathbf{S}^*$ such that $\|\mathbf{S}^*\|_{\infty,\infty} \leq \alpha_1/\sqrt{d_1 d_2}$. Note that this condition is essential for sparse recovery as illustrated in Klopp et al. (2014).

**Corollary 4.15** (Partially Observed RPCA). Consider partially observed robust PCA under uniform sampling model. Suppose $\mathbf{X}^*$, $\mathbf{S}^*$ and $\mathcal{L}_n$ satisfy the same conditions as in Theorem 4.9. There exist constants $\{c_i\}_{i=1}^{10}$ such that under the robustness $\beta \leq 1/(c_1 r\kappa)$, $s \leq c_2 d_1 d_2/(\alpha^2 r^2 \kappa^2)$, and sample size $n \geq c_3(r^2 d + s)\log d$, if we perform $L = O(1)$ number of iterations in Algorithm 2 with appropriate step size $\eta', \tau'$ and $\lambda, \lambda'$ large enough, then with probability at least $1 - c_4 d$, the output of Algorithm 1, with step size $\eta = c_5/\sigma_1, \tau = c_6$ and $\gamma, \gamma'$ large enough, satisfies

$$D(\mathbf{Z}^T, \mathbf{S}^T) \leq \rho^T D(\mathbf{Z}^0, \mathbf{S}^0) + c_7 \max\{\nu^2, \alpha^2 r\}\frac{rd \log d}{n} + c_8 \max\{\alpha_1^2, \nu^2\}\frac{s \log d}{n} + \frac{\alpha_1^2 s}{d_1 d_2}, \quad (4.3)$$

where $\rho = \max\{1 - c_9 \eta \sigma_r, 1 - c_{10}\tau\}$, and $\alpha_1 = \sqrt{d_1 d_2}\|\mathbf{S}^*\|_{\infty,\infty}$.

**Remark 4.16.** Note that the extra fourth term $\alpha_1^2 s/(d_1 d_2)$, on the right hand side of (4.3), is due to the unobserved corruption entries but is in fact dominated by the third term. Corollary 4.15 suggests that, after $O\big(\kappa \log\big(n/((r^2d+s)\log d)\big)\big)$ number of iterations, the output of our algorithm achieves $O\big((r^2 d + s)\log d/n\big)$ statistical error, and the term $O(r^2 d \log d/n)$ denotes the statistical error for the low-rank matrix. The term $O(s \log d/n)$ corresponds to the statistical error for the sparse matrix, which matches the minimax lower bound (Raskutti et al., 2011). Moreover, compared with existing nonconvex robust PCA algorithms (Yi et al., 2016; Cherapanamjeri et al., 2016), our algorithm achieves better computational complexity while matching the best-known sample complexity provided that $s \leq r^2 d$. The detailed comparisons are summarized in Table 1.

## 5 Experiments

In this section, we illustrate our experimental results to further demonstrate the performance of our proposed algorithm. Firstly, we investigate the performance of our algorithm with respect to robust matrix sensing and robust PCA on synthetic data. For robust matrix sensing, we compare our algorithm with SpaRCS (Waters et al., 2011). For robust PCA, we compare our algorithm with several state-of-the-art algorithms, including NcRPCA (Netrapalli et al., 2014), Fast RPCA (Yi et al., 2016), and PG-RMC (Cherapanamjeri et al., 2016). Note that all the experimental results are based on the optimal parameters, which are selected by cross validation, and averaged over 30 trials. Secondly, we compare our algorithm with several existing robust PCA algorithms, including GoDec (Zhou and Tao, 2011), Alt RPCA (Gu and Banerjee, 2016), and Fast RPCA (Yi et al., 2016) on real-world data.

### 5.1 Simulations on Sythetic Data

**Robust Matrix Sensing.** Our data are generated from the model $\mathbf{y} = \mathcal{A}(\mathbf{X}^* + \mathbf{S}^*) + \epsilon$. We generate $\mathbf{X}^* \in \mathbb{R}^{d_1 \times d_2}$ via $\mathbf{X}^* = \mathbf{U}^*\mathbf{V}^{*\top}$, where each entry of $\mathbf{U}^* \in \mathbb{R}^{d_1 \times r}$ and $\mathbf{V}^* \in \mathbb{R}^{d_2 \times r}$ is generated independently from standard Gaussian distribution. Besides, we generate the unknown sparse matrix $\mathbf{S}^*$ with each element sampled from Bernoulli distribution with parameter $1 - \beta$, where $\beta$ is the corruption parameter. The value of each nonzero element of $\mathbf{S}^*$ is drawn uniformly from $[-\alpha, \alpha]$. And each element of the sensing matrix $\mathbf{A}_i$ is drawn from i.i.d. standard normal distribution. For the noisy setting, we consider $\epsilon_i$ follows i.i.d. zero mean normal distribution with variance $\nu^2$.

For robust matrix sensing, we study the following experimental settings: (i) $d_1 = d_2 = 100, r = 3$; (ii) $d_1 = d_2 = 150, r = 4$; (iii) $d_1 = d_2 = 200, r = 5$. Furthermore, we consider the noiseless case, choose $\alpha = r, \beta = 0.1$, and set the the number of observation $n = 0.2 \cdot d_1 d_2$. First, we report the relative error and its standard deviation of low-rank structure ($\|\widehat{\mathbf{X}} - \mathbf{X}^*\|_F / \|\mathbf{X}^*\|_F$) as well as CPU time for different algorithms in Table 2. Note that we didn't show the results of sparse structure since it has similar performance to low-rank structure. The results show that our proposed algorithm outperforms the baseline algorithms in terms of relative error and CPU time.

In addition, we demonstrate the experimental results for robust matrix sensing regarding the linear convergence rate, sample complexity, and statistical rate of our proposed algorithm in Figure 1. Figure 1(a) and 1(c) illustrate the relative error $\|\widehat{\mathbf{X}} - \mathbf{X}^*\|_F^2 / \|\mathbf{X}^*\|_F^2$ in log scale versus number of iterations. Note that, we only lay out results under the setting $d_1 = d_2 = 100, r = 3$ with number of observations $n = 0.2 \cdot d_1 d_2$ to avoid redundancy. These plots demonstrate the linear rate of convergence of our algorithm. Figure 1(b) demonstrates the sample complexity requirement to achieve exact recovery for low-rank structure in the noiseless setting. Note that we say $\widehat{\mathbf{X}}$ achieves exact recovery if $\|\widehat{\mathbf{X}} - \mathbf{X}^*\|_F / \|\mathbf{X}^*\|_F \leq 10^{-3}$. It confirms our theoretical results regarding the sample complexity. The statistical error for the low-rank matrix is demonstrated in Figure 1(d), which is consistent with our result $O(rd/n)$.

**Robust PCA.** We generate the data according to $\mathbf{Y} = \mathbf{X}^* + \mathbf{S}^* + \mathbf{E}$, where the matrices $\mathbf{X}^*, \mathbf{S}^* \in$ $\mathbb{R}^{d_1 \times d_2}$ are generated by the same procedures as in robust matrix sensing. In the noisy setting, each element of the noisy matrix $\mathbf{E} \in \mathbb{R}^{d_1 \times d_2}$ is drawn from i.i.d. zero mean Gaussian distribution with variance $\nu^2$.

Table 2: Results for robust matrix sensing in terms of relative error ($\times 10^{-3}$) and CPU time.

| | $d_1 = d_2 = 100, r = 3$ | | $d_1 = d_2 = 150, r = 4$ | | $d_1 = d_2 = 200, r = 5$ | |
|---|---|---|---|---|---|---|
| **Methods** | Error | Time (s) | Error | Time (s) | Error | Time (s) |
| SpaRCS | 28.6 (1.24) | 30.21 | 26.83 (1.18) | 107.71 | 25.73 (1.47) | 275.40 |
| Ours | 5.51 (0.60) | 24.31 | 5.33 (0.57) | 63.05 | 4.75 (0.62) | 177.24 |

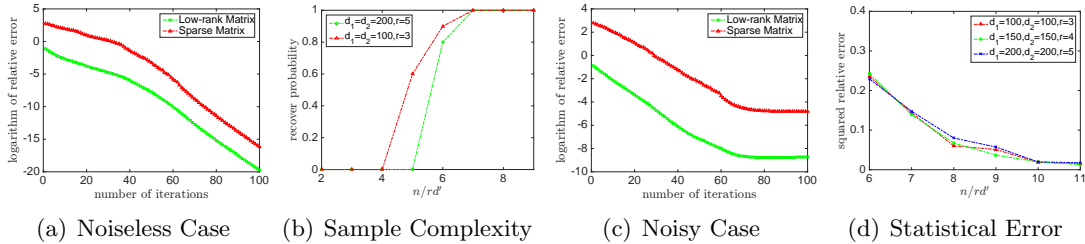

(a) Noiseless Case  (b) Sample Complexity  (c) Noisy Case  (d) Statistical Error

Figure 1: Experimental results for robust matrix sensing. (a),(c) Relative error in log scale vs. number of iterations in the noiseless and noisy settings respectively. (b) Recovering probability of low-rank matrix vs. scaled sample size in the noiseless setting. (d) Relative error vs. scaled sample size in the noisy setting.

For robust PCA, we study the following experimental settings: (i) $d_1 = d_2 = 100, r = 3$; (ii) $d_1 = d_2 = 1000, r = 20$; (iii) $d_1 = d_2 = 5000, r = 50$. In addition, we consider the noiseless case and choose $\alpha = r, \beta = 0.1$. Note that all the experimental results are based on the optimal parameters, which are selected by cross validation, and averaged over 30 trials. We report the averaged root mean square error (RMSE) and its standard deviation of low-rank structure ($\|\widehat{\mathbf{X}} - \mathbf{X}^*\|_F / \sqrt{d_1 d_2}$) as well as CPU time for different algorithms in Table 3. Note that we didn't show the RMSE of sparse structure since it has similar results to low-rank structure. The results show that all the algorithms perform well in terms of RMSE. However, our algorithm outperforms the baseline algorithms in terms of CPU time, especially when the dimension is large, which aligns well with our theory.

The experimental results for robust PCA regarding the linear convergence rate, sample complexity, and statistical rate are summarized in Figure 2. In detail, Figures 2(a) and 2(c) report the squared estimation error $\|\widehat{\mathbf{X}} - \mathbf{X}^*\|_F^2 / (d_1 d_2)$ in log scale versus number of iterations. Note that we only lay out the results under fully observed model with setting $d_1 = d_2 = 200, r = 5$, because other settings will give us similar plots, and we leave them out for simplicity. The results verify the linear convergence rate of our algorithm. In the noiseless setting, the sample complexity for achieving exactly recovery of the low-rank matrix is illustrated in Figure 2(b). The result of recovery probability indicates the sample complexity requirement $n = O(rd \log d)$ for robust PCA. Finally, Figure 2(d) demonstrates the statistical error for the low-rank matrix, which is at the order $O(rd \log d/n)$. Although our theoretical results suggest $O(r^2 d \log d)$ sample complexity and $O(r^2 d \log d)$ statistical error, the simulation results indicate that both the sample complexity and the statistical error scale linearly

Table 3: Results for robust PCA in terms of RMSE ($\times 10^{-3}$) and CPU time.

|  | $d_1 = d_2 = 100, r = 5$ | | $d_1 = d_2 = 1000, r = 20$ | | $d_1 = d_2 = 5000, r = 50$ | |
| --- | --- | --- | --- | --- | --- | --- |
| **Methods** | RMSE | Time (s) | RMSE | Time (s) | RMSE | Time (s) |
| NcRPCA | 5.12 (1.84) | 0.164 | 4.39 (2.27) | 2.12 | 5.48 (1.44) | 61.78 |
| Fast RPCA | 4.67 (0.22) | 0.179 | 4.25 (0.47) | 1.86 | 4.78 (0.39) | 43.16 |
| PG-RMC | 5.45 (2.15) | 0.185 | 3.97 (1.27) | 3.23 | 6.88 (1.06) | 89.29 |
| Ours | 3.97 (0.16) | 0.121 | 3.74 (0.15) | 1.54 | 3.67 (0.17) | 35.72 |

with $rd$.

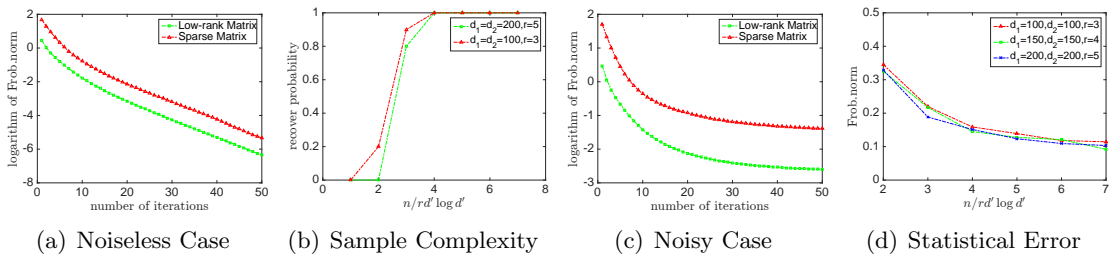

(a) Noiseless Case    (b) Sample Complexity    (c) Noisy Case    (d) Statistical Error

Figure 2: Experimental results for robust PCA. (a),(c) Squared estimation error in log scale vs. number of iterations in the noiseless and noisy settings respectively. (b) Recovering probability of low-rank matrix vs. scaled sample size in the noiseless setting. (d) Squared relative error vs. scaled sample size in the noisy setting.

## 5.2 Real-World Experiments

We evaluate our proposed method through the problem of background modeling (Li et al., 2004). The goal of background modeling is to reveal the correlation between video frames, reconstruct the static background and detect moving objects in foreground. More specifically, a video sequence has a low-rank plus sparse structure, because backgrounds of all frames are related, while the moving objects in foregrounds are sparse and independent. Due to this superstructure property, robust PCA has been widely used for background modeling (Zhou and Tao, 2011; Gu and Banerjee, 2016; Yi et al., 2016). We apply our proposed method to one surveillance video (Li et al., 2004), which includes 200 frames with the resolution $144 \times 176$. In particular, we convert each frame to a vector and form a $25344 \times 200$ data matrix $\mathbf{Y}$. Figure 3 illustrates the estimated background frames (i.e., low-rank structure) by different methods. The background frames estimated by our method are comparable to others. However, compared with GoDec (taking about 32 seconds), Alt RPCA (taking about 22 seconds), and Fast RPCA (taking about 26 seconds), our proposed method only takes around 18 seconds to process the video sequence. All of these experimental results demonstrate the superiority of our proposed method.

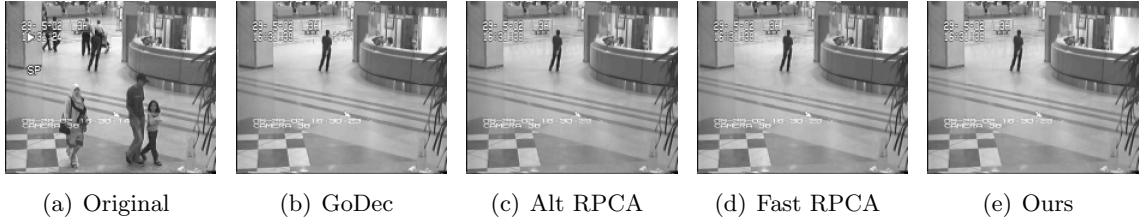

| (a) Original | (b) GoDec | (c) Alt RPCA | (d) Fast RPCA | (e) Ours |

Figure 3: Background reconstruction of *Hall of a business building* video. (a) The original frame. (b)-(e) Background frames estimated by GoDec (Zhou and Tao, 2011), Alt RPCA (Gu and Banerjee, 2016), Fast RPCA (Yi et al., 2016), and our algorithm respectively.

## 6 Conclusions

We proposed a generic framework for low-rank plus sparse matrix recovery via nonconvex optimization, which integrates both optimization-theoretic and statistical analyses. However, there still exist some open problems along this line of research, e.g., (1) How to achieve $O(1/r)$ robustness guarantee for the initialization phase targeted for general loss functions? (2) How to improve the sample complexity from $O(r^2 d \log d)$ to $O(rd \log d)$ for robust PCA based on nonconvex optimization? We hope these open problems can be addressed in future study.

## A  Proof of the Main Theory

In this section, we establish the proof of our main theory. Before proceeding any further, we introduce the following notations. For any index set $\Omega \subseteq [d_1] \times [d_2]$, let $\Omega_{i,*}$ and $\Omega_{*,j}$ be the $i$-th row and $j$-th column of $\Omega$ respectively. Denote the column and row space of $\mathbf{A}$ by $\text{col}(\mathbf{A})$ and $\text{row}(\mathbf{A})$ respectively. Let the top $d_1 \times r$ and bottom $d_2 \times r$ matrices of any matrix $\mathbf{A} \in \mathbb{R}^{(d_1+d_2)\times r}$ be $\mathbf{A}_U$ and $\mathbf{A}_V$ respectively. Let the nuclear norm of any matrix $\mathbf{A}$ be $\|\mathbf{A}\|_*$. Denote $\mathbf{Z} = [\mathbf{U}; \mathbf{V}] \in \mathbb{R}^{(d_1+d_2) \times r}$, then according to (3.3), we reformulate the regularized objective function as follows

$$\widetilde{F}_n(\mathbf{Z}, \mathbf{S}) = F_n(\mathbf{U}, \mathbf{V}, \mathbf{S}) = \mathcal{L}_n(\mathbf{U}\mathbf{V}^\top + \mathbf{S}) + \frac{1}{8}\|\mathbf{U}^\top\mathbf{U} - \mathbf{V}^\top\mathbf{V}\|_F^2. \quad (\text{A.1})$$

Therefore, the corresponding gradient regarding to $\mathbf{Z}$ is as follows

$$\nabla_{\mathbf{Z}} \widetilde{F}_n(\mathbf{Z}, \mathbf{S}) = \begin{bmatrix} \nabla_{\mathbf{U}} \mathcal{L}_n(\mathbf{U}\mathbf{V}^\top + \mathbf{S}) + \frac{1}{2}\mathbf{U}(\mathbf{U}^\top\mathbf{U} - \mathbf{V}^\top\mathbf{V}) \\ \nabla_{\mathbf{V}} \mathcal{L}_n(\mathbf{U}\mathbf{V}^\top + \mathbf{S}) + \frac{1}{2}\mathbf{V}(\mathbf{U}^\top\mathbf{U} - \mathbf{V}^\top\mathbf{V}) \end{bmatrix}. \quad (\text{A.2})$$

### A.1 Proof of Theorem 4.6

In order to prove Theorem 4.6, we need to make use of the following lemmas. Since both low-rank and sparse structures exist in our model, it is necessary to derive the convergence results for both structures. Lemma A.1, proved in Section B.1 characterizes the convergence of the low rank structure, while Lemma A.2, proved in Section B.2 corresponds to the convergence of the sparse structure.

**Lemma A.1** (Convergence for Low-Rank Structure). Suppose the sample loss function $\mathcal{L}_n$ satisfies Conditions 4.2 and 4.4. Recall that $\mathbf{X}^* = \mathbf{U}^*\mathbf{V}^{*\top}$ is the unknown rank-$r$ matrix that satisfies (3.1), $\mathbf{S}^*$ is the unknown $s$-sparse matrix with at most $\beta$-fraction nonzero entries per row and column. There exist constants $c_1, c_2$ and $c_3$ such that if $\mathbf{Z}^t \in \mathbb{B}(c_2\sqrt{\sigma_r})$ with $c_2 \leq \min\{1/4, \sqrt{\mu_1'/[10(L_1 + 1 + 8/\mu_2)]}\}$, and we set the step size $\eta = c_1/\sigma_1$ with $c_1 \leq \min\{1/32, \mu_1/(192L_1^2)\}$, then the output of Algorithm 1 $\mathbf{Z}^t = [\mathbf{U}^t; \mathbf{V}^t]$ satisfies

$$d^2(\mathbf{Z}^{t+1}, \mathbf{Z}^*) \leq \rho_1 d^2(\mathbf{Z}^t, \mathbf{Z}^*) - \frac{\eta\mu_1}{4}\|\mathbf{X}^t - \mathbf{X}^*\|_F^2 + \Gamma_1\|\mathbf{S}^t - \mathbf{S}^*\|_F^2 + \Gamma_2\|\nabla_\mathbf{X}\mathcal{L}_n(\mathbf{X}^* + \mathbf{S}^*)\|_2^2,$$

provided that $\beta \leq 1/(c_3\alpha r\kappa)$ with $c_3 \geq 720(\gamma+1)\mu_2/\mu_1'$, where contraction parameter $\rho_1 = 1 - \eta\mu_1'\sigma_r/40$, $\mu_1' = \min\{\mu_1, 2\}$, $\Gamma_1 = 48\eta^2(1+K)^2\sigma_1 + \eta(\mu_2 + 4K^2/\mu_1)$, and $\Gamma_2 = 48\eta^2 r\sigma_1 + 2\eta(8r/\mu_1 + r/L_1)$.

**Lemma A.2** (Convergence for Sparse Structure). Suppose the sample loss function $\mathcal{L}_n$ satisfies Conditions 4.3 and 4.4. Recall that $\mathbf{X}^*$ is the unknown rank-$r$ matrix, $\mathbf{S}^*$ is the unknown $s$-sparse matrix. If we set the step size $\tau \leq 1/(3L_2)$ and choose appropriate parameters $\gamma, \gamma'$, then the output of Algorithm 1 satisfies

$$\|\mathbf{S}^{t+1} - \mathbf{S}^*\|_F^2 \leq \rho_2\|\mathbf{S}^t - \mathbf{S}^*\|_F^2 + \Gamma_3\|\mathbf{X}^t - \mathbf{X}^*\|_F^2 + \Gamma_4\|\mathbf{H}^t\|_F^2 + \Gamma_5\|\nabla_\mathbf{S}\mathcal{L}_n(\mathbf{X}^* + \mathbf{S}^*)\|_{\infty,\infty}^2.$$

Here, $\rho_2$ is the contraction parameter satisfying $\rho_2 = C(\gamma, \gamma') \cdot (1 - \mu_2\tau/4) < 1$, where $C(\gamma, \gamma')$ is defined in Theorem 4.6, and $\Gamma_3, \Gamma_4$ and $\Gamma_5$ are constants satisfying

$$\Gamma_3 = C(\gamma, \gamma') \cdot \left(\frac{4\tau K^2}{\mu_2} + 3\tau^2(1+K)^2\right), \quad \Gamma_4 = C(\gamma, \gamma') \cdot \frac{\tau(\gamma+1)\beta\alpha r\sigma_1}{\mu_2},$$

$$\Gamma_5 = C(\gamma, \gamma') \cdot \left(\frac{4\tau(\gamma'+1)s}{\mu_2} + 3\tau^2(2\gamma'+1)s\right).$$

Now we are ready to prove Theorem 4.6.

*Proof of Theorem 4.6.* Given a fixed step size $\tau$, we set $\gamma, \gamma'$ such that $\gamma' \geq 1 + 256/(\mu_2^2\tau^2)$ and $\gamma \geq \max\{5, 1 + 64^2/(\mu_2\tau)^2\}$, then we obtain

$$\rho_2 = \left(1 + \sqrt{\frac{2}{\gamma-1}}\right)^2 \cdot \left(1 + \frac{2}{\sqrt{\gamma'-1}}\right) \cdot \left(1 - \frac{\mu_2\tau}{4}\right) \leq 1 - \frac{\mu_2\tau}{16}.$$

Consider iteration stage $t$. According to Lemmas A.1 and A.2, we have

$$d^2(\mathbf{Z}^{t+1}, \mathbf{Z}^*) + \frac{1}{\sigma_1}\|\mathbf{S}^{t+1} - \mathbf{S}^*\|_F^2 \leq \left(\rho_1 + \frac{\Gamma_4}{\sigma_1}\right) \cdot d^2(\mathbf{Z}^t, \mathbf{Z}^*) + \frac{1}{\sigma_1}(\rho_2 + \Gamma_1\sigma_1) \cdot \|\mathbf{S}^t - \mathbf{S}^*\|_F^2$$
$$+ \left(-\frac{\eta\mu_1}{4} + \frac{\Gamma_3}{\sigma_1}\right) \cdot \|\mathbf{X}^t - \mathbf{X}^*\|_F^2 + \Gamma_2\|\nabla_\mathbf{X}\mathcal{L}_n(\mathbf{X}^* + \mathbf{S}^*)\|_2^2 + \frac{\Gamma_5}{\sigma_1}\|\nabla_\mathbf{X}\mathcal{L}_n(\mathbf{X}^* + \mathbf{S}^*)\|_{\infty,\infty}^2.$$

Recall the formula of $\Gamma_1$ and $\Gamma_3, \Gamma_4$ from Lemmas A.1 and A.2 respectively. Note that under condition $\eta = c_1/\sigma_1$ and $\beta = c_3/(\alpha r\kappa)$, we can set $c_3$ to be sufficiently small such that

$$\Gamma_4 = C(\gamma, \gamma') \cdot \frac{\tau(\gamma+1)\beta\alpha r\sigma_1}{\mu_2} \leq \frac{c_1\mu_1'\sigma_r}{80},$$

where $\mu_1' = \min\{\mu_1, 2\}$, which implies that $\rho_1 + \Gamma_4/\sigma_1 \leq 1 - \eta\mu_1'\sigma_r/80$. Besides, under condition that $K$ is sufficiently small, we can set $c_1 \leq \min\{\mu_2/50, \tau/96\}$ such that the following inequality holds

$$\Gamma_1\sigma_1 = 48c_1^2(1+K)^2 + c_1\left(\frac{4K^2}{\mu_1} + \mu_2\right) \leq 50c_1^2 + 2c_1\mu_2 \leq 3\mu_2 c_1 \leq \frac{\mu_2\tau}{32}. \quad (A.3)$$

Finally, consider the formula of $\Gamma_3$. Note that similarly we can set $K$ to be small enough such that

$$\Gamma_3 = C(\gamma, \gamma') \cdot \left(\frac{4\tau K^2}{\mu_2} + 3\tau^2(1+K)^2\right) \leq 4\tau^2,$$

thus as long as $\tau$ is sufficiently small, there exist $c_1$ such that $16\tau^2/\mu_1 \leq c_1 \leq \min\{\mu_2/50, \tau/96\}$, which implies $\Gamma_3 \leq c_1\mu_1/4$ while ensuring (A.3) holds as well. Therefore, we obtain

$$d^2(\mathbf{Z}^{t+1}, \mathbf{Z}^*) + \frac{1}{\sigma_1}\|\mathbf{S}^{t+1} - \mathbf{S}^*\|_F^2 \leq \left(1 - \frac{\eta\mu_1'\sigma_r}{80}\right) \cdot d^2(\mathbf{Z}^t, \mathbf{Z}^*) + \frac{1}{\sigma_1}\left(1 - \frac{\mu_2\tau}{32}\right) \cdot \|\mathbf{S}^t - \mathbf{S}^*\|_F^2$$
$$+ \Gamma_2\|\nabla_\mathbf{X}\mathcal{L}_n(\mathbf{X}^* + \mathbf{S}^*)\|_2^2 + \frac{\Gamma_5}{\sigma_1}\|\nabla_\mathbf{S}\mathcal{L}_n(\mathbf{X}^* + \mathbf{S}^*)\|_{\infty,\infty}^2.$$

For simplicity, we denote $D(\mathbf{Z}^t, \mathbf{S}^t) = d^2(\mathbf{Z}^t, \mathbf{Z}^*) + \|\mathbf{S}^t - \mathbf{S}^*\|_F^2/\sigma_1$, and $\rho = \max\{1 - \eta\mu_1'\sigma_r/80, 1 - \mu_2\tau/32\} \in (0, 1)$, then we have

$$D(\mathbf{Z}^{t+1}, \mathbf{S}^{t+1}) \leq \rho D(\mathbf{Z}^t, \mathbf{S}^t) + \Gamma_2\|\nabla_\mathbf{X}\mathcal{L}_n(\mathbf{X}^* + \mathbf{S}^*)\|_2^2 + \frac{\Gamma_5}{\sigma_1}\|\nabla_\mathbf{S}\mathcal{L}_n(\mathbf{X}^* + \mathbf{S}^*)\|_{\infty,\infty}^2.$$

Recall the formula of $\Gamma_2$ and $\Gamma_5$ in Lemmas A.1 and A.2 respectively. Under Condition 4.5, we can always set the sample size $n$ to be large enough such that

$$\Gamma_2\|\nabla_\mathbf{X}\mathcal{L}_n(\mathbf{X}^* + \mathbf{S}^*)\|_2^2 + \frac{\Gamma_5}{\sigma_1}\|\nabla_\mathbf{S}\mathcal{L}_n(\mathbf{X}^* + \mathbf{S}^*)\|_{\infty,\infty}^2 \leq \Gamma_2\epsilon_1^2(n,\delta) + \frac{\Gamma_5}{\sigma_1}\epsilon_2^2(n,\delta) \leq (1-\rho)c_2^2\sigma_r$$

holds with probability at least $1 - \delta$. Thus as long as $D(\mathbf{Z}^0, \mathbf{S}^0) \leq c_2^2\sigma_r$, we have by induction $D(\mathbf{Z}^t, \mathbf{S}^t) \leq c_2^2\sigma_r$ for any $t \geq 0$, which implies $\mathbf{Z}^t \in \mathbb{B}(c_2\sqrt{\sigma_r})$, for any $t \geq 0$. Hence, we obtain

$$D(\mathbf{Z}^t, \mathbf{S}^t) \leq \rho^t D(\mathbf{Z}^0, \mathbf{S}^0) + \frac{\Gamma_2}{1-\rho}\|\nabla_\mathbf{X}\mathcal{L}_n(\mathbf{X}^* + \mathbf{S}^*)\|_2^2 + \frac{\Gamma_5}{(1-\rho)\sigma_1}\|\nabla_\mathbf{S}\mathcal{L}_n(\mathbf{X}^* + \mathbf{S}^*)\|_{\infty,\infty}^2,$$

which completes the proof. □

### A.2 Proof of Theorem 4.8

In order to prove Theorem 4.8, we need to make use of the following lemma. Lemma A.3 characterizes a variation of regularity condition for the sample loss function $\mathcal{L}_n$ with respect to the sparse structure, which is proved in Section B.3.

**Lemma A.3.** Suppose the sample loss function $\mathcal{L}_n$ satisfies Condition 4.3. Given a fixed rank-$r$ matrix $\mathbf{X}$, for any sparse matrices $\mathbf{S}_1, \mathbf{S}_2 \in \mathbb{R}^{d_1 \times d_2}$ with cardinality at most $\gamma' s$, we have

$$\langle \nabla_\mathbf{S}\mathcal{L}_n(\mathbf{X} + \mathbf{S}_1) - \nabla_\mathbf{S}\mathcal{L}_n(\mathbf{X} + \mathbf{S}_2), \mathbf{S}_1 - \mathbf{S}_2 \rangle \geq \frac{\mu_2}{2}\|\mathbf{S}_1 - \mathbf{S}_2\|_F^2$$
$$+ \frac{1}{2L_2}\|\mathcal{P}_\Omega(\nabla_\mathbf{S}\mathcal{L}_n(\mathbf{X} + \mathbf{S}_1) - \nabla_\mathbf{S}\mathcal{L}_n(\mathbf{X} + \mathbf{S}_2))\|_F^2,$$

where $\Omega \subseteq [d_1] \times [d_2]$ is an index set with cardinality at most $\widetilde{s}$ such that $\text{supp}(\mathbf{S}_1) \subseteq \Omega$ and $\mathcal{P}_\Omega$ is the projection operator onto $\Omega$.

*Proof of Theorem 4.8.* Consider a fixed iteration $\ell$ in Algorithm 2. As for the sparse structure, we have

$$\mathbf{S}_{\ell+1} = \mathcal{H}_{\lambda s}(\mathbf{S}_\ell - \tau' \nabla_{\mathbf{S}} \mathcal{L}_n(\mathbf{X}_\ell + \mathbf{S}_\ell)).$$

Denote $\Omega' = \text{supp}(\mathbf{S}^*) \cup \text{supp}(\mathbf{S}_\ell) \cup \text{supp}(\mathbf{S}_{\ell+1})$, then we have $\lambda s \leq |\Omega'| \leq (2\lambda + 1)s$. We further denote $\widetilde{\mathbf{S}}_{\ell+1} = \mathcal{P}_{\Omega'}(\mathbf{S}_\ell - \tau' \nabla_{\mathbf{S}} \mathcal{L}_n(\mathbf{X}_\ell + \mathbf{S}_\ell))$, then we obtain $\mathbf{S}_{\ell+1} = \mathcal{H}_{\lambda s}(\widetilde{\mathbf{S}}_{\ell+1})$. Thus, according to Lemma 3.3 in Li et al. (2016), we have

$$\|\mathbf{S}_{\ell+1} - \mathbf{S}^*\|_F^2 \leq \left(1 + \frac{2}{\sqrt{\lambda' - 1}}\right) \cdot \|\widetilde{\mathbf{S}}_{\ell+1} - \mathbf{S}^*\|_F^2. \tag{A.4}$$

Therefore, it is sufficient to upper bound $\|\widetilde{\mathbf{S}}_{\ell+1} - \mathbf{S}^*\|_F$ for the sparse structure. We have

$$\|\widetilde{\mathbf{S}}_{\ell+1} - \mathbf{S}^*\|_F = \|\mathbf{S}_\ell - \mathbf{S}^* - \tau' \mathcal{P}_{\Omega'}(\nabla_{\mathbf{S}} \mathcal{L}_n(\mathbf{X}_\ell + \mathbf{S}_\ell))\|_F$$
$$\leq \underbrace{\|\mathbf{S}_\ell - \mathbf{S}^* - \tau' \mathcal{P}_{\Omega'}(\nabla_{\mathbf{S}} \mathcal{L}_n(\mathbf{X}_\ell + \mathbf{S}_\ell) - \nabla_{\mathbf{S}} \mathcal{L}_n(\mathbf{X}_\ell + \mathbf{S}^*))\|_F}_{I_1}$$
$$+ \tau' \underbrace{\|\mathcal{P}_{\Omega'}(\nabla_{\mathbf{S}} \mathcal{L}_n(\mathbf{X}_\ell + \mathbf{S}^*) - \nabla_{\mathbf{S}} \mathcal{L}_n(\mathbf{X}^* + \mathbf{S}^*))\|_F}_{I_2} + \tau' \underbrace{\|\mathcal{P}_{\Omega'}(\nabla_{\mathbf{S}} \mathcal{L}_n(\mathbf{X}^* + \mathbf{S}^*))\|_F}_{I_3}, \tag{A.5}$$

where the second inequality follows from the triangle inequality. As for the first term $I_1$ in (A.5), according to Lemma A.3, we have

$$I_1^2 \leq (1 - \mu_2 \tau') \cdot \|\mathbf{S}_\ell - \mathbf{S}^*\|_F^2 - \left(\frac{\tau'}{L_2} - \tau'^2\right) \cdot \|\mathcal{P}_{\Omega'}(\nabla_{\mathbf{S}} \mathcal{L}_n(\mathbf{X}_\ell + \mathbf{S}_\ell) - \nabla_{\mathbf{S}} \mathcal{L}_n(\mathbf{X}_\ell + \mathbf{S}^*))\|_F^2$$
$$\leq (1 - \mu_2 \tau') \cdot \|\mathbf{S}_\ell - \mathbf{S}^*\|_F^2, \tag{A.6}$$

provided that $\tau' \leq 1/L_2$. Consider the second term $I_2$ in (A.5). Note that $|\Omega'| \leq (2\lambda + 1)s$, thus according to the definition of Frobenius norm, we have

$$I_2 = \sup_{\|\mathbf{W}\|_F \leq 1} \langle \mathcal{P}_{\Omega'}(\nabla_{\mathbf{S}} \mathcal{L}_n(\mathbf{X}_\ell + \mathbf{S}^*) - \nabla_{\mathbf{S}} \mathcal{L}_n(\mathbf{X}^* + \mathbf{S}^*)), \mathbf{W} \rangle$$
$$\leq \sup_{\|\mathbf{W}\|_F \leq 1} \{|\langle \mathbf{X}_\ell - \mathbf{X}^*, \mathcal{P}_{\Omega'}(\mathbf{W}) \rangle| + K\|\mathbf{X}_\ell - \mathbf{X}^*\|_F \cdot \|\mathcal{P}_{\Omega'}(\mathbf{W})\|_F\}$$
$$\leq \|\mathbf{X}_\ell - \mathbf{X}^*\|_{\infty,\infty} \cdot \|\mathcal{P}_{\Omega'}(\mathbf{W})\|_{1,1} + K\|\mathbf{X}_\ell - \mathbf{X}^*\|_F \leq 4\zeta^* \sqrt{\lambda s} + K\|\mathbf{X}_\ell - \mathbf{X}^*\|_F, \tag{A.7}$$

where the first inequality follows from Condition 4.4, the second inequality holds because $|\langle \mathbf{A}, \mathbf{B} \rangle| \leq \|\mathbf{A}\|_{1,1} \cdot \|\mathbf{B}\|_{\infty,\infty}$ and $\|\mathcal{P}_{\Omega'}(\mathbf{W})\|_F \leq \|\mathbf{W}\|_F \leq 1$, and the last inequality is due to the fact that $\|\mathbf{X}_\ell\|_{\infty,\infty} \leq \zeta^*, \|\mathbf{X}^*\|_{\infty,\infty} \leq \zeta^*$ and the triangle inequality. And for the third term $I_3$, we have

$$I_3 \leq \sqrt{(2\lambda + 1)s} \cdot \|\nabla_{\mathbf{S}} \mathcal{L}_n(\mathbf{X}^* + \mathbf{S}^*)\|_{\infty,\infty}. \tag{A.8}$$

Therefore, plugging (A.6), (A.7) and (A.8) into (A.5), we obtain

$$\|\widetilde{\mathbf{S}}_{\ell+1} - \mathbf{S}^*\|_F \leq \sqrt{1-\mu_2\tau'} \cdot \|\mathbf{S}_\ell - \mathbf{S}^*\|_F + \tau'K\|\mathbf{X}_\ell - \mathbf{X}^*\|_F + 2\tau'\zeta^*\sqrt{s} \\ + \tau'\sqrt{(2\lambda+1)s} \cdot \|\nabla_{\mathbf{S}}\mathcal{L}_n(\mathbf{X}^* + \mathbf{S}^*)\|_{\infty,\infty}. \quad (A.9)$$

Hence, combining (A.4) and (A.9), we obtain the following result for sparse structure

$$\|\mathbf{S}_{\ell+1} - \mathbf{S}^*\|_F \leq \left(1 + \frac{2}{\sqrt{\lambda-1}}\right) \cdot \left(\sqrt{1-\mu_2\tau'} \cdot \|\mathbf{S}_\ell - \mathbf{S}^*\|_F + \tau'K\|\mathbf{X}_\ell - \mathbf{X}^*\|_F\right) \\ + \tau'\left(1 + \frac{2}{\sqrt{\lambda-1}}\right) \cdot (4\zeta^*\sqrt{\lambda s} + \sqrt{3\lambda s} \cdot \|\nabla_{\mathbf{S}}\mathcal{L}_n(\mathbf{X}^* + \mathbf{S}^*)\|_{\infty,\infty}). \quad (A.10)$$

Next, let us consider the low-rank structure. According to Algorithm 2, we have

$$\mathbf{X}_{\ell+1} = \mathcal{P}_{\lambda',\zeta^*}(\mathbf{X}_\ell - \eta'\nabla_{\mathbf{X}}\mathcal{L}_n(\mathbf{X}_\ell + \mathbf{S}_\ell)),$$

where the projection operator $\mathcal{P}_{\lambda',\zeta^*}$ is defined as

$$\mathcal{P}_{\lambda',\zeta^*}(\mathbf{X}) = \underset{\text{rank}(\mathbf{Y})\leq\lambda'r, \|\mathbf{Y}\|_{\infty,\infty}\leq\zeta^*}{\arg\min} \|\mathbf{Y} - \mathbf{X}\|_F, \text{ for any } \mathbf{X} \in \mathbb{R}^{d_1 \times d_2}.$$

Let the singular value decomposition of $\mathbf{X}_\ell, \mathbf{X}_{\ell+1}$ be $\mathbf{X}_\ell = \overline{\mathbf{U}}^\ell \mathbf{\Sigma}^\ell \overline{\mathbf{V}}^{\ell\top}$ and $\mathbf{X}_{\ell+1} = \overline{\mathbf{U}}^{\ell+1} \mathbf{\Sigma}^{\ell+1} \overline{\mathbf{V}}^{\ell+1\top}$ respectively. Define the following subspace spanned by the column vectors of $\overline{\mathbf{U}}^*, \overline{\mathbf{U}}^\ell$ and $\overline{\mathbf{U}}^{\ell+1}$ as

$$\text{span}(\widetilde{\mathbf{U}}) = \text{span}\{\overline{\mathbf{U}}^*, \overline{\mathbf{U}}^\ell, \overline{\mathbf{U}}^{\ell+1}\} = \text{col}(\overline{\mathbf{U}}^*) + \text{col}(\overline{\mathbf{U}}^\ell) + \text{col}(\overline{\mathbf{U}}^{\ell+1}),$$

where each column vector of $\widetilde{\mathbf{U}}$ is a basis vector of the above subspace. Similarly, we define the subspace spanned by the column vectors of $\overline{\mathbf{V}}^*, \overline{\mathbf{V}}^\ell$ and $\overline{\mathbf{V}}^{\ell+1}$ as

$$\text{span}(\widetilde{\mathbf{V}}) = \text{span}\{\overline{\mathbf{V}}^*, \overline{\mathbf{V}}^\ell, \overline{\mathbf{V}}^{\ell+1}\} = \text{col}(\overline{\mathbf{V}}^*) + \text{col}(\overline{\mathbf{V}}^\ell) + \text{col}(\overline{\mathbf{V}}^{\ell+1}),$$

Note that $\mathbf{X}^*$ has rank $r$, $\mathbf{X}_\ell$ and $\mathbf{X}_{\ell+1}$ has rank at most $\lambda'r$, thus both $\widetilde{\mathbf{U}}$ and $\widetilde{\mathbf{V}}$ have at most $(2\lambda'+1)r$ columns. Moreover, we further define the following subspace

$$\mathcal{A} = \{\mathbf{\Delta} \in \mathbb{R}^{d_1 \times d_2} \mid \text{row}(\mathbf{\Delta}) \subseteq \text{span}(\widetilde{\mathbf{V}}) \text{ and } \text{col}(\mathbf{\Delta}) \subseteq \text{span}(\widetilde{\mathbf{U}})\}.$$

Let $\Pi_\mathcal{A}$ be the projection operator onto $\mathcal{A}$, then for any $\mathbf{X} \in \mathbb{R}^{d_1 \times d_2}$, we have $\Pi_\mathcal{A}(\mathbf{X}) = \widetilde{\mathbf{U}}\widetilde{\mathbf{U}}^\top \mathbf{X} \widetilde{\mathbf{V}}\widetilde{\mathbf{V}}^\top$. Note that for any $\mathbf{X} \in \mathbb{R}^{d_1 \times d_2}$, we have $\text{rank}(\Pi_\mathcal{A}(\mathbf{X})) \leq (2\lambda'+1)r$, since $\text{rank}(\mathbf{AB}) \leq \min\{\text{rank}(\mathbf{A}), \text{rank}(\mathbf{B})\}$. Besides, we denote

$$\widetilde{\mathbf{X}}_{\ell+1} = \mathbf{X}_\ell - \eta'\Pi_\mathcal{A}(\nabla_{\mathbf{X}}\mathcal{L}_n(\mathbf{X}_\ell + \mathbf{S}_\ell)).$$

Similar to the proof of Theorem 5.9 in Wang et al. (2016), we have $\mathbf{X}_{\ell+1}$ is actually the best rank-$\lambda'r$ approximation of $\widetilde{\mathbf{X}}_{\ell+1}$ satisfying the infinity norm constraint, or in other words, $\mathbf{X}_{\ell+1} = \mathcal{P}_{\lambda',\zeta^*}(\widetilde{\mathbf{X}}_{\ell+1})$. Note that $\mathcal{P}_{\lambda',\zeta^*}(\mathbf{X}^*) = \mathbf{X}^*$, thus according to Lemma 3.18 in Li et al. (2016), we obtain

$$\|\mathbf{X}_{\ell+1} - \mathbf{X}^*\|_F^2 = \|\mathcal{P}_{\lambda',\zeta^*}(\widetilde{\mathbf{X}}_{\ell+1}) - \mathbf{X}^*\|_F^2 \leq \left(1 + \frac{2}{\sqrt{\lambda'-1}}\right) \cdot \|\widetilde{\mathbf{X}}_{\ell+1} - \mathbf{X}^*\|_F^2. \quad (A.11)$$

Thus, it suffices to bound the term $\|\widetilde{\mathbf{X}}_{\ell+1} - \mathbf{X}^*\|_F$. Note that $\mathbf{X}^* \in \mathcal{A}$, thus according to the triangle inequality, we have

$$\|\widetilde{\mathbf{X}}_{\ell+1} - \mathbf{X}^*\|_F \leq \underbrace{\|\mathbf{X}_\ell - \mathbf{X}^* - \eta'\Pi_\mathcal{A}(\nabla_\mathbf{X}\mathcal{L}_n(\mathbf{X}_\ell + \mathbf{S}_\ell) - \nabla_\mathbf{X}\mathcal{L}_n(\mathbf{X}^* + \mathbf{S}_\ell))\|_F}_{I_1'}$$
$$+ \eta'\underbrace{\|\Pi_\mathcal{A}(\nabla_\mathbf{X}\mathcal{L}_n(\mathbf{X}^* + \mathbf{S}_\ell) - \nabla_\mathbf{X}\mathcal{L}_n(\mathbf{X}^* + \mathbf{S}^*))\|_F}_{I_2'} + \eta'\underbrace{\|\Pi_\mathcal{A}(\nabla_\mathbf{X}\mathcal{L}_n(\mathbf{X}^* + \mathbf{S}^*))\|_F}_{I_3'}. \tag{A.12}$$

Consider $I_1'$ in (A.12) first. According to Lemma B.2 in Wang et al. (2017), we have

$$I_1'^2 \leq (1 - \eta'\mu_1) \cdot \|\mathbf{X}_\ell - \mathbf{X}^*\|_F^2, \tag{A.13}$$

provided that $\eta' \leq 1/L_1$. As for the second term $I_2'$ in (A.12), by the definition of Frobenius norm, we have

$$I_2' = \sup_{\|\mathbf{W}\|_F \leq 1} \langle \Pi_{\mathcal{A}_{3r}}(\nabla_\mathbf{X}\mathcal{L}_n(\mathbf{X}^* + \mathbf{S}_\ell) - \nabla_\mathbf{X}\mathcal{L}_n(\mathbf{X}^* + \mathbf{S}^*)), \mathbf{W}\rangle$$
$$\leq (1 + K) \cdot \|\mathbf{S}_\ell - \mathbf{S}^*\|_F \cdot \|\Pi_\mathcal{A}(\mathbf{W})\|_F \leq (1 + K) \cdot \|\mathbf{S}_\ell - \mathbf{S}^*\|_F, \tag{A.14}$$

where the first inequality holds because of Condition 4.4. As for $I_3'$, we have

$$I_3' \leq \sqrt{(2\lambda' + 1)r} \cdot \|\Pi_\mathcal{A}(\nabla_\mathbf{X}\mathcal{L}_n(\mathbf{X}^* + \mathbf{S}^*))\|_2 \leq \sqrt{(2\lambda' + 1)r} \cdot \|\nabla_\mathbf{X}\mathcal{L}_n(\mathbf{X}^* + \mathbf{S}^*)\|_2. \tag{A.15}$$

Therefore, plugging (A.13), (A.14) and (A.15) into (A.12), we obtain

$$\|\widetilde{\mathbf{X}}_{\ell+1} - \mathbf{X}^*\|_F \leq \sqrt{1 - \eta'\mu_1} \cdot \|\mathbf{X}_\ell - \mathbf{X}^*\|_F + \eta'(1 + K) \cdot \|\mathbf{S}_\ell - \mathbf{S}^*\|_F + \eta'\sqrt{3\lambda'r}\|\nabla_\mathbf{X}\mathcal{L}_n(\mathbf{X}^* + \mathbf{S}^*)\|_2. \tag{A.16}$$

Finally, combining (A.11) and (A.16), we obtain the following result for low rank structure

$$\|\mathbf{X}_{\ell+1} - \mathbf{X}^*\|_F \leq \left(1 + \frac{2}{\sqrt{\lambda' - 1}}\right) \cdot \left(\sqrt{1 - \eta'\mu_1} \cdot \|\mathbf{X}_\ell - \mathbf{X}^*\|_F + \eta'(1 + K) \cdot \|\mathbf{S}_\ell - \mathbf{S}^*\|_F\right)$$
$$+ \eta'\left(1 + \frac{2}{\sqrt{\lambda' - 1}}\right) \cdot \sqrt{3\lambda'r}\|\nabla_\mathbf{X}\mathcal{L}_n(\mathbf{X}^* + \mathbf{S}^*)\|_2. \tag{A.17}$$

Hence, combining (A.10) and (A.17), we obtain

$$\|\mathbf{X}_{\ell+1} - \mathbf{X}^*\|_F + \|\mathbf{S}_{\ell+1} - \mathbf{S}^*\|_F \leq \rho_1'\|\mathbf{X}_\ell - \mathbf{X}^*\|_F + \rho_2'\|\mathbf{S}_\ell - \mathbf{S}^*\|_F + 4\sqrt{\lambda}\tau'\left(1 + \frac{2}{\sqrt{\lambda - 1}}\right) \cdot \zeta^*\sqrt{s}$$
$$+ \Gamma_1\|\nabla_\mathbf{X}\mathcal{L}_n(\mathbf{X}^* + \mathbf{S}^*)\|_2 + \Gamma_2\|\nabla_\mathbf{S}\mathcal{L}_n(\mathbf{X}^* + \mathbf{S}^*)\|_{\infty,\infty}, \tag{A.18}$$

where $\Gamma_1 = \eta'(1 + 2/\sqrt{\lambda' - 1})\sqrt{3\lambda'r}$, $\Gamma_2 = \tau'(1 + 2/\sqrt{\lambda - 1})\sqrt{3\lambda s}$, and contraction parameter $\rho_1', \rho_2'$ are defined in Theorem 4.8. Note that we set $\eta' = 1/(6\mu_1) \leq 1/L_1$, $\tau' = 3/(4\mu_2) \leq 1/L_2$, and we assume $\mu_1 \geq 1/3$. Then with sufficient large $\lambda$ and $\lambda'$ and structural Lipschitz gradient parameter $K$ small enough, we could guarantee $\rho_1', \rho_2' \in (0, 19/20)$. Plugging in the definition of $\zeta^* = c_0\alpha r\kappa/\sqrt{d_1 d_2}$, we complete the proof by induction. □

## A.3 Proof of Theorem 4.9

*Proof.* To prove Theorem 4.9, it is sufficient to verify the assumption $D(\mathbf{Z}^0, \mathbf{S}^0) \leq c_4^2 \sigma_r$ in Theorem 4.6. Thus, according to Theorem 4.8, it is sufficient to make sure the right hand side of (4.2) is small enough.

As for the optimization error, i.e., the first term on the R.H.S. of (4.2), we can perform $L \geq \log\{c\sigma_r/(2\|\mathbf{X}^*\|_F + 2\|\mathbf{S}^*\|_F\}/\log(\rho')$ iterations in Algorithm 2 to make sure the optimization error is sufficiently small such that $\rho'^L \cdot (\|\mathbf{X}^*\|_F + \|\mathbf{S}^*\|_F) \leq c\sigma_r/2$, where $c = \min\{1/2, c_4/4\}$.

On the other hand, for the statistical error, i.e., the last three terms on the R.H.S. of (4.2), we assume $s \leq cd_1 d_2/(\alpha^2 r^2 \kappa^2)$, where $c$ is a small enough constant, and sample size $n$ is sufficiently large such that $\Gamma_1 \sqrt{r}\epsilon_1(n,\delta) + \Gamma_2 \sqrt{s}\epsilon_2(n,\delta) \leq c\sigma_r/4$. Putting pieces together, we arrive at $\|\mathbf{X}^0 - \mathbf{X}^*\|_F + \|\mathbf{S}^0 - \mathbf{S}^*\|_F \leq c \cdot \sigma_r$. Finally, based on Lemma 5.14 in Tu et al. (2015), the initial assumption that $D(\mathbf{Z}^0, \mathbf{S}^0) \leq c_4^2 \sigma_r$ in Theorem 4.6 is satisfied, which completes the proof. $\square$

# B  Proofs of Technical Lemmas in Appendix A

## B.1  Proof of Lemma A.1

In order to prove Theorem A.1, we need to make use of the following lemmas. Lemma B.1 characterizes a local curvature property of the low-rank structure, which gives us the lower bound of the inner product term. We provide its proof in Section C.1. Lemma B.2, proved in Section C.2, characterizes a local smoothness property of the low-rank structure and gives us an upper bound of the Frobenius term.

**Lemma B.1** (Local Curvature Property for Low-Rank Structure). Suppose the sample loss function $\mathcal{L}_n$ satisfies Conditions 4.2 and 4.4. Recall that $\mathbf{X}^* = \mathbf{U}^* \mathbf{V}^{*\top}$ is the unknown rank-$r$ matrix that satisfies (3.1), and $\mathbf{S}^*$ is the unknown $s$-sparse matrix. Let $\mathbf{Z} \in \mathbb{R}^{(d_1+d_2) \times r}$ be any matrix with $\mathbf{Z} = [\mathbf{U}; \mathbf{V}]$, where $\mathbf{U} \in \mathbb{R}^{d_1 \times r}$, $\mathbf{V} \in \mathbb{R}^{d_2 \times r}$ satisfy $\|\mathbf{U}\|_{2,\infty} \leq 2\sqrt{\alpha r \sigma_1/d_1}$ and $\|\mathbf{V}\|_{2,\infty} \leq 2\sqrt{\alpha r \sigma_1/d_2}$. Let $\mathbf{S} \in \mathbb{R}^{d_1 \times d_2}$ be any matrix with at most $\beta'$-fraction nonzero entries per row and column and satisfying $\|\mathbf{S}\|_0 \leq s' \leq \widetilde{s}$. Denote the optimal rotation with respect to $\mathbf{Z}$ by $\mathbf{R} = \mathrm{argmin}_{\widetilde{\mathbf{R}} \in \mathbb{Q}_r} \|\mathbf{Z} - \mathbf{Z}^* \widetilde{\mathbf{R}}\|_F$, and $\mathbf{H} = \mathbf{Z} - \mathbf{Z}^* \mathbf{R}$, then we have

$$\langle \nabla_\mathbf{Z} \widetilde{F}_n(\mathbf{Z}, \mathbf{S}), \mathbf{H} \rangle \geq \frac{\mu_1}{4}\|\mathbf{X} - \mathbf{X}^*\|_F^2 + \frac{1}{16}\|\widetilde{\mathbf{Z}}^\top \mathbf{Z}\|_F^2 + \left(\frac{\mu_1'}{20}\sigma_r - C\right) \cdot \|\mathbf{H}\|_F^2 - \left(\frac{L_1+1}{8} + \frac{1}{\mu_2}\right) \cdot \|\mathbf{H}\|_F^4$$
$$- \left(\frac{\mu_2}{2} + \frac{2K^2}{\mu_1}\right) \cdot \|\mathbf{S} - \mathbf{S}^*\|_F^2 - \left(\frac{8r}{\mu_1} + \frac{r}{L_1}\right) \cdot \|\nabla_\mathbf{X} \mathcal{L}_n(\mathbf{X}^* + \mathbf{S}^*)\|_2^2,$$

where $\mathbf{X} = \mathbf{U}\mathbf{V}^\top$, $\mu_1' = \min\{\mu_1, 2\}$, and $C = 18(\beta' + \beta)\alpha r \sigma_1/\mu_2$.

**Lemma B.2** (Local Smoothness Property for Low-Rank Structure). Suppose the sample loss function $\mathcal{L}_n$ satisfies Conditions 4.2 and 4.4. Recall that $\mathbf{X}^*$ is the unknown rank-$r$ matrix and $\mathbf{S}^*$ is the unknown $s$-sparse matrix. For any matrix $\mathbf{Z} = [\mathbf{U}; \mathbf{V}] \in \mathbb{R}^{(d_1+d_2) \times r}$ and $\mathbf{S} \in \mathbb{R}^{d_1 \times d_2}$ with at most $s'$ nonzero entries satisfying $s' \leq \widetilde{s}$, we have

$$\|\nabla_\mathbf{Z} \widetilde{F}_n(\mathbf{Z}, \mathbf{S})\|_F^2 \leq \left(12L_1^2 \|\mathbf{X} - \mathbf{X}^*\|_F^2 + 12(1+K)^2 \cdot \|\mathbf{S} - \mathbf{S}^*\|_F^2 + \|\mathbf{U}^\top \mathbf{U} - \mathbf{V}^\top \mathbf{V}\|_F^2\right) \cdot \|\mathbf{Z}\|_2^2$$
$$+ 12r\|\nabla_\mathbf{X}\mathcal{L}_n(\mathbf{X}^* + \mathbf{S}^*)\|_2^2 \cdot \|\mathbf{Z}\|_2^2,$$

where $\mathbf{X} = \mathbf{U}\mathbf{V}^\top$.

*Proof of Lemma A.1.* Recall $\mathbf{Z}^* = [\mathbf{U}^*; \mathbf{V}^*]$ and $\mathbf{X}^* = \mathbf{U}^*\mathbf{V}^{*\top}$, where $\mathbf{U}^* = \overline{\mathbf{U}}^*(\mathbf{\Sigma}^*)^{1/2}$, $\mathbf{V}^* = \overline{\mathbf{V}}^*(\mathbf{\Sigma}^*)^{1/2}$, we have $\|\mathbf{Z}^*\|_2 = \sqrt{2\sigma_1}$. According to our initial ball assumption $\mathbf{Z}^0 \in \mathbb{B}(\sqrt{\sigma_r}/4)$, there exists an orthogonal matrix $\mathbf{R} \in \mathbb{R}^{r \times r}$ such that $\|\mathbf{Z}^0 - \mathbf{Z}^*\mathbf{R}\|_F \leq \sqrt{\sigma_r}/4$, thus we obtain

$$\sqrt{\sigma_1} \leq \|\mathbf{Z}^*\|_2 - \|\mathbf{Z}^0 - \mathbf{Z}^*\mathbf{R}\|_2 \leq \|\mathbf{Z}^0\|_2 \leq \|\mathbf{Z}^*\|_2 + \|\mathbf{Z}^0 - \mathbf{Z}^*\mathbf{R}\|_F \leq 2\sqrt{\sigma_1}.$$

Recall (3.1) and the definition of $\mathcal{C}_1, \mathcal{C}_2$ in (3.2), then it is obvious that $\mathbf{U}^* \in \mathcal{C}_1$ and $\mathbf{V}^* \in \mathcal{C}_2$. Consider a fixed iteration stage $t$, we denote

$$\widetilde{\mathbf{U}}^{t+1} = \mathbf{U}^t - \eta\nabla_\mathbf{U}\mathcal{L}_n(\mathbf{U}^t\mathbf{V}^{t\top} + \mathbf{S}^t) - \frac{1}{2}\eta\mathbf{U}^t(\mathbf{U}^{t\top}\mathbf{U}^t - \mathbf{V}^{t\top}\mathbf{V}^t),$$

$$\widetilde{\mathbf{V}}^{t+1} = \mathbf{V}^t - \eta\nabla_\mathbf{V}\mathcal{L}_n(\mathbf{U}^t\mathbf{V}^{t\top} + \mathbf{S}^t) - \frac{1}{2}\eta\mathbf{V}^t(\mathbf{V}^{t\top}\mathbf{V}^t - \mathbf{U}^{t\top}\mathbf{U}^t).$$

Denote $\widetilde{\mathbf{Z}}^{t+1} = [\widetilde{\mathbf{U}}^{t+1}; \widetilde{\mathbf{V}}^{t+1}]$, and $\mathbf{Z}^t = [\mathbf{U}^t; \mathbf{V}^t]$, for any iteration stage $t$, then according to (A.2), we have $\widetilde{\mathbf{Z}}^{t+1} = \mathbf{Z}^t - \eta\nabla_\mathbf{Z}\widetilde{F}_n(\mathbf{Z}^t, \mathbf{S}^t)$. Besides, according to Algorithm 1, we obtain

$$\mathbf{U}^{t+1} = \mathcal{P}_{\mathcal{C}_1}(\widetilde{\mathbf{U}}^{t+1}) \quad \text{and} \quad \mathbf{V}^{t+1} = \mathcal{P}_{\mathcal{C}_2}(\widetilde{\mathbf{V}}^{t+1}).$$

Recall $\mathbf{Z}^* = [\mathbf{U}^*; \mathbf{V}^*]$, and $\mathbf{R}^t = \arg\min_{\mathbf{R} \in \mathbb{Q}_r} \|\mathbf{Z}^t - \mathbf{Z}^*\mathbf{R}\|_F$, for any $t$. Denote $\mathbf{H}^t = \mathbf{Z} - \mathbf{Z}^*\mathbf{R}^t$. Since $\mathcal{C}_1, \mathcal{C}_2$ are both rotation-invariant constraint sets, and $\mathbf{U}^* \in \mathcal{C}_1$, $\mathbf{V}^* \in \mathcal{C}_2$, we have

$$\begin{aligned} d^2(\mathbf{Z}^{t+1}, \mathbf{Z}^*) &\leq \|\mathbf{Z}^{t+1} - \mathbf{Z}^*\mathbf{R}^t\|_F^2 \\ &\leq \|\mathbf{Z}^t - \eta\nabla_\mathbf{Z}\widetilde{F}_n(\mathbf{Z}^t, \mathbf{S}^t) - \mathbf{Z}^*\mathbf{R}^t\|_F^2 \\ &= d^2(\mathbf{Z}^t, \mathbf{Z}^*) - 2\eta\langle\nabla_\mathbf{Z}\widetilde{F}_n(\mathbf{Z}^t, \mathbf{S}^t), \mathbf{H}^t\rangle + \eta^2\|\nabla_\mathbf{Z}\widetilde{F}_n(\mathbf{Z}^t, \mathbf{S}^t)\|_F^2, \end{aligned} \quad (B.1)$$

where the first inequality follows from Definition 4.1, and the second inequality is due to the nonexpansive property of projection $\mathcal{P}_{\mathcal{C}_i}$ onto $\mathcal{C}_i$, where $i \in \{1, 2\}$. Therefore, it suffices to lower bound the inner product term $\langle\nabla_\mathbf{Z}\widetilde{F}_n(\mathbf{Z}^t, \mathbf{S}^t), \mathbf{H}^t\rangle$ and upper bound the term $\|\nabla_\mathbf{Z}\widetilde{F}_n(\mathbf{Z}^t, \mathbf{S}^t)\|_F^2$, respectively. According to Algorithm 1, we have $(\mathbf{U}^t, \mathbf{V}^t)$ satisfies the condition of $(\mathbf{U}, \mathbf{V})$ in Lemma B.1, and $\mathbf{S}^t$ has at most $\gamma\beta$-fraction nonzero entries per row and column with $\|\mathbf{S}\|_0 \leq \gamma's \leq \widetilde{s}$. Denote $\mathbf{X}^t = \mathbf{U}^t\mathbf{V}^{t\top}$, then according to Lemma B.1, we obtain

$$\begin{aligned} \langle\nabla_\mathbf{Z}\widetilde{F}_n(\mathbf{Z}^t, \mathbf{S}^t), \mathbf{H}^t\rangle &\geq \frac{\mu_1}{4}\|\mathbf{X}^t - \mathbf{X}^*\|_F^2 + \frac{1}{16}\|\mathbf{U}^{t\top}\mathbf{U}^t - \mathbf{V}^{t\top}\mathbf{V}^t\|_F^2 + \left(\frac{\mu_1'}{20}\sigma_r - C\right) \cdot \|\mathbf{H}^t\|_F^2 \\ &- \left(\frac{L_1+1}{8} + \frac{1}{\mu_2}\right) \cdot \|\mathbf{H}^t\|_F^4 - \left(\frac{\mu_2}{2} + \frac{2K^2}{\mu_1}\right) \cdot \|\mathbf{S}^t - \mathbf{S}^*\|_F^2 - \left(\frac{8r}{\mu_1} + \frac{r}{L_1}\right) \cdot \|\nabla_\mathbf{X}\mathcal{L}_n(\mathbf{X}^* + \mathbf{S}^*)\|_2^2, \end{aligned}$$

where $\mu_1' = \min\{\mu_1, 2\}$, and $C = 18(\gamma+1)\beta\alpha r\sigma_1/\mu_2$. Besides, according to Lemma B.2, we have

$$\begin{aligned} \|\nabla_\mathbf{Z}\widetilde{F}_n(\mathbf{Z}^t, \mathbf{S}^t)\|_F^2 &\leq \left(12L_1^2\|\mathbf{X}^t - \mathbf{X}^*\|_F^2 + 12(1+K)^2 \cdot \|\mathbf{S}^t - \mathbf{S}^*\|_F^2 + \|\mathbf{U}^{t\top}\mathbf{U}^t - \mathbf{V}^{t\top}\mathbf{V}^t\|_F^2\right) \cdot \|\mathbf{Z}^t\|_2^2 \\ &+ 12r\|\nabla_\mathbf{X}\mathcal{L}_n(\mathbf{X}^* + \mathbf{S}^*)\|_2^2 \cdot \|\mathbf{Z}^t\|_2^2. \end{aligned}$$

Note that under the assumption of $\mathbf{Z}^t \in \mathbb{B}(c_2\sqrt{\sigma_r})$, where $c_2 \leq 1/4$, we have $\|\mathbf{Z}^t\|_2 \leq \|\mathbf{Z}^*\mathbf{R}^t\|_2 + \|\mathbf{Z}^t - \mathbf{Z}^*\mathbf{R}^t\|_2 \leq 2\sqrt{\sigma_1}$, since $\|\mathbf{Z}^*\|_2^2 = 2\sigma_1$. Thus, if we set the step size $\eta = c_1/\sigma_1$, where

$c_1 \leq \min\{1/32, \mu_1/(192L_1^2)\}$, and we assume $\beta \leq 1/(c_3 \alpha r \kappa)$ with $c_3$ large enough such that $c_3 \geq 720(\gamma+1)\mu_2/\mu_1'$, we have

$$-2\eta \langle \nabla_{\mathbf{Z}} \widetilde{F}_n(\mathbf{Z}^t, \mathbf{S}^t), \mathbf{H}^t \rangle + \eta^2 \|\nabla_{\mathbf{Z}} \widetilde{F}_n(\mathbf{Z}^t, \mathbf{S}^t)\|_F^2 \leq -\frac{\eta \mu_1}{4}\|\mathbf{X}^t - \mathbf{X}^*\|_F^2 - \frac{\eta \mu_1' \sigma_r}{20}\|\mathbf{H}^t\|_F^2$$
$$+ \eta\left(\frac{L_1+1}{4} + \frac{2}{\mu_2}\right) \cdot \|\mathbf{H}^t\|_F^4 + C_1\|\mathbf{S}^t - \mathbf{S}^*\|_F^2 + C_2\|\nabla_{\mathbf{X}}\mathcal{L}_n(\mathbf{X}^* + \mathbf{S}^*)\|_2^2,$$

where $C_1 = 48\eta^2(1+K)^2\sigma_1 + \eta(\mu_2 + 4K^2/\mu_1)$, and $C_2 = 48\eta^2 r\sigma_1 + 2\eta(8r/\mu_1 + r/L_1)$. Note that according to our assumption, $\|\mathbf{H}^t\|_F^2 \leq c_2^2 \sigma_r$ with $c_2^2 \leq \mu_1'/[10(L_1+1+8/\mu_2)]$, thus by (B.1), we obtain

$$d^2(\mathbf{Z}^{t+1}, \mathbf{Z}^*) \leq \left(1 - \frac{\eta \mu_1' \sigma_r}{40}\right) d^2(\mathbf{Z}^t, \mathbf{Z}^*) - \frac{\eta \mu_1}{4}\|\mathbf{X}^t - \mathbf{X}^*\|_F^2 + C_1\|\mathbf{S}^t - \mathbf{S}^*\|_F^2 + C_2\|\nabla_{\mathbf{X}}\mathcal{L}_n(\mathbf{X}^* + \mathbf{S}^*)\|_2^2,$$

which completes the proof. $\square$

## B.2 Proof of Lemma A.2

In order to prove Lemma A.2, we need to utilize the following lemma. Inspired by Yi et al. (2016), we present Lemma B.3, which characterizes a nearly non-expansiveness property of the truncation operator $\mathcal{T}_\theta$, as long as $\theta$ is large enough. We provides its proof in Section C.3 for completeness.

**Lemma B.3.** Suppose $\mathbf{S}^* \in \mathbb{R}^{d_1 \times d_2}$ is the unknown sparse matrix with at most $\beta$-fraction nonzero entries per row and column. For any matrix $\mathbf{S} \in \mathbb{R}^{d_1 \times d_2}$, we have

$$\|\mathcal{T}_{\gamma\beta}(\mathbf{S}) - \mathbf{S}^*\|_F^2 \leq \left(1 + \sqrt{\frac{2}{\gamma-1}}\right)^2 \cdot \|\mathbf{S} - \mathbf{S}^*\|_F^2,$$

where $\gamma > 1$ is a parameter.

Now we are ready to prove Lemma A.2.

*Proof of Lemma A.2.* Consider a fixed iteration stage $t$. For the sparse structure, according to Algorithm 1, we have

$$\mathbf{S}^{t+1} = \mathcal{T}_{\gamma\beta} \circ \mathcal{H}_{\gamma' s}\big(\mathbf{S}^t - \tau \nabla_{\mathbf{S}} \mathcal{L}_n(\mathbf{U}^t \mathbf{V}^{t\top} + \mathbf{S}^t)\big).$$

Denote $\bar{\mathbf{S}}^{t+1} = \mathcal{H}_{\gamma' s}\big(\mathbf{S}^t - \tau \nabla_{\mathbf{S}} \mathcal{L}_n(\mathbf{U}^t \mathbf{V}^{t\top} + \mathbf{S}^t)\big)$, then we have $\mathbf{S}^{t+1} = \mathcal{T}_{\gamma\beta}(\bar{\mathbf{S}}^{t+1})$. To begin with according to Lemma B.3, we have

$$\|\mathbf{S}^{t+1} - \mathbf{S}^*\|_F^2 = \|\mathcal{T}_{\gamma\beta}(\bar{\mathbf{S}}^{t+1}) - \mathbf{S}^*\|_F^2 \leq \left(1 + \sqrt{\frac{2}{\gamma-1}}\right)^2 \cdot \|\bar{\mathbf{S}}^{t+1} - \mathbf{S}^*\|_F^2. \quad (B.2)$$

Moreover, denote $\Omega = \Omega^* \cup \Omega^t \cup \Omega^{t+1}$, where $\Omega^* = \text{supp}(\mathbf{S}^*)$, $\Omega^t = \text{supp}(\mathbf{S}^t)$ and $\Omega^{t+1} = \text{supp}(\bar{\mathbf{S}}^{t+1})$. Obviously, the cardinality of $\Omega$ satisfies $\gamma' s \leq |\Omega| \leq (2\gamma' + 1)s$. Based on $\Omega$, we define $\widetilde{\mathbf{S}}^{t+1}$ as follows

$$\widetilde{\mathbf{S}}^{t+1} = \mathcal{P}_\Omega\big(\mathbf{S}^t - \tau \nabla_{\mathbf{S}} \mathcal{L}_n(\mathbf{U}^t \mathbf{V}^{t\top} + \mathbf{S}^t)\big) = \mathbf{S}^t - \tau \mathcal{P}_\Omega\big(\nabla_{\mathbf{S}} \mathcal{L}_n(\mathbf{U}^t \mathbf{V}^{t\top} + \mathbf{S}^t)\big), \quad (B.3)$$

where $\mathcal{P}_\Omega$ is the projection operator onto the index set $\Omega$. Note that $\Omega^{t+1} \subseteq \Omega$, thus we have $\bar{\mathbf{S}}^{t+1} = \mathcal{H}_{\gamma s}(\widetilde{\mathbf{S}}^{t+1})$. According to Lemma 3.3 in Li et al. (2016), we have

$$\|\bar{\mathbf{S}}^{t+1} - \mathbf{S}^*\|_F^2 \leq \left(1 + \frac{2}{\sqrt{\gamma'-1}}\right) \cdot \|\widetilde{\mathbf{S}}^{t+1} - \mathbf{S}^*\|_F^2. \tag{B.4}$$

Therefore, it is sufficient to upper bound $\|\widetilde{\mathbf{S}}^{t+1} - \mathbf{S}^*\|_F^2$. By (B.3), we have

$$\|\widetilde{\mathbf{S}}^{t+1} - \mathbf{S}^*\|_F^2 = \|\mathbf{S}^t - \mathbf{S}^*\|_F^2 - 2\tau \underbrace{\langle \nabla_\mathbf{S} \mathcal{L}_n(\mathbf{X}^t + \mathbf{S}^t), \mathbf{S}^t - \mathbf{S}^* \rangle}_{I_1} + \tau^2 \underbrace{\|\mathcal{P}_\Omega(\nabla_\mathbf{S} \mathcal{L}_n(\mathbf{X}^t + \mathbf{S}^t))\|_F^2}_{I_2}, \tag{B.5}$$

where the equality holds because $\langle \mathcal{P}_\Omega(\mathbf{A}), \mathbf{B} \rangle = \langle \mathbf{A}, \mathcal{P}_\Omega(\mathbf{B}) \rangle$. In the following discussions, we are going to bound $I_1$ and $I_2$ respectively. Consider the term $I_1$ first, we have

$$I_1 = \underbrace{\langle \nabla_\mathbf{S} \mathcal{L}_n(\mathbf{X}^t + \mathbf{S}^t) - \nabla_\mathbf{S} \mathcal{L}_n(\mathbf{X}^t + \mathbf{S}^*), \mathbf{S}^t - \mathbf{S}^* \rangle}_{I_{11}} + \underbrace{\langle \nabla_\mathbf{S} \mathcal{L}_n(\mathbf{X}^t + \mathbf{S}^*) - \nabla_\mathbf{S} \mathcal{L}_n(\mathbf{X}^* + \mathbf{S}^*), \mathbf{S}^t - \mathbf{S}^* \rangle}_{I_{12}}$$
$$+ \underbrace{\langle \nabla_\mathbf{S} \mathcal{L}_n(\mathbf{X}^* + \mathbf{S}^*), \mathbf{S}^t - \mathbf{S}^* \rangle}_{I_{13}}. \tag{B.6}$$

As for the first term $I_{11}$ in (B.6), according to Lemma A.3, we have

$$I_{11} \geq \frac{\mu_2}{2}\|\mathbf{S}^t - \mathbf{S}^*\|_F^2 + \frac{1}{2L_2}\|\mathcal{P}_\Omega(\nabla_\mathbf{S}\mathcal{L}_n(\mathbf{X}^t + \mathbf{S}^t) - \nabla_\mathbf{S}\mathcal{L}_n(\mathbf{X}^t + \mathbf{S}^*))\|_F^2. \tag{B.7}$$

Note that we have $\text{supp}(\mathbf{S}^t - \mathbf{S}^*) \subseteq \Omega^t \cup \Omega^*$, where $\Omega^t \cup \Omega^*$ has at most $(\gamma + 1)\beta$-fraction nonzero entries per row and column. Denote $\mathbf{R}^t$ as the optimal rotation with respect to $\mathbf{Z}^t = [\mathbf{U}^t; \mathbf{V}^t]$, and $\mathbf{H}^t = \mathbf{Z}^t - \mathbf{Z}^*\mathbf{R}^t$. According to Condition 4.4, we obtain the bound of $I_{12}$ in (B.6)

$$|I_{12}| \leq |\langle \mathbf{X}^t - \mathbf{X}^*, \mathbf{S}^t - \mathbf{S}^* \rangle| + K\|\mathbf{X}^t - \mathbf{X}^*\|_F \cdot \|\mathbf{S}^t - \mathbf{S}^*\|_F$$
$$\leq \|\mathcal{P}_{\Omega^t \cup \Omega^*}(\mathbf{X}^t - \mathbf{X}^*)\|_F \cdot \|\mathbf{S}^t - \mathbf{S}^*\|_F + K\|\mathbf{X}^t - \mathbf{X}^*\|_F \cdot \|\mathbf{S}^t - \mathbf{S}^*\|_F$$
$$\leq \sqrt{18(\gamma+1)\beta\alpha r\sigma_1}\|\mathbf{H}^t\|_F \cdot \|\mathbf{S}^t - \mathbf{S}^*\|_F + K\|\mathbf{X}^t - \mathbf{X}^*\|_F \cdot \|\mathbf{S}^t - \mathbf{S}^*\|_F, \tag{B.8}$$

where the second inequality holds because $|\langle \mathbf{A}, \mathbf{B} \rangle| \leq \|\mathbf{A}\|_F \cdot \|\mathbf{B}\|_F$, and the last inequality follows from Lemma 14 in Yi et al. (2016). As for the last term $I_{13}$ in (B.6), we have

$$|I_{13}| \leq \|\nabla_\mathbf{S}\mathcal{L}_n(\mathbf{X}^* + \mathbf{S}^*)\|_{\infty,\infty} \cdot \|\mathbf{S}^t - \mathbf{S}^*\|_{1,1} \leq \sqrt{(\gamma'+1)s} \cdot \|\nabla_\mathbf{S}\mathcal{L}_n(\mathbf{X}^* + \mathbf{S}^*)\|_{\infty,\infty} \cdot \|\mathbf{S}^t - \mathbf{S}^*\|_F, \tag{B.9}$$

where the first inequality holds because $|\langle \mathbf{A}, \mathbf{B} \rangle| \leq \|\mathbf{A}\|_{\infty,\infty} \cdot \|\mathbf{B}\|_{1,1}$, and the second inequality follows from the fact that $\mathbf{S}^t - \mathbf{S}^*$ has at most $(\gamma'+1)s$ nonzero entries. Therefore, plugging (B.7), (B.8) and (B.9) into (B.6), we obtain the lower bound of $I_1$

$$I_1 \geq \frac{\mu_2}{8}\|\mathbf{S}^t - \mathbf{S}^*\|_F^2 + \frac{1}{2L_2}\|\mathcal{P}_\Omega(\nabla_\mathbf{S}\mathcal{L}_n(\mathbf{X}^t + \mathbf{S}^t) - \nabla_\mathbf{S}\mathcal{L}_n(\mathbf{X}^t + \mathbf{S}^*))\|_F^2 - \frac{2K^2}{\mu_2}\|\mathbf{X}^t - \mathbf{X}^*\|_F^2$$
$$- \frac{36(\gamma+1)\beta\alpha r\sigma_1}{\mu_2}\|\mathbf{H}^t\|_F^2 - \frac{2(\gamma'+1)s}{\mu_2}\|\nabla_\mathbf{S}\mathcal{L}_n(\mathbf{X}^* + \mathbf{S}^*)\|_{\infty,\infty}^2. \tag{B.10}$$

Next, consider the term $I_2$ in (B.5). We have

$$I_2 \leq 3\|\mathcal{P}_\Omega\big(\nabla_\mathbf{S}\mathcal{L}_n(\mathbf{X}^t+\mathbf{S}^t) - \nabla_\mathbf{S}\mathcal{L}_n(\mathbf{X}^t+\mathbf{S}^*)\big)\|_F^2 + 3\|\mathcal{P}_\Omega\big(\nabla_\mathbf{S}\mathcal{L}_n(\mathbf{X}^t+\mathbf{S}^*) - \nabla_\mathbf{S}\mathcal{L}_n(\mathbf{X}^*+\mathbf{S}^*)\big)\|_F^2$$
$$+ 3\|\mathcal{P}_\Omega\big(\nabla_\mathbf{S}\mathcal{L}_n(\mathbf{X}^*+\mathbf{S}^*)\big)\|_F^2 \quad \text{(B.11)}$$

As for the second term in (B.11), according to the definition of Frobenius norm, we have

$$\|\mathcal{P}_\Omega\big(\nabla_\mathbf{S}\mathcal{L}_n(\mathbf{X}^t+\mathbf{S}^*) - \nabla_\mathbf{S}\mathcal{L}_n(\mathbf{X}^*+\mathbf{S}^*)\big)\|_F = \sup_{\|\mathbf{W}\|\leq 1} \langle \mathcal{P}_\Omega\big(\nabla_\mathbf{S}\mathcal{L}_n(\mathbf{X}^t+\mathbf{S}^*) - \nabla_\mathbf{S}\mathcal{L}_n(\mathbf{X}^*+\mathbf{S}^*)\big), \mathbf{W}\rangle$$
$$= \sup_{\|\mathbf{W}\|\leq 1} \langle \nabla_\mathbf{S}\mathcal{L}_n(\mathbf{X}^t+\mathbf{S}^*) - \nabla_\mathbf{S}\mathcal{L}_n(\mathbf{X}^*+\mathbf{S}^*), \mathcal{P}_\Omega(\mathbf{W})\rangle$$
$$\leq (1+K)\cdot\|\mathbf{X}^t-\mathbf{X}^*\|_F \cdot \|\mathcal{P}_\Omega(\mathbf{W})\|_F$$
$$\leq (1+K)\cdot\|\mathbf{X}^t-\mathbf{X}^*\|_F, \quad \text{(B.12)}$$

where the second equality holds because $\langle\mathcal{P}_\Omega(\mathbf{A}),\mathbf{B}\rangle = \langle\mathbf{A},\mathcal{P}_\Omega(\mathbf{B})\rangle$, and the first inequality holds because of Condition 4.4. As for the last term in (B.11), note that $|\Omega| \leq (2\gamma'+1)s$, thus we have

$$\|\mathcal{P}_\Omega\big(\nabla_\mathbf{S}\mathcal{L}_n(\mathbf{X}^*+\mathbf{S}^*)\big)\|_F^2 \leq (2\gamma'+1)s\|\nabla_\mathbf{S}\mathcal{L}_n(\mathbf{X}^*+\mathbf{S}^*)\|_{\infty,\infty}^2. \quad \text{(B.13)}$$

Therefore, plugging (B.12) and (B.13) into (B.11), we obtain the upper bound of $I_2$

$$I_2 \leq 3\|\mathcal{P}_\Omega\big(\nabla_\mathbf{S}\mathcal{L}_n(\mathbf{X}^t+\mathbf{S}^t) - \nabla_\mathbf{S}\mathcal{L}_n(\mathbf{X}^t+\mathbf{S}^*)\big)\|_F^2 + 3(1+K)^2\cdot\|\mathbf{X}^t-\mathbf{X}^*\|_F^2$$
$$+ 3(2\gamma'+1)s\|\nabla_\mathbf{S}\mathcal{L}_n(\mathbf{X}^*+\mathbf{S}^*)\|_{\infty,\infty}^2. \quad \text{(B.14)}$$

If we set the step size $\tau \leq 1/(3L_2)$, then by plugging (B.10) and (B.14) into (B.5), we have

$$\|\widetilde{\mathbf{S}}^{t+1}-\mathbf{S}^*\|_F^2 \leq \left(1-\frac{\mu_2\tau}{4}\right)\cdot\|\mathbf{S}^t-\mathbf{S}^*\|_F^2 + C_3\|\mathbf{X}^t-\mathbf{X}^*\|_F^2 + C_4\|\mathbf{H}^t\|_F^2 + C_5\|\nabla_\mathbf{S}\mathcal{L}_n(\mathbf{X}^*+\mathbf{S}^*)\|_{\infty,\infty}^2,$$
(B.15)

where $C_3 = 4\tau K^2/\mu_2 + 3\tau^2(1+K)^2$, $C_4 = 72\tau(\gamma+1)\beta\alpha r\sigma_1/\mu_2$ and $C_5 = 4\tau(\gamma'+1)s/\mu_2 + 3\tau^2(2\gamma'+1)s$. Thus combining (B.2), (B.4) and (B.15), we obtain

$$\|\mathbf{S}^{t+1}-\mathbf{S}^*\|_F^2 \leq \rho\|\mathbf{S}^t-\mathbf{S}^*\|_F^2 + C(\gamma,\gamma')\cdot\big(C_3\|\mathbf{X}^t-\mathbf{X}^*\|_F^2 + C_4\|\mathbf{H}^t\|_F^2 + C_5\|\nabla_\mathbf{S}\mathcal{L}_n(\mathbf{X}^*+\mathbf{S}^*)\|_{\infty,\infty}^2\big),$$

which completes the proof. $\square$

### B.3 Proof of Lemma A.3

In order to proof Lemma A.3, we need to make use of the following lemma, which can be derived following the standard proof of Lipschitz continuous gradient property (Nesterov, 2004).

**Lemma B.4.** Suppose the sample loss function $\mathcal{L}_n$ satisfies Conditions 4.3. Given a fixed rank-$r$ matrix $\mathbf{X} \in \mathbb{R}^{d_1\times d_2}$, then for any sparse matrices $\mathbf{S}_1, \mathbf{S}_2 \in \mathbb{R}^{d_1\times d_2}$ with cardinality at most $\widetilde{s}$, we have

$$\mathcal{L}_n(\mathbf{X}+\mathbf{S}_1) \geq \mathcal{L}_n(\mathbf{X}+\mathbf{S}_2) + \langle\nabla_\mathbf{S}\mathcal{L}_n(\mathbf{X}+\mathbf{S}_2), \mathbf{S}_1-\mathbf{S}_2\rangle$$
$$+ \frac{1}{2L_2}\big\|\mathcal{P}_\Omega\big(\nabla_\mathbf{S}\mathcal{L}_n(\mathbf{X}+\mathbf{S}_1) - \nabla_\mathbf{S}\mathcal{L}_n(\mathbf{X}+\mathbf{S}_2)\big)\big\|_F^2,$$

where $\Omega \subseteq [d_1]\times[d_2]$ is an index set with cardinality at most $\widetilde{s}$ such that $\mathrm{supp}(\mathbf{S}_1) \subseteq \Omega$ and $\mathcal{P}_\Omega$ is the projection operator onto $\Omega$.

Now we are ready to prove Lemma A.3.

*Proof of Lemma A.3.* Since the sample loss function $\mathcal{L}_n$ satisfies the restricted strong convexity Condition 4.3, we have

$$\mathcal{L}_n(\mathbf{X} + \mathbf{S}_2) \geq \mathcal{L}_n(\mathbf{X} + \mathbf{S}_1) + \langle \nabla_{\mathbf{S}} \mathcal{L}_n(\mathbf{X} + \mathbf{S}_1), \mathbf{S}_2 - \mathbf{S}_1 \rangle + \frac{\mu_2}{2} \|\mathbf{S}_2 - \mathbf{S}_1\|_F^2. \tag{B.16}$$

According to Lemma B.4, we have

$$\begin{aligned}\mathcal{L}_n(\mathbf{X} + \mathbf{S}_1) \geq{}& \mathcal{L}_n(\mathbf{X} + \mathbf{S}_2) + \langle \nabla_{\mathbf{S}} \mathcal{L}_n(\mathbf{X} + \mathbf{S}_2), \mathbf{S}_1 - \mathbf{S}_2 \rangle \\ & + \frac{1}{2L_2} \big\| \mathcal{P}_\Omega \big( \nabla_{\mathbf{S}} \mathcal{L}_n(\mathbf{X} + \mathbf{S}_1) - \nabla_{\mathbf{S}} \mathcal{L}_n(\mathbf{X} + \mathbf{S}_2) \big) \big\|_F^2.\end{aligned} \tag{B.17}$$

Therefore, combining (B.16) and (B.17), we obtain

$$\begin{aligned}\langle \nabla_{\mathbf{S}} \mathcal{L}_n(\mathbf{X} + \mathbf{S}_1) - \nabla_{\mathbf{S}} \mathcal{L}_n(\mathbf{X} + \mathbf{S}_2), \mathbf{S}_1 - \mathbf{S}_2 \rangle \geq{}& \frac{\mu_2}{2} \|\mathbf{S}_1 - \mathbf{S}_2\|_F^2 \\ & + \frac{1}{2L_2} \big\| \mathcal{P}_\Omega \big( \nabla_{\mathbf{S}} \mathcal{L}_n(\mathbf{X} + \mathbf{S}_1) - \nabla_{\mathbf{S}} \mathcal{L}_n(\mathbf{X} + \mathbf{S}_2) \big) \big\|_F^2,\end{aligned}$$

which completes the proof. $\square$

## C  Proofs of Auxiliary Lemmas in Appendix B

To begin with, we introduce the following notations for simplicity. Consider $\mathbf{Z} \in \mathbb{R}^{(d_1+d_2) \times r}$, for $\mathbf{U} \in \mathbb{R}^{d_1 \times r}$ and $\mathbf{V} \in \mathbb{R}^{d_2 \times r}$, and $\mathbf{X} = \mathbf{U}\mathbf{V}^\top$, we let $\mathbf{Z} = [\mathbf{U}; \mathbf{V}]$. Let $\mathbf{R} = \operatorname{argmin}_{\widetilde{\mathbf{R}} \in \mathbb{Q}_r} \|\mathbf{Z} - \mathbf{Z}^* \widetilde{\mathbf{R}}\|_F$ be the optimal rotation regarding to $\mathbf{Z}$, and $\mathbf{H} = \mathbf{Z} - \mathbf{Z}^* \mathbf{R} = [\mathbf{H}_U; \mathbf{H}_V]$ with $\mathbf{H}_U \in \mathbb{R}^{d_1 \times r}$ and $\mathbf{H}_V \in \mathbb{R}^{d_2 \times r}$.

Besides, we introduce the following projection metrics, which are essential for proving the following lemmas. Denote by $\overline{\mathbf{U}}_1, \overline{\mathbf{U}}_2, \overline{\mathbf{U}}_3$ the left singular matrices of $\mathbf{X}, \mathbf{U}, \mathbf{H}_U$ respectively. Let $\widetilde{\mathbf{U}}$ be the matrix spanned by the column vectors of $\overline{\mathbf{U}}_1, \overline{\mathbf{U}}_2$ and $\overline{\mathbf{U}}_3$, i.e.,

$$\operatorname{col}(\widetilde{\mathbf{U}}) = \operatorname{span}\{\overline{\mathbf{U}}_1, \overline{\mathbf{U}}_2, \overline{\mathbf{U}}_3\} = \operatorname{col}(\overline{\mathbf{U}}_1) + \operatorname{col}(\overline{\mathbf{U}}_2) + \operatorname{col}(\overline{\mathbf{U}}_3). \tag{C.1}$$

It is easy to show that $\widetilde{\mathbf{U}}$ is an orthonormal matrix with at most $3r$ columns. Here, the sum of two subspaces is defined as $\mathbf{U}_1 + \mathbf{U}_2 = \{\mathbf{u}_1 + \mathbf{u}_2 \mid \mathbf{u}_1 \in \mathbf{U}_1, \mathbf{u}_2 \in \mathbf{U}_2\}$. Similarly, denote by $\overline{\mathbf{V}}_1, \overline{\mathbf{V}}_2, \overline{\mathbf{V}}_3$ the right singular matrices of $\mathbf{X}, \mathbf{V}, \mathbf{H}_V$ respectively. Again, let $\widetilde{\mathbf{V}}$ be the matrix spanned by the column of $\overline{\mathbf{V}}_1, \overline{\mathbf{V}}_2$ and $\overline{\mathbf{V}}_3$, i.e.,

$$\operatorname{col}(\widetilde{\mathbf{V}}) = \operatorname{span}\{\overline{\mathbf{V}}_1, \overline{\mathbf{V}}_2, \overline{\mathbf{V}}_3\} = \operatorname{col}(\overline{\mathbf{V}}_1) + \operatorname{col}(\overline{\mathbf{V}}_2) + \operatorname{col}(\overline{\mathbf{V}}_3), \tag{C.2}$$

where the rank of $\widetilde{\mathbf{V}}$ is at most $3r$.

## C.1 Proof of Lemma B.1

*Proof.* Recall $\mathbf{Z} = [\mathbf{U}; \mathbf{V}]$. We denote $\widetilde{\mathbf{Z}} = [\mathbf{U}; -\mathbf{V}] \in \mathbb{R}^{(d_1+d_2) \times r}$, then we can rewrite the regularization term $\|\mathbf{U}^\top \mathbf{U} - \mathbf{V}^\top \mathbf{V}\|_F^2$ as $\|\widetilde{\mathbf{Z}}^\top \mathbf{Z}\|_F^2$ and its gradient with respect to $\mathbf{Z}$ as $\nabla_{\mathbf{Z}}(\|\mathbf{U}^\top \mathbf{U} - \mathbf{V}^\top \mathbf{V}\|_F^2) = 4\widetilde{\mathbf{Z}}\widetilde{\mathbf{Z}}^\top \mathbf{Z}$. According to the formula of $\nabla \widetilde{F}_n(\mathbf{Z}, \mathbf{S})$ in (A.2), we have

$$\langle \nabla_{\mathbf{Z}} \widetilde{F}_n(\mathbf{Z}, \mathbf{S}), \mathbf{H} \rangle = \underbrace{\langle \nabla_{\mathbf{U}} \mathcal{L}_n(\mathbf{U}\mathbf{V}^\top + \mathbf{S}), \mathbf{H}_U \rangle + \langle \nabla_{\mathbf{V}} \mathcal{L}_n(\mathbf{U}\mathbf{V}^\top + \mathbf{S}), \mathbf{H}_V \rangle}_{I_1} + \frac{1}{2} \underbrace{\langle \widetilde{\mathbf{Z}}\widetilde{\mathbf{Z}}^\top \mathbf{Z}, \mathbf{H} \rangle}_{I_2}, \quad (C.3)$$

where $\widetilde{\mathbf{Z}} = [\mathbf{U}; -\mathbf{V}]$, and $\mathbf{H}_U, \mathbf{H}_V$ denote the top $d_1 \times r$ and bottom $d_2 \times r$ submatrices of $\mathbf{H}$ respectively. Note that $\nabla_{\mathbf{U}} \mathcal{L}_n(\mathbf{U}\mathbf{V}^\top + \mathbf{S}) = \nabla_{\mathbf{X}} \mathcal{L}_n(\mathbf{X} + \mathbf{S})\mathbf{V}$, and $\nabla_{\mathbf{V}} \mathcal{L}_n(\mathbf{U}\mathbf{V}^\top + \mathbf{S}) = [\nabla_{\mathbf{X}} \mathcal{L}_n(\mathbf{X} + \mathbf{S})]^\top \mathbf{U}$. Consider the term $I_1$ in (C.3) first, we have

$$I_1 = \langle \nabla_{\mathbf{X}} \mathcal{L}_n(\mathbf{X} + \mathbf{S}), \mathbf{U}\mathbf{V}^\top - \mathbf{U}^*\mathbf{V}^{*\top} + \mathbf{H}_U \mathbf{H}_V^\top \rangle$$
$$= \underbrace{\langle \nabla_{\mathbf{X}} \mathcal{L}_n(\mathbf{X}^* + \mathbf{S}^*), \mathbf{X} - \mathbf{X}^* + \mathbf{H}_U \mathbf{H}_V^\top \rangle}_{I_{11}} + \underbrace{\langle \nabla_{\mathbf{X}} \mathcal{L}_n(\mathbf{X}^* + \mathbf{S}) - \nabla_{\mathbf{X}} \mathcal{L}_n(\mathbf{X}^* + \mathbf{S}^*), \mathbf{X} - \mathbf{X}^* + \mathbf{H}_U \mathbf{H}_V^\top \rangle}_{I_{12}}$$
$$+ \underbrace{\langle \nabla_{\mathbf{X}} \mathcal{L}_n(\mathbf{X} + \mathbf{S}) - \mathcal{L}_n(\mathbf{X}^* + \mathbf{S}), \mathbf{X} - \mathbf{X}^* + \mathbf{H}_U \mathbf{H}_V^\top \rangle}_{I_{13}}. \quad (C.4)$$

In the following discussions, we are going to bound $I_{11}, I_{12}$ and $I_{13}$ respectively. For the first term $I_{11}$ in (C.4), we have

$$|I_{11}| \leq \|\nabla_{\mathbf{X}} \mathcal{L}_n(\mathbf{X}^* + \mathbf{S}^*)\|_2 \cdot \big(\|\mathbf{X} - \mathbf{X}^*\|_* + \|\mathbf{H}_U \mathbf{H}_V^\top\|_*\big)$$
$$\leq \|\nabla_{\mathbf{X}} \mathcal{L}_n(\mathbf{X}^* + \mathbf{S}^*)\|_2 \cdot \big(\sqrt{2r}\|\mathbf{X} - \mathbf{X}^*\|_F + \sqrt{r}\|\mathbf{H}_U \mathbf{H}_V\|_F\big)$$
$$\leq \frac{\mu_1}{16}\|\mathbf{X} - \mathbf{X}^*\|_F^2 + \frac{L_1}{16}\|\mathbf{H}\|_F^4 + \left(\frac{8r}{\mu_1} + \frac{r}{L_1}\right) \cdot \|\nabla_{\mathbf{X}} \mathcal{L}_n(\mathbf{X}^* + \mathbf{S}^*)\|_2^2, \quad (C.5)$$

where the first inequality holds because of Von Neumann trace inequality, the second inequality is due to $\mathbf{X} - \mathbf{X}^*$ has rank at most $2r$ and $\mathbf{H}_U \mathbf{H}_V^\top$ has rank at most $r$, and the last inequality holds because $\|\mathbf{H}_U \mathbf{H}_V^\top\|_F \leq \|\mathbf{H}_U\| \cdot \|\mathbf{H}_V\|_F \leq \|\mathbf{H}\|_F^2/2$ and $2ab \leq ta^2 + b^2/t$, for any $t > 0$. As for the second term $I_{12}$ in (C.4), note that $\mathbf{X} - \mathbf{X}^* + \mathbf{H}_U \mathbf{H}_V^\top$ has rank at most $3r$, thus according to the structural Lipschitz gradient Condition 4.4, we have

$$|I_{12}| \leq |\langle \mathbf{S} - \mathbf{S}^*, \mathbf{X} - \mathbf{X}^* + \mathbf{H}_U \mathbf{H}_V^\top \rangle| + K\|\mathbf{X} - \mathbf{X}^* + \mathbf{H}_U \mathbf{H}_V^\top\|_F \cdot \|\mathbf{S} - \mathbf{S}^*\|_F$$
$$\leq |\langle \mathbf{S} - \mathbf{S}^*, \mathbf{X} - \mathbf{X}^* \rangle| + \|\mathbf{S} - \mathbf{S}^*\|_F \cdot \|\mathbf{H}_U \mathbf{H}_V^\top\|_F + K\|\mathbf{X} - \mathbf{X}^* + \mathbf{H}_U \mathbf{H}_V^\top\|_F \cdot \|\mathbf{S} - \mathbf{S}^*\|_F$$
$$\leq |\langle \mathbf{S} - \mathbf{S}^*, \mathbf{X} - \mathbf{X}^* \rangle| + \frac{1+K}{2}\|\mathbf{S} - \mathbf{S}^*\|_F \cdot \|\mathbf{H}\|_F^2 + K\|\mathbf{X} - \mathbf{X}^*\|_F \cdot \|\mathbf{S} - \mathbf{S}^*\|_F, \quad (C.6)$$

where the second inequality follows from triangle inequality and the fact that $|\langle \mathbf{A}, \mathbf{B} \rangle| \leq \|\mathbf{A}\|_F \cdot \|\mathbf{B}\|_F$, and the last inequality is due to triangle inequality and the fact that $\|\mathbf{H}_U \mathbf{H}_V^\top\|_F \leq \|\mathbf{H}_U\|_F \cdot \|\mathbf{H}_V\|_F \leq \|\mathbf{H}\|_F^2/2$. Therefore, it suffices to bound the first term $|\langle \mathbf{S} - \mathbf{S}^*, \mathbf{X} - \mathbf{X}^* \rangle|$. Denote the support of $\mathbf{S} - \mathbf{S}^*$ by $\Omega$, then according to our assumption, $\Omega$ has at most $\beta' + \beta$ fraction nonzero entries per

row and column. By Lemma 14 in Yi et al. (2016), we further obtain

$$
\begin{aligned}
|I_{12}| &\leq \|\mathbf{S} - \mathbf{S}^*\|_F \cdot \|\mathcal{P}_\Omega(\mathbf{U}\mathbf{V}^\top - \mathbf{U}\mathbf{V}^*)\|_F + \frac{1+K}{2}\|\mathbf{S} - \mathbf{S}^*\|_F \cdot \|\mathbf{H}\|_F^2 + K\|\mathbf{X} - \mathbf{X}^*\|_F \cdot \|\mathbf{S} - \mathbf{S}^*\|_F \\
&\leq \sqrt{18(\beta' + \beta)\alpha r \sigma_1}\|\mathbf{H}\|_F \cdot \|\mathbf{S} - \mathbf{S}^*\|_F + \frac{1+K}{2}\|\mathbf{S} - \mathbf{S}^*\|_F \cdot \|\mathbf{H}\|_F^2 + K\|\mathbf{X} - \mathbf{X}^*\|_F \cdot \|\mathbf{S} - \mathbf{S}^*\|_F \\
&\leq \frac{\mu_1}{8}\|\mathbf{X} - \mathbf{X}^*\|_F^2 + \left(\frac{\mu_2}{2} + \frac{2K^2}{\mu_1}\right) \cdot \|\mathbf{S} - \mathbf{S}^*\|_F^2 + \frac{18(\beta' + \beta)\alpha r \sigma_1}{\mu_2}\|\mathbf{H}\|_F^2 + \frac{(1+K)^2}{4\mu_2}\|\mathbf{H}\|_F^4,
\end{aligned}
\tag{C.7}
$$

where the first inequality holds because $|\langle \mathbf{A}, \mathcal{P}_\Omega(\mathbf{B})\rangle| \leq \|\mathcal{P}_\Omega(\mathbf{A})\|_F \cdot \|\mathbf{B}\|_F$, and the second inequality is due to Lemma 14 in Yi et al. (2016), and the last inequality holds because $2ab \leq ta^2 + b^2/t$, for any $t > 0$. Finally, we consider the last term $I_{13}$ in (C.4). Recall the orthonormal projection matrices $\widetilde{\mathbf{U}}$ and $\widetilde{\mathbf{V}}$ in (C.1) and (C.2). According to Lemma B.2 in Wang et al. (2017), we have

$$
\begin{aligned}
\langle \nabla_\mathbf{X}\mathcal{L}_n(\mathbf{X} + \mathbf{S}) - \nabla_\mathbf{X}\mathcal{L}_n(\mathbf{X}^* + \mathbf{S}), \mathbf{X} - \mathbf{X}^*\rangle &\geq \frac{1}{4L_1}\|\widetilde{\mathbf{U}}^\top(\nabla_\mathbf{X}\mathcal{L}_n(\mathbf{X} + \mathbf{S}) - \nabla_\mathbf{X}\mathcal{L}_n(\mathbf{X}^* + \mathbf{S}))\|_F^2 \\
&+ \frac{1}{4L_1}\|(\nabla_\mathbf{X}\mathcal{L}_n(\mathbf{X} + \mathbf{S}) - \nabla_\mathbf{X}\mathcal{L}_n(\mathbf{X}^* + \mathbf{S}))\widetilde{\mathbf{V}}\|_F^2 + \frac{\mu_1}{2}\|\mathbf{X} - \mathbf{X}^*\|_F^2.
\end{aligned}
\tag{C.8}
$$

As for the remaining term in $I_{13}$, we have

$$
\begin{aligned}
|\langle \nabla_\mathbf{X}\mathcal{L}_n(\mathbf{X} + \mathbf{S}) - \nabla_\mathbf{X}\mathcal{L}_n(\mathbf{X}^* + \mathbf{S}), \mathbf{H}_U\mathbf{H}_V^\top\rangle| &= |\langle \nabla_\mathbf{X}\mathcal{L}_n(\mathbf{X} + \mathbf{S}) - \nabla_\mathbf{X}\mathcal{L}_n(\mathbf{X}^* + \mathbf{S}), \widetilde{\mathbf{U}}\widetilde{\mathbf{U}}^\top\mathbf{H}_U\mathbf{H}_V^\top\rangle| \\
&\leq \frac{1}{2}\|\widetilde{\mathbf{U}}^\top(\nabla_\mathbf{X}\mathcal{L}_n(\mathbf{X} + \mathbf{S}) - \nabla_\mathbf{X}\mathcal{L}_n(\mathbf{X}^* + \mathbf{S}))\|_F \cdot \|\mathbf{H}\|_F^2 \\
&\leq \frac{1}{2L_1}\|\widetilde{\mathbf{U}}^\top(\nabla_\mathbf{X}\mathcal{L}_n(\mathbf{X} + \mathbf{S}) - \nabla_\mathbf{X}\mathcal{L}_n(\mathbf{X}^* + \mathbf{S})\|_F^2 + \frac{L_1}{8}\|\mathbf{H}\|_F^4,
\end{aligned}
\tag{C.9}
$$

where the equality is due to the fact that $\text{col}(\overline{\mathbf{U}}_3) \subseteq \text{col}(\widetilde{\mathbf{U}})$, where $\overline{\mathbf{U}}_3$ is the left singular matrix of $\mathbf{H}_U$, which implies that $\widetilde{\mathbf{U}}\widetilde{\mathbf{U}}^\top\mathbf{H}_U = \mathbf{H}_U$, the first inequality holds because $|\langle \mathbf{A}, \mathbf{BC}\rangle| \leq \|\mathbf{A}\|_F \cdot \|\mathbf{BC}\|_F \leq \|\mathbf{A}\|_F \cdot \|\mathbf{B}\|_2 \cdot \|\mathbf{C}\|_F$ and $\|\widetilde{\mathbf{U}}\|_2 = 1$, and the last inequality holds because $2ab \leq ta^2 + b^2/t$, for any $t > 0$. Similarly, we have

$$
\begin{aligned}
|\langle \nabla_\mathbf{X}\mathcal{L}_n(\mathbf{X} + \mathbf{S}) - \nabla_\mathbf{X}\mathcal{L}_n(\mathbf{X}^* + \mathbf{S}), \mathbf{H}_U\mathbf{H}_V^\top\rangle| &\leq \frac{1}{2}\|(\nabla_\mathbf{X}\mathcal{L}_n(\mathbf{X} + \mathbf{S}) - \nabla_\mathbf{X}\mathcal{L}_n(\mathbf{X}^* + \mathbf{S}))\widetilde{\mathbf{V}}\|_F \cdot \|\mathbf{H}\|_F^2 \\
&\leq \frac{1}{2L_1}\|(\nabla_\mathbf{X}\mathcal{L}_n(\mathbf{X} + \mathbf{S}) - \nabla_\mathbf{X}\mathcal{L}_n(\mathbf{X}^* + \mathbf{S}))\widetilde{\mathbf{V}}\|_F^2 + \frac{L_1}{8}\|\mathbf{H}\|_F^4.
\end{aligned}
\tag{C.10}
$$

Therefore, combining (C.8), (C.9) and (C.10), we obtain the lower bound of $I_{13}$

$$
I_{13} \geq \frac{\mu_1}{2}\|\mathbf{X} - \mathbf{X}^*\|_F^2 - \frac{L_1}{8}\|\mathbf{H}\|_F^4.
\tag{C.11}
$$

Hence, combining (C.5), (C.7) and (C.11), we further obtain the lower bound of $I_1$ in (C.3)

$$
\begin{aligned}
I_1 &\geq \frac{3\mu_1}{8}\|\mathbf{X} - \mathbf{X}^*\|_F^2 - \frac{18(\beta' + \beta)\alpha r \sigma_1}{\mu_2}\|\mathbf{H}\|_F^2 - \left(\frac{L_1}{8} + \frac{(1+K)^2}{4\mu_2}\right) \cdot \|\mathbf{H}\|_F^4 \\
&- \left(\frac{\mu_2}{2} + \frac{2K^2}{\mu_1}\right) \cdot \|\mathbf{S} - \mathbf{S}^*\|_F^2 - \left(\frac{8r}{\mu_1} + \frac{r}{L_1}\right) \cdot \|\nabla_\mathbf{X}\mathcal{L}_n(\mathbf{X}^* + \mathbf{S}^*)\|_2^2.
\end{aligned}
\tag{C.12}
$$

Besides, according to Lemma C.1 in Wang et al. (2016), we obtain the following lower bound regarding $I_2$ in (C.3)

$$I_2 \geq \frac{1}{2}\|\widetilde{\mathbf{Z}}^\top \mathbf{Z}\|_F^2 - \frac{1}{2}\|\widetilde{\mathbf{Z}}^\top \mathbf{Z}\|_F \cdot \|\mathbf{H}\|_F^2 \geq \frac{1}{4}\|\widetilde{\mathbf{Z}}^\top \mathbf{Z}\|_F^2 - \frac{1}{4}\|\mathbf{H}\|_F^4. \tag{C.13}$$

Note that $K \in (0, 1)$, by plugging (C.12) and (C.13) into (C.3), we have

$$\langle \nabla_{\mathbf{Z}} \widetilde{F}_n(\mathbf{Z}, \mathbf{S}), \mathbf{H} \rangle \geq \frac{3\mu_1}{8}\|\mathbf{X} - \mathbf{X}^*\|_F^2 + \frac{1}{8}\|\widetilde{\mathbf{Z}}^\top \mathbf{Z}\|_F^2 - \frac{18(\beta' + \beta)\alpha r \sigma_1}{\mu_2}\|\mathbf{H}\|_F^2 - \left(\frac{L_1 + 1}{8} + \frac{1}{\mu_2}\right) \cdot \|\mathbf{H}\|_F^4$$
$$- \left(\frac{\mu_2}{2} + \frac{2K^2}{\mu_1}\right) \cdot \|\mathbf{S} - \mathbf{S}^*\|_F^2 - \left(\frac{8r}{\mu_1} + \frac{r}{L_1}\right) \cdot \|\nabla_{\mathbf{X}} \mathcal{L}_n(\mathbf{X}^* + \mathbf{S}^*)\|_2^2. \tag{C.14}$$

Furthermore, denote $\widetilde{\mathbf{Z}}^* = [\mathbf{U}^*; -\mathbf{V}^*]$, then we obtain the following result

$$\|\widetilde{\mathbf{Z}}^\top \mathbf{Z}\|_F^2 = \langle \mathbf{Z}\mathbf{Z}^\top - \mathbf{Z}^*\mathbf{Z}^{*\top}, \widetilde{\mathbf{Z}}\widetilde{\mathbf{Z}}^\top - \widetilde{\mathbf{Z}}^*\widetilde{\mathbf{Z}}^{*\top} \rangle + \langle \mathbf{Z}^*\mathbf{Z}^{*\top}, \widetilde{\mathbf{Z}}\widetilde{\mathbf{Z}}^\top \rangle + \langle \mathbf{Z}\mathbf{Z}^\top, \widetilde{\mathbf{Z}}^*\widetilde{\mathbf{Z}}^{*\top} \rangle$$
$$\geq \langle \mathbf{Z}\mathbf{Z}^\top - \mathbf{Z}^*\mathbf{Z}^{*\top}, \widetilde{\mathbf{Z}}\widetilde{\mathbf{Z}}^\top - \widetilde{\mathbf{Z}}^*\widetilde{\mathbf{Z}}^{*\top} \rangle$$
$$= \|\mathbf{U}\mathbf{U}^\top - \mathbf{U}^*\mathbf{U}^{*\top}\|_F^2 + \|\mathbf{V}\mathbf{V}^\top - \mathbf{V}^*\mathbf{V}^{*\top}\|_F^2 - 2\|\mathbf{X} - \mathbf{X}^*\|_F^2, \tag{C.15}$$

where the first equality follows from $\widetilde{\mathbf{Z}}^{*\top}\mathbf{Z}^* = 0$, and the inequality is due to the fact that $\langle \mathbf{A}\mathbf{A}^\top, \mathbf{B}\mathbf{B}^\top \rangle = \|\mathbf{A}^\top \mathbf{B}\|_F^2 \geq 0$. Therefore, by (C.15), we obtain

$$4\|\mathbf{X} - \mathbf{X}^*\|_F^2 + \|\widetilde{\mathbf{Z}}^\top \mathbf{Z}\|_F^2 = \|\mathbf{Z}\mathbf{Z}^\top - \mathbf{Z}^*\mathbf{Z}^{*\top}\|_F^2 \geq 4(\sqrt{2} - 1)\sigma_r\|\mathbf{H}\|_F^2, \tag{C.16}$$

where the inequality follows from Lemma 5.4 in Tu et al. (2015) and the fact that $\sigma_r^2(\mathbf{Z}^*) = 2\sigma_r(\mathbf{X}^*) = 2\sigma_r$. Denote $\mu_1' = \min\{\mu_1, 2\}$, then by plugging (C.16) into (C.14), we obtain

$$\langle \nabla_{\mathbf{Z}} \widetilde{F}_n(\mathbf{Z}, \mathbf{S}), \mathbf{H} \rangle \geq \frac{\mu_1}{4}\|\mathbf{X} - \mathbf{X}^*\|_F^2 + \frac{1}{16}\|\widetilde{\mathbf{Z}}^\top \mathbf{Z}\|_F^2 + \left(\frac{\mu_1'}{20}\sigma_r - \frac{18(\beta' + \beta)\alpha r \sigma_1}{\mu_2}\right) \cdot \|\mathbf{H}\|_F^2$$
$$- \left(\frac{L_1 + 1}{8} + \frac{1}{\mu_2}\right) \cdot \|\mathbf{H}\|_F^4 - \left(\frac{\mu_2}{2} + \frac{2K^2}{\mu_1}\right) \cdot \|\mathbf{S} - \mathbf{S}^*\|_F^2 - \left(\frac{8r}{\mu_1} + \frac{r}{L_1}\right) \cdot \|\nabla_{\mathbf{X}} \mathcal{L}_n(\mathbf{X}^* + \mathbf{S}^*)\|_2^2,$$

which completes the proof. $\square$

### C.2 Proof of Lemma B.2

*Proof.* According to the formula of $\nabla_{\mathbf{Z}} \widetilde{F}_n(\mathbf{Z}, \mathbf{S})$ in (A.2), we have

$$\|\nabla_{\mathbf{Z}} \widetilde{F}_n(\mathbf{Z}, \mathbf{S})\|_F^2 \leq 2\|\nabla_{\mathbf{U}} \mathcal{L}_n(\mathbf{U}\mathbf{V}^\top + \mathbf{S})\|_F^2 + 2\|\nabla_{\mathbf{V}} \mathcal{L}_n(\mathbf{U}\mathbf{V}^\top + \mathbf{S})\|_F^2 + \|\mathbf{U}^\top \mathbf{U} - \mathbf{V}^\top \mathbf{V}\|_F^2 \cdot \|\mathbf{Z}\|_2^2, \tag{C.17}$$

where the inequality follows from the fact that $\|\mathbf{A} + \mathbf{B}\|_F^2 \leq 2\|\mathbf{A}\|_F^2 + 2\|\mathbf{B}\|_F^2$, $\|\mathbf{A}\mathbf{B}\|_F \leq \|\mathbf{A}\|_2 \cdot \|\mathbf{B}\|_F$, and $\max\{\|\mathbf{U}\|_2, \|\mathbf{V}\|_2\} \leq \|\mathbf{Z}\|_2$. Consider the first term $\|\nabla_{\mathbf{U}} \mathcal{L}_n(\mathbf{U}\mathbf{V}^\top + \mathbf{S})\|_F^2$. Denote $\mathbf{X} = \mathbf{U}\mathbf{V}^\top$, then we have

$$\|\nabla_{\mathbf{U}} \mathcal{L}_n(\mathbf{U}\mathbf{V}^\top + \mathbf{S})\|_F^2 \leq 3\underbrace{\|(\nabla_{\mathbf{X}} \mathcal{L}_n(\mathbf{X} + \mathbf{S}) - \nabla_{\mathbf{X}} \mathcal{L}_n(\mathbf{X}^* + \mathbf{S}))\mathbf{V}\|_F^2}_{I_1}$$
$$+ 3\underbrace{\|(\nabla_{\mathbf{X}} \mathcal{L}_n(\mathbf{X}^* + \mathbf{S}) - \nabla_{\mathbf{X}} \mathcal{L}_n(\mathbf{X}^* + \mathbf{S}^*))\mathbf{V}\|_F^2}_{I_2} + 3\underbrace{\|\nabla_{\mathbf{X}} \mathcal{L}_n(\mathbf{X}^* + \mathbf{S}^*)\mathbf{V}\|_F^2}_{I_3}, \tag{C.18}$$

where the inequality holds because $\nabla_{\mathbf{U}}\mathcal{L}_n(\mathbf{UV}^\top + \mathbf{S}) = \nabla_{\mathbf{X}}\mathcal{L}_n(\mathbf{X} + \mathbf{S})\mathbf{V}$ and $\|\mathbf{A} + \mathbf{B} + \mathbf{C}\|_F^2 \leq 3(\|\mathbf{A}\|_F^2 + \|\mathbf{B}\|_F^2 + \|\mathbf{C}\|_F^2)$. In the following discussion, we are going to upper bound $I_1, I_2$ and $I_3$ separately. As for $I_1$, according to the orthonormal projection matrix $\widetilde{\mathbf{V}}$ defined in (C.2), we have

$$
\begin{aligned}
I_1 &= \left\|\left(\nabla_{\mathbf{X}}\mathcal{L}_n(\mathbf{X} + \mathbf{S}) - \nabla_{\mathbf{X}}\mathcal{L}_n(\mathbf{X}^* + \mathbf{S})\right)\widetilde{\mathbf{V}}\widetilde{\mathbf{V}}^\top \mathbf{V}\right\|_F^2 \\
&\leq \left\|\left(\nabla_{\mathbf{X}}\mathcal{L}_n(\mathbf{X} + \mathbf{S}) - \nabla_{\mathbf{X}}\mathcal{L}_n(\mathbf{X}^* + \mathbf{S})\right)\widetilde{\mathbf{V}}\right\|_F^2 \cdot \|\widetilde{\mathbf{V}}^\top \mathbf{V}\|_2^2 \\
&\leq \left\|\left(\nabla_{\mathbf{X}}\mathcal{L}_n(\mathbf{X} + \mathbf{S}) - \nabla_{\mathbf{X}}\mathcal{L}_n(\mathbf{X}^* + \mathbf{S})\right)\widetilde{\mathbf{V}}\right\|_F^2 \cdot \|\mathbf{V}\|_2^2,
\end{aligned} \quad \text{(C.19)}
$$

where the equality holds because $\text{col}(\mathbf{V}) \subseteq \text{col}(\widetilde{\mathbf{V}})$, which implies that $\widetilde{\mathbf{V}}\widetilde{\mathbf{V}}^\top \mathbf{V} = \mathbf{V}$, the first inequality is due to the fact that $\|\mathbf{AB}\|_F \leq \|\mathbf{A}\|_F \cdot \|\mathbf{B}\|_2$, and the last inequality holds because $\|\mathbf{AB}\|_2 \leq \|\mathbf{A}\|_2 \cdot \|\mathbf{B}\|_2$ and the fact that $\widetilde{\mathbf{V}}$ is orthonormal. Moreover, consider the second term $I_2$ in (C.18). According to the definition of Frobenius norm, we have

$$
\begin{aligned}
\left\|\left(\nabla_{\mathbf{X}}\mathcal{L}_n(\mathbf{X}^* + \mathbf{S}) - \nabla_{\mathbf{X}}\mathcal{L}_n(\mathbf{X}^* + \mathbf{S}^*)\right)\mathbf{V}\right\|_F &= \sup_{\|\mathbf{W}\|_F \leq 1} \left\langle \left(\nabla_{\mathbf{X}}\mathcal{L}_n(\mathbf{X}^* + \mathbf{S}) - \nabla_{\mathbf{X}}\mathcal{L}_n(\mathbf{X}^* + \mathbf{S}^*)\right)\mathbf{V}, \mathbf{W}\right\rangle \\
&\leq \sup_{\|\mathbf{W}\|_F \leq 1} (1+K) \cdot \|\mathbf{S} - \mathbf{S}^*\|_F \cdot \|\mathbf{W}\mathbf{V}^\top\|_F \\
&\leq (1+K) \cdot \|\mathbf{S} - \mathbf{S}^*\|_F \cdot \|\mathbf{V}\|_2, \quad \text{(C.20)}
\end{aligned}
$$

where the first inequality follows from the structural Lipschitz gradient Condition 4.4 and the fact that $|\langle \mathbf{A}, \mathbf{B}\rangle| \leq \|\mathbf{A}\|_F \cdot \|\mathbf{B}\|_F$, and the second one holds because $\|\mathbf{AB}\|_F \leq \|\mathbf{A}\|_F \cdot \|\mathbf{B}\|_2$ and $\|\mathbf{W}\|_F \leq 1$. Finally, consider the last term $I_3$ in (C.18), we have

$$
I_3 \leq \|\nabla_{\mathbf{X}}\mathcal{L}_n(\mathbf{X}^* + \mathbf{S}^*)\|_2^2 \cdot \|\mathbf{V}\|_F^2 \leq r\|\nabla_{\mathbf{X}}\mathcal{L}_n(\mathbf{X}^* + \mathbf{S}^*)\|_2^2 \cdot \|\mathbf{V}\|_2^2. \quad \text{(C.21)}
$$

Thus, combining (C.19), (C.20) and (C.21), we obtain

$$
\begin{aligned}
\|\nabla_{\mathbf{U}}\mathcal{L}_n(\mathbf{UV}^\top + \mathbf{S})\|_F^2 \leq &\, 3\left\|\left(\nabla_{\mathbf{X}}\mathcal{L}_n(\mathbf{X} + \mathbf{S}) - \nabla_{\mathbf{X}}\mathcal{L}_n(\mathbf{X}^* + \mathbf{S})\right)\widetilde{\mathbf{V}}\right\|_F^2 \cdot \|\mathbf{V}\|_2^2 \\
&+ 3(1+K)^2 \cdot \|\mathbf{S} - \mathbf{S}^*\|_F^2 \cdot \|\mathbf{V}\|_2^2 + 3r\|\nabla_{\mathbf{X}}\mathcal{L}_n(\mathbf{X}^* + \mathbf{S}^*)\|_2^2 \cdot \|\mathbf{V}\|_2^2. \quad \text{(C.22)}
\end{aligned}
$$

As for the second term $\|\nabla_{\mathbf{V}}\mathcal{L}_n(\mathbf{UV}^\top + \mathbf{S})\|_F^2$ in (C.17), based on similar techniques, we obtain

$$
\begin{aligned}
\|\nabla_{\mathbf{V}}\mathcal{L}_n(\mathbf{UV}^\top + \mathbf{S})\|_F^2 \leq &\, 3\left\|\widetilde{\mathbf{U}}^\top\left(\nabla_{\mathbf{X}}\mathcal{L}_n(\mathbf{X} + \mathbf{S}) - \nabla_{\mathbf{X}}\mathcal{L}_n(\mathbf{X}^* + \mathbf{S})\right)\right\|_F^2 \cdot \|\mathbf{U}\|_2^2 \\
&+ 3(1+K)^2 \cdot \|\mathbf{S} - \mathbf{S}^*\|_F^2 \cdot \|\mathbf{U}\|_2^2 + 3r\|\nabla_{\mathbf{X}}\mathcal{L}_n(\mathbf{X}^* + \mathbf{S}^*)\|_2^2 \cdot \|\mathbf{U}\|_2^2, \quad \text{(C.23)}
\end{aligned}
$$

where $\widetilde{\mathbf{U}}$ is an orthonormal matrix defined in (C.1). According to Lemma C.1 in Wang et al. (2017) and Condition 4.2, we have

$$
\begin{aligned}
I &= \left\|\left(\nabla_{\mathbf{X}}\mathcal{L}_n(\mathbf{X}^* + \mathbf{S}) - \nabla_{\mathbf{X}}\mathcal{L}_n(\mathbf{X} + \mathbf{S})\right)\widetilde{\mathbf{V}}\right\|_F^2 + \left\|\widetilde{\mathbf{U}}^\top\left(\nabla_{\mathbf{X}}\mathcal{L}_n(\mathbf{X}^* + \mathbf{S}) - \nabla_{\mathbf{X}}\mathcal{L}_n(\mathbf{X} + \mathbf{S})\right)\right\|_F^2 \\
&\leq 4L_1\big(\mathcal{L}_n(\mathbf{X}^* + \mathbf{S}) - \mathcal{L}_n(\mathbf{X} + \mathbf{S}) - \langle\nabla_{\mathbf{X}}\mathcal{L}_n(\mathbf{X} + \mathbf{S}), \mathbf{X}^* - \mathbf{X}\rangle\big) \leq 2L_1^2 \cdot \|\mathbf{X} - \mathbf{X}^*\|_F^2. \quad \text{(C.24)}
\end{aligned}
$$

Therefore, plugging (C.22), (C.23) and (C.24) into (C.17), we obtain

$$
\begin{aligned}
\|\nabla_{\mathbf{Z}}\widetilde{F}_n(\mathbf{Z}, \mathbf{S})\|_F^2 \leq &\, \Big(12L_1^2\|\mathbf{X} - \mathbf{X}^*\|_F^2 + 12(1+K)^2 \cdot \|\mathbf{S} - \mathbf{S}^*\|_F^2 + \|\mathbf{U}^\top\mathbf{U} - \mathbf{V}^\top\mathbf{V}\|_F^2\Big) \cdot \|\mathbf{Z}\|_2^2 \\
&+ 12r\|\nabla_{\mathbf{X}}\mathcal{L}_n(\mathbf{X}^* + \mathbf{S}^*)\|_2^2 \cdot \|\mathbf{Z}\|_2^2,
\end{aligned}
$$

where the inequality holds because $\max\{\|\mathbf{U}\|_2, \|\mathbf{V}\|_2\} \leq \|\mathbf{Z}\|_2$. Thus, we finish the proof. $\square$

## C.3 Proof of Lemma B.3

*Proof.* Denote the support of $\mathbf{S}^*$ and $\mathcal{T}_{\gamma\beta}(\mathbf{S})$ by $\Omega^*$ and $\Omega$ respectively. According to the definition of the truncation operator $\mathcal{T}_\alpha$, we have

$$\begin{aligned}\|\mathcal{T}_{\gamma\beta}(\mathbf{S}) - \mathbf{S}^*\|_F^2 &= \|\mathcal{P}_\Omega(\mathcal{T}_{\gamma\beta}(\mathbf{S}) - \mathbf{S}^*)\|_F^2 + \|\mathcal{P}_{\Omega^*\setminus\Omega}(\mathcal{T}_{\gamma\beta}(\mathbf{S}) - \mathbf{S}^*)\|_F^2 \\ &= \|\mathcal{P}_\Omega(\mathbf{S} - \mathbf{S}^*)\|_F^2 + \|\mathcal{P}_{\Omega^*\setminus\Omega}(-\mathbf{S}^*)\|_F^2,\end{aligned} \quad (C.25)$$

where the second inequality holds because $[\mathcal{T}_{\gamma\beta}(\mathbf{S})]_{i,j} = S_{i,j}$ if $(i,j) \in \Omega$, and $[\mathcal{T}_{\gamma\beta}(\mathbf{S})]_{i,j} = 0$ otherwise. For any $(i,j) \in \Omega^* \setminus \Omega$, we claim

$$\left|(\mathbf{S} - \mathbf{S}^* + \mathbf{S}^*)_{i,j}\right| \leq \max\left\{\underbrace{\left|(\mathbf{S} - \mathbf{S}^*)_{i,*}^{(\gamma\beta d_2 - \beta d_2)}\right|}_{I_1}, \underbrace{\left|(\mathbf{S} - \mathbf{S}^*)_{*,j}^{(\gamma\beta d_1 - \beta d_1)}\right|}_{I_2}\right\}, \quad (C.26)$$

where we denote the $k$-th largest element in magnitude of $(\mathbf{S} - \mathbf{S}^*)_{i,*}$ by $(\mathbf{S} - \mathbf{S}^*)_{i,*}^{(k)}$, and the $k$-th largest element in magnitude of $(\mathbf{S} - \mathbf{S}^*)_{*,j}$ by $(\mathbf{S} - \mathbf{S}^*)_{*,j}^{(k)}$. In the following discussion, we are going to prove claim (C.26) by contradiction. Suppose $\left|(\mathbf{S} - \mathbf{S}^* + \mathbf{S}^*)_{i,j}\right| = |S_{i,j}| > \max\{I_1, I_2\}$, where $(i,j) \in \Omega^* \setminus \Omega$. Noticing $\mathbf{S}^*$ has at most $\beta$-fraction nonzero entries per row and column, we have

$$I_1 \geq \left|\mathbf{S}_{i,*}^{(\gamma\beta d_2 - \beta d_2)}\right| \quad \text{and} \quad I_2 \geq \left|\mathbf{S}_{*,j}^{(\gamma\beta d_1 - \beta d_1)}\right|.$$

Thus we have $|S_{i,j}| \geq \max\left\{\left|\mathbf{S}_{i,*}^{(\gamma\beta d_2 - \beta d_2)}\right|, \left|\mathbf{S}_{*,j}^{(\gamma\beta d_1 - \beta d_1)}\right|\right\}$, which contradicts with the fact that $(i,j) \in \Omega^* \setminus \Omega$. Therefore, based on (C.26), we obtain

$$\begin{aligned}\|\mathcal{P}_{\Omega^*\setminus\Omega}(\mathbf{S} - \mathbf{S}^* + \mathbf{S}^*)\|_F^2 &= \sum_{(i,j)\in\Omega^*\setminus\Omega}\left|(\mathbf{S} - \mathbf{S}^* + \mathbf{S}^*)_{i,j}\right|^2 \\ &\leq \sum_{(i,j)\in\Omega^*\setminus\Omega}\frac{\|(\mathbf{S} - \mathbf{S}^*)_{i,*}\|_2^2}{(\gamma-1)\beta d_2} + \sum_{(i,j)\in\Omega^*\setminus\Omega}\frac{\|(\mathbf{S} - \mathbf{S}^*)_{*,j}\|_2^2}{(\gamma-1)\beta d_1} \\ &\leq \frac{2}{\gamma-1}\|\mathbf{S} - \mathbf{S}^*\|_F^2,\end{aligned} \quad (C.27)$$

where the first inequality is due to (C.26), and the second inequality holds because for each row and column of $\Omega^*$, it has at most $\beta$-fraction nonzero elements. Thus we obtain

$$\begin{aligned}\|\mathcal{P}_{\Omega^*\setminus\Omega}(-\mathbf{S}^*)\|_F^2 &= \|\mathcal{P}_{\Omega^*\setminus\Omega}(\mathbf{S} - \mathbf{S}^*) - \mathcal{P}_{\Omega^*\setminus\Omega}(\mathbf{S} - \mathbf{S}^* + \mathbf{S}^*)\|_F^2 \\ &\leq (1+c)\cdot\|\mathcal{P}_{\Omega^*\setminus\Omega}(\mathbf{S} - \mathbf{S}^*)\|_F^2 + \left(1 + \frac{1}{c}\right)\cdot\|\mathcal{P}_{\Omega^*\setminus\Omega}(\mathbf{S} - \mathbf{S}^* + \mathbf{S}^*)\|_F^2 \\ &\leq (1+c)\cdot\|\mathcal{P}_{\Omega^*\setminus\Omega}(\mathbf{S} - \mathbf{S}^*)\|_F^2 + \frac{c+1}{c}\cdot\frac{2}{\gamma-1}\|\mathbf{S} - \mathbf{S}^*\|_F^2,\end{aligned} \quad (C.28)$$

where the second inequality holds because $\|\mathbf{A}+\mathbf{B}\|_F^2 \leq (1+c) \cdot \|\mathbf{A}\|_F^2 + (1+1/c) \cdot \|\mathbf{B}\|_F^2$, for any $c > 0$, and the second inequality is due to (C.27). Therefore, plugging in (C.28) into (C.25), we have

$$\|\mathcal{T}_{\gamma\beta}(\mathbf{S}) - \mathbf{S}^*\|_F^2 \leq \|\mathcal{P}_\Omega(\mathbf{S}-\mathbf{S}^*)\|_F^2 + (1+c) \cdot \|\mathcal{P}_{\Omega^*\setminus\Omega}(\mathbf{S}-\mathbf{S}^*)\|_F^2 + \frac{c+1}{c} \cdot \frac{2}{\gamma-1}\|\mathbf{S}-\mathbf{S}^*\|_F^2$$

$$\leq \left(1 + c + \frac{2(c+1)}{c(\gamma-1)}\right) \cdot \|\mathbf{S}-\mathbf{S}^*\|_F^2$$

$$= \left(1 + \frac{2}{\gamma-1} + 2\sqrt{\frac{2}{\gamma-1}}\right) \cdot \|\mathbf{S}-\mathbf{S}^*\|_F^2,$$

where we set $c = \sqrt{(\gamma-1)/2}$ in the last step. Thus we complete the proof. $\square$

# D  Proofs for Specific Models

In this section, we provide proofs for specific models. In the following discussions, we let $d = \max\{d_1, d_2\}$.

## D.1  Proofs for Robust Matrix Sensing

For matrix sensing, recall that we have the linear measurement operator $\mathcal{A}$ with each sensing matrix $\mathbf{A}_i$ sampled independently from $\boldsymbol{\Sigma}$-Gaussian ensemble, where $\text{vec}(\mathbf{A}_i) \sim N(0, \boldsymbol{\Sigma})$. In particular, we consider $\boldsymbol{\Sigma} = \mathbf{I}$ and here $\text{vec}(\mathbf{A}_i)$ denotes the vectorization of matrix $\mathbf{A}_i$. In order to prove the results for matrix sensing, we first lay out several lemmas, which are essential to prove the results for robust matrix sensing. The first lemma is useful to verify the restricted strong convexity and smoothness conditions in Condition 4.2.

**Lemma D.1.** (Negahban and Wainwright, 2011) Suppose we have the linear measurement operator $\mathcal{A}$ with each sensing matrix $\mathbf{A}_i$ sampled independently from $\mathbf{I}$-Gaussian ensemble, then there exists constants $c_0, c_1$ such that for all $\boldsymbol{\Delta} \in \mathbb{R}^{d_1 \times d_2}$ with rank at most $2\widetilde{r}$, it holds with probability at least $1 - \exp(-c_0 n)$ that

$$\left|\frac{\|\mathcal{A}(\boldsymbol{\Delta})\|_2^2}{n} - \frac{1}{2}\|\text{vec}(\boldsymbol{\Delta})\|_2^2\right| \leq c_1 \frac{\widetilde{r}d}{n}\|\boldsymbol{\Delta}\|_F^2. \tag{D.1}$$

The second lemma is useful to verify the restricted strong convexity and smoothness conditions in Condition 4.3.

**Lemma D.2.** (Raskutti et al., 2010) For any random matrix $\mathbf{A} \in \mathbb{R}^{n \times d_1 d_2}$, which is drawn from the $\boldsymbol{\Sigma}$-Gaussian ensemble, and the cardinalities of all vector $\mathbf{s} \in \mathbb{R}^{d_1 d_2}$ satisfy $|\mathbf{s}| \leq \widetilde{s}$. If we have sample size $n \geq c_2 \widetilde{s} \log d$, then the following inequality holds with probability at least $1 - c_3 \exp(-c_4 n)$

$$c_5\|\boldsymbol{\Sigma}^{1/2}\mathbf{s}\|_2^2 - c_6\frac{\log d}{n}\|\mathbf{s}\|_1 \leq \frac{\|\mathbf{A}\mathbf{s}\|_2^2}{n} \leq c_7\|\boldsymbol{\Sigma}^{1/2}\mathbf{s}\|_2^2 + c_8\frac{\log d}{n}\|\mathbf{s}\|_1,$$

where $\{c_i\}_{i=2}^8$ are universal constants.

The next lemma verifies the structural Lipschitz gradient condition in Condition 4.4.

**Lemma D.3.** Consider robust matrix sensing with objective loss function defined in section 4.2. There exist constants $C_0, C_1$ such that the following inequality holds with probability at least $1 - \exp(-C_0 d)$

$$|\langle \nabla_{\mathbf{X}} \mathcal{L}_n(\mathbf{X}^* + \mathbf{S}) - \nabla_{\mathbf{X}} \mathcal{L}_n(\mathbf{X}^* + \mathbf{S}^*), \mathbf{X} \rangle - \langle \mathbf{S} - \mathbf{S}^*, \mathbf{X} \rangle| \leq K \|\mathbf{X}\|_F \cdot \|\mathbf{S} - \mathbf{S}^*\|_F,$$
$$|\langle \nabla_{\mathbf{S}} \mathcal{L}_n(\mathbf{X} + \mathbf{S}^*) - \nabla_{\mathbf{S}} \mathcal{L}_n(\mathbf{X}^* + \mathbf{S}^*), \mathbf{S} \rangle - \langle \mathbf{X} - \mathbf{X}^*, \mathbf{S} \rangle| \leq K \|\mathbf{X} - \mathbf{X}^*\|_F \cdot \|\mathbf{S}\|_F,$$

for all low-rank matrices $\mathbf{X}, \mathbf{X}^*$ with rank at most $\widetilde{r}$ and all sparse matrices $\mathbf{S}, \mathbf{S}^*$ with sparsity at most $\widetilde{s}$, where $\widetilde{r}, \widetilde{s}$ are defined in Condition 4.2, and the structural Lipschitz gradient parameter $K = C_1 \sqrt{(rd + s) \log d / n}$.

The last lemma verifies the condition in Condition 4.5 for robust matrix sensing.

**Lemma D.4.** Consider robust matrix sensing, suppose each sensing matrix $\mathbf{A}_i$ is sampled independently from **I**-Gaussian ensemble and each element of noise vector $\boldsymbol{\epsilon}$ follows i.i.d. sub-Gaussian distribution with parameter $\nu$. Then we have the following inequalities hold with probability at least $1 - C_2/d$ in terms of spectral norm and infinity norm respectively

$$\left\| \frac{1}{n} \sum_{i=1}^n \epsilon_i \mathbf{A}_i \right\|_2 \leq C_3 \nu \sqrt{\frac{d}{n}} \quad \text{and} \quad \left\| \frac{1}{n} \sum_{i=1}^n \epsilon_i \mathbf{A}_i \right\|_{\infty,\infty} \leq C_4 \nu \sqrt{\frac{\log d}{n}},$$

where $C_2, C_3, C_4$ are universal constants.

Now, we are ready to prove Corollary 4.11.

*Proof of Corollary 4.11.* In order to prove Corollary 4.11, we only need to verify the restricted strong convex and smoothness conditions in Conditions 4.2 and 4.3, the structural Lipschitz gradient condition in Condition 4.4, and the condition in Condition 4.5.

Recall that we have the sample loss function for robust matrix sensing as $\mathcal{L}_n(\mathbf{X} + \mathbf{S}) := \|\mathbf{y} - \mathcal{A}_n(\mathbf{X} + \mathbf{S})\|_2^2/(2n)$. Therefore, for all given sparse matrices $\mathbf{S}$, we have the following holds for all matrices $\mathbf{X}_1, \mathbf{X}_2$ with rank at most $\widetilde{r}$

$$\mathcal{L}_n(\mathbf{X}_1 + \mathbf{S}) - \mathcal{L}_n(\mathbf{X}_2 + \mathbf{S}) - \langle \nabla_{\mathbf{X}} \mathcal{L}_n(\mathbf{X}_2 + \mathbf{S}), \mathbf{X}_2 - \mathbf{X}_1 \rangle = \frac{\|\mathcal{A}(\boldsymbol{\Delta})\|_2^2}{n},$$

where $\boldsymbol{\Delta} = \mathbf{X}_2 - \mathbf{X}_1$. According to Lemma D.1, if we have $n > c_1' \widetilde{r} d$, where $c'$ is some constants. Then, with probability at least $1 - \exp(-c_0 n)$, we have the restricted strong convexity and smoothness conditions in Condition 4.2 hold with parameter $\mu_1 = 4/9$ and $L_1 = 5/9$. In addition, for all given low-rank matrices $\mathbf{X}$, we have the following holds for all matrices $\mathbf{S}_1, \mathbf{S}_2$ with sparsity at most $\widetilde{s}$

$$\mathcal{L}_n(\mathbf{X} + \mathbf{S}_1) - \mathcal{L}_n(\mathbf{X} + \mathbf{S}_2) - \langle \nabla_{\mathbf{S}} \mathcal{L}_n(\mathbf{X} + \mathbf{S}_2), \mathbf{S}_2 - \mathbf{S}_1 \rangle = \frac{\|\mathcal{A}(\boldsymbol{\Delta})\|_2^2}{n},$$

where $\boldsymbol{\Delta} = \mathbf{S}_2 - \mathbf{S}_1$. Furthermore, we can obtain $\|\mathcal{A}(\boldsymbol{\Delta})\|_2^2 = \|\mathbf{A}\boldsymbol{\delta}\|_2^2$, where we have $\mathbf{A} \in \mathbb{R}^{n \times d_1 d_2}$ with each row $\mathbf{A}_{i*} = \text{vec}(\mathbf{A}_i)$, and $\boldsymbol{\delta} = \text{vec}(\boldsymbol{\Delta})$. Therefore, according to Lemma D.2, we have

$$c_1 \|\boldsymbol{\delta}\|_2^2 - c_2 \frac{\log d}{n} \|\boldsymbol{\delta}\|_1 \leq \frac{\|\mathbf{A}\boldsymbol{\delta}\|_2^2}{n} \leq c_3 \|\boldsymbol{\delta}\|_2^2 + c_4 \frac{\log d}{n} \|\boldsymbol{\delta}\|_1.$$

Thus provided that $n > c_5 \widetilde{s} \log d$, with probability at least $1 - c_3 \exp(-c_4 n)$, the restricted strong convexity and smoothness conditions in Condition 4.3 hold with parameters $\mu_2 = 4/9$ and $L_2 = 5/9$.

Next, according to Lemma D.3, with probability at least $1 - \exp(-C_0 d)$, we can establish the structural Lipschitz gradient condition in Condition 4.5 with parameter $K = C_1\sqrt{(rd+s)\log d/n}$.

Finally, we will verify the condition in Condition 4.5. By the definition of the objective loss function for robust matrix sensing, we have $\nabla_{\mathbf{X}} \mathcal{L}_n(\mathbf{X}^* + \mathbf{S}^*) = \sum_{i=1}^n \epsilon_i \mathbf{A}_i/n$ and $\nabla_{\mathbf{S}} \mathcal{L}_n(\mathbf{X}^* + \mathbf{S}^*) = \sum_{i=1}^n \epsilon_i \mathbf{A}_i/n$. Therefore, according to Lemma D.4, with probability at least $1 - C_2/d$, we can establish the condition in Condition 4.5 with parameters $\epsilon_1 = C_3 \nu \sqrt{d/n}$ and $\epsilon_2 = C_4 \nu \sqrt{\log d/n}$. This completes the proof. $\square$

## D.2 Proofs for Robust PCA

Note that since robust PCA under fully observed model is a special case of robust PCA under partially observed model, thus we just lay out the proofs of robust PCA under partially observed model. To prove the results of partially observed robust PCA, we need the following lemmas, which are essential to establish the restricted strong convexity and smoothness conditions in Conditions 4.2 and 4.3. Note that the following lemmas only work for robust PCA under noisy observation model.

**Lemma D.5.** (Negahban and Wainwright, 2012) There exist universal constants $\{c_i\}_{i=1}^4$ such that if the number of observations $n \geq c_1 rd \log d$, and the following condition is satisfied for all $\mathbf{\Delta} \in \mathbb{R}^{d_1 \times d_2}$

$$\sqrt{\frac{d_1 d_2}{r}} \frac{\|\mathbf{\Delta}\|_{\infty,\infty}}{\|\mathbf{\Delta}\|_F} \cdot \frac{\|\mathbf{\Delta}\|_*}{\|\mathbf{\Delta}\|_F} \leq \frac{1}{c_2}\sqrt{n/(d\log d)}, \tag{D.2}$$

we have, with probability at least $1 - c_3/d$, that the following holds

$$\left| \frac{\|\mathcal{A}(\mathbf{\Delta})\|_2}{\sqrt{n}} - \frac{\|\mathbf{\Delta}\|_F}{\sqrt{d_1 d_2}} \right| \leq \frac{1}{10} \frac{\|\mathbf{\Delta}\|_F}{\sqrt{d_1 d_2}} \left( 1 + \frac{c_4 \sqrt{d_1 d_2} \|\mathbf{\Delta}\|_{\infty,\infty}}{\sqrt{n} \|\mathbf{\Delta}\|_F} \right).$$

**Lemma D.6.** There exist universal constants $\{c_i\}_{i=1}^5$ such that as long as $n \geq c_1 \log d$, we have with probability at least $1 - c_2 \exp(-c_3 \log d)$ that

$$\left| \frac{\|\mathcal{A}(\mathbf{\Delta})\|_2}{\sqrt{n}} - \frac{\|\mathbf{\Delta}\|_F}{\sqrt{d_1 d_2}} \right| \leq \frac{1}{2} \frac{\|\mathbf{\Delta}\|_F}{\sqrt{d_1 d_2}} + \frac{c_5 \|\mathbf{\Delta}\|_{\infty,\infty}}{\sqrt{n}} \quad \text{for all} \quad \mathbf{\Delta} \in \mathcal{C}(n), \tag{D.3}$$

where we have the set $\mathcal{C}(n)$ as follows

$$\mathcal{C}(n) = \left\{ \mathbf{\Delta} \in \mathbb{R}^{d_1 \times d_2} \mid \frac{\|\mathbf{\Delta}\|_{1,1}}{\|\mathbf{\Delta}\|_F} \cdot \frac{\|\mathbf{\Delta}\|_{\infty,\infty}}{\|\mathbf{\Delta}\|_F} \leq c_4 \sqrt{\frac{n}{d_1 d_2 \log d}} \right\}.$$

The next lemma verifies the structural Lipschitz gradient condition in Condition 4.4.

**Lemma D.7.** Consider partially observed robust PCA with objective loss function defined in section 4.2. There exist constants $C_0, C_1$ such that the following inequality holds with probability at least $1 - \exp(-C_0 d)$

$$|\langle \nabla_{\mathbf{X}} \mathcal{L}_n(\mathbf{X}^* + \mathbf{S}) - \nabla_{\mathbf{X}} \mathcal{L}_n(\mathbf{X}^* + \mathbf{S}^*), \mathbf{X} \rangle - \langle \mathbf{S} - \mathbf{S}^*, \mathbf{X} \rangle| \leq K\|\mathbf{X}\|_F \cdot \|\mathbf{S} - \mathbf{S}^*\|_F,$$

$$|\langle \nabla_{\mathbf{S}} \mathcal{L}_n(\mathbf{X} + \mathbf{S}^*) - \nabla_{\mathbf{S}} \mathcal{L}_n(\mathbf{X}^* + \mathbf{S}^*), \mathbf{S} \rangle - \langle \mathbf{X} - \mathbf{X}^*, \mathbf{S} \rangle| \leq K\|\mathbf{X} - \mathbf{X}^*\|_F \cdot \|\mathbf{S}\|_F,$$

for all low-rank matrices $\mathbf{X}, \mathbf{X}^*$ with rank at most $\widetilde{r}$ and all sparse matrices $\mathbf{S}, \mathbf{S}^*$ with sparsity at most $\widetilde{s}$, where $\widetilde{r}, \widetilde{s}$ are defined in Condition 4.2, and the structural Lipschitz gradient parameter $K = C_1 \sqrt{(rd+s)\log d/n}$.

The last lemma verifies the condition in Condition 4.5 for partially observed robust PCA.

**Lemma D.8.** Consider partially observed robust PCA. If $\mathbf{A}_{jk} = \mathbf{e}_j \mathbf{e}_k^\top$ is uniformly distributed on $\Omega$, then for i.i.d. zero mean random variables $\epsilon_{jk}$ with variance $\nu^2$, we have the following inequalities hold with probability at least $1 - C_2/d$ in terms of spectral norm and infinity norm respectively

$$\left\| \frac{1}{p} \sum_{j,k \in \Omega} \epsilon_{jk} \mathbf{A}_{jk} \right\|_2 \leq C_3 \nu \sqrt{\frac{d \log d}{p}} \quad \text{and} \quad \left\| \frac{1}{p} \sum_{jk \in \Omega} \epsilon_{jk} \mathbf{A}_{jk} \right\|_{\infty,\infty} \leq C_4 \nu \sqrt{\frac{\log d}{p}},$$

where $C_2, C_3, C_4$ are universal constants, and $p = n/(d_1 d_2)$.

Now, we are ready to prove Corollary 4.15.

*Proof of Corollary 4.15.* To prove Corollary 4.15, we need to verify the restricted strong convexity and smoothness conditions in Conditions 4.2 and 4.3, the structural Lipschitz gradient condition in Condition 4.4, and the condition in Condition 4.5.

In the following discussion, we let $\mathbf{A}_{jk} = \mathbf{e}_j \mathbf{e}_k^\top$, where $\mathbf{e}_i, \mathbf{e}_j$ are basis vectors with $d_1$ and $d_2$ dimensions, and we let $\mathcal{A}$ be the corresponding transformation operator. In addition, let the number of observations to be $|\Omega| = n$. Therefore, the objective loss function for robust PCA in 4.2 can be rewritten as

$$\mathcal{L}_n(\mathbf{X} + \mathbf{S}) := \frac{1}{2p} \sum_{(j,k) \in \Omega} \left( \langle \mathbf{A}_{jk}, \mathbf{X} + \mathbf{S} \rangle - Y_{jk} \right)^2.$$

Therefore, for all given sparse matrices $\mathbf{S}$, we have the following holds for all matrices $\mathbf{X}_1, \mathbf{X}_2$ satisfying incoherence condition with rank at most $\widetilde{r}$

$$\mathcal{L}_n(\mathbf{X}_1 + \mathbf{S}) - \mathcal{L}_n(\mathbf{X}_2 + \mathbf{S}) - \langle \nabla_{\mathbf{X}} \mathcal{L}_n(\mathbf{X}_2 + \mathbf{S}), \mathbf{X}_2 - \mathbf{X}_1 \rangle = \frac{\|\mathcal{A}(\mathbf{\Delta})\|_2^2}{p},$$

where $\mathbf{\Delta} = \mathbf{X}_1 - \mathbf{X}_2$, and $p = n/(d_1 d_2)$. Now, we are ready to prove the restricted strong convexity and smoothness conditions in Condition 4.2.

**Case 1:** If $\mathbf{\Delta}$ not satisfies condition (D.2), then we have

$$\|\mathbf{\Delta}\|_F^2 \leq C_0 \left( \sqrt{d_1 d_2} \|\mathbf{\Delta}\|_\infty \right) \|\mathbf{\Delta}\|_* \sqrt{\frac{d \log d}{nr}}$$

$$\leq 2C_0 \alpha_1 \sqrt{d_1 d_2} \|\mathbf{\Delta}\|_* \sqrt{\frac{d \log d}{nr}}$$

$$\leq 2C_0 \alpha_1 \sqrt{2\widetilde{r} d_1 d_2} \|\mathbf{\Delta}\|_F \sqrt{\frac{d \log d}{nr}},$$

where $\widetilde{\alpha} = \alpha r / \sqrt{d_1 d_2}$ due to the incoherence condition of low rank matrices $\mathbf{X}_1$ and $\mathbf{X}_2$, and the last inequality comes from $\text{rank}(\mathbf{\Delta}) \leq 2\widetilde{r}$. Thus, by the definition of $\widetilde{r}$, we can obtain

$$\|\mathbf{\Delta}\|_F^2 \leq C_1 \alpha^2 \sigma_1^2 \frac{r^2 d \log d}{n}. \tag{D.4}$$

**Case 2:** If $\boldsymbol{\Delta}$ satisfies condition (D.2), then according to Lemma D.5, we have

$$\left|\frac{\|\mathcal{A}(\boldsymbol{\Delta})\|_2}{\sqrt{p}} - \|\boldsymbol{\Delta}\|_F\right| \leq \frac{\|\boldsymbol{\Delta}\|_F}{10}\left(1 + \frac{C_2\sqrt{d_1d_2}\|\boldsymbol{\Delta}\|_{\infty,\infty}}{\sqrt{n}\|\boldsymbol{\Delta}\|_F}\right).$$

Thus if $C_2\sqrt{d_1d_2}\|\boldsymbol{\Delta}\|_{\infty,\infty}/(\sqrt{n}\|\boldsymbol{\Delta}\|_F) \geq C_3$, we have

$$\|\boldsymbol{\Delta}\|_F^2 \leq C_4\frac{\widetilde{\alpha}^2}{p}.$$

Otherwise, if $C_2\sqrt{d_1d_2}\|\boldsymbol{\Delta}\|_{\infty,\infty}/(\sqrt{n}\|\boldsymbol{\Delta}\|_F) \leq C_3$, we have

$$\frac{8}{9}\|\boldsymbol{\Delta}\|_F^2 \leq \frac{\|\mathcal{A}(\boldsymbol{\Delta})\|_2^2}{p} \leq \frac{10}{9}\|\boldsymbol{\Delta}\|_F^2,$$

which gives us the restricted strong convexity and smoothness conditions in Condition 4.2 with parameters $\mu_1 = 8/9, L_1 = 10/9$.

Next, we prove the restricted strong convexity and smoothness conditions in Condition 4.3. For all given low-rank matrices $\mathbf{X}$, we have the following holds for all matrices $\mathbf{S}_1, \mathbf{S}_2$ with at most $\widetilde{s}$ nonzero entries and infinity norm bound $\alpha_1/\sqrt{d_1d_2}$

$$\mathcal{L}_n(\mathbf{X} + \mathbf{S}_1) - \mathcal{L}_n(\mathbf{X} + \mathbf{S}_2) - \langle\nabla_{\mathbf{S}}\mathcal{L}_n(\mathbf{X} + \mathbf{S}_2), \mathbf{S}_2 - \mathbf{S}_1\rangle = \frac{\|\mathcal{A}(\boldsymbol{\Delta})\|_2^2}{p},$$

where $\boldsymbol{\Delta} = \mathbf{S}_1 - \mathbf{S}_2$, and $p = n/(d_1d_2)$.

**Case 1:** If $\boldsymbol{\Delta} \notin \mathcal{C}(n)$, then we can get

$$\|\boldsymbol{\Delta}\|_F^2 \leq C_5\left(\sqrt{d_1d_2}\|\boldsymbol{\Delta}\|_{\infty,\infty}\right) \cdot \|\boldsymbol{\Delta}\|_{1,1}\sqrt{\frac{\log d}{n}} \leq 2C_5\alpha_1\|\boldsymbol{\Delta}\|_{1,1}\sqrt{\frac{\log d}{n}},$$

where the last inequality is due to the fact that $\|\boldsymbol{\Delta}\|_\infty = \|\mathbf{S}_1 - \mathbf{S}_2\|_\infty \leq 2\alpha_1/\sqrt{d_1d_2}$. Therefore, we can obtain

$$\|\boldsymbol{\Delta}\|_F^2 \leq 2C_5\sqrt{2\widetilde{s}}\alpha_1\|\boldsymbol{\Delta}\|_F\sqrt{\frac{\log d}{n}},$$

where the inequality holds because $\boldsymbol{\Delta}$ has at most $2\widetilde{s}$ nonzero entries. Therefore, by the definition of $\widetilde{s}$, we have

$$\|\boldsymbol{\Delta}\|_F^2 \leq C_6\alpha_1^2\frac{s\log d}{n}. \tag{D.5}$$

**Case 2:** If $\boldsymbol{\Delta} \in \mathcal{C}(n)$, we have

$$\left|\frac{\|\mathcal{A}(\boldsymbol{\Delta})\|_2}{\sqrt{p}} - \|\boldsymbol{\Delta}\|_F\right| \leq \frac{1}{2}\|\boldsymbol{\Delta}\|_F + \frac{c_5\sqrt{d_1d_2}\|\boldsymbol{\Delta}\|_{\infty,\infty}}{\sqrt{n}\|\boldsymbol{\Delta}\|_F},$$

If $\sqrt{n}\|\boldsymbol{\Delta}\|_F \leq C_7\sqrt{d_1d_2}\|\boldsymbol{\Delta}\|_{\infty,\infty}$, we can obtain $\|\boldsymbol{\Delta}\|_F^2 \leq C_7'\alpha_1^2/n$. Otherwise, if we have $\sqrt{n}\|\boldsymbol{\Delta}\|_F \geq C_7\sqrt{d_1d_2}\|\boldsymbol{\Delta}\|_{\infty,\infty}$, according to Lemma D.6, we obtain

$$\frac{8}{9}\|\boldsymbol{\Delta}\|_F^2 \leq \frac{\|\mathcal{A}(\boldsymbol{\Delta})\|_2^2}{p} \leq \frac{10}{9}\|\boldsymbol{\Delta}\|_F^2,$$

which implies the restricted strong convexity and smoothness conditions in Condition 4.3 hold with parameters $\mu_2 = 8/9, L_2 = 10/9$.

Next, according to Lemma D.7, with probability at least $1 - \exp(-C_0 d)$, we can establish the structural Lipschitz gradient condition in Condition 4.5 with parameter $K = C_1 \sqrt{(rd + s) \log d/n}$.

Finally, we verify the condition in Condition 4.5. By the definition of the objective loss function for robust PCA, we have $\nabla_{\mathbf{X}} \mathcal{L}_n(\mathbf{X}^* + \mathbf{S}^*) = \sum_{j,k \in \Omega} \epsilon_{jk} \mathbf{A}_{jk}/p$ and $\nabla_{\mathbf{S}} \mathcal{L}_n(\mathbf{X}^* + \mathbf{S}^*) = \sum_{j,k \in \Omega} \epsilon_{jk} \mathbf{A}_{jk}/p$, where $\epsilon_{jk}$ are i.i.d. Gaussian variables with variance $\nu^2/(d_1 d_2)$. Therefore, according to Lemma D.8, with probability at least $1 - C_8'/d$, we have $\|\sum_{j,k \in \Omega} \epsilon_{jk} \mathbf{A}_{jk}/p\|_2^2 \leq C_8 \nu^2 d \log d/n$. In addition, we have $\|\sum_{jk \in \Omega} \epsilon_{jk} \mathbf{A}_{jk}/p\|_{\infty,\infty}^2 \leq C_9 \nu^2 \log d/n$. Furthermore, we have additional estimation error bounds (D.4) and (D.5) when we derive the restricted strong convexity and smoothness conditions. Therefore, we can establish the condition in Condition 4.5 with parameters $\epsilon_1^2 = C_8 \max\{\alpha_1^2, \nu^2\} d/n$ and $\epsilon_2^2 = C_9 \max\{\alpha_1^2, \nu^2\} \log d/n$. This completes the proof. $\square$

# E   Proofs of Technical Lemmas in Appendix D

## E.1   Proof of Lemma D.3

*Proof.* In order to verify the structural Lipschitz gradient condition, we need to make use of the Bernstein-type inequality for sub-exponential random variables in Vershynin (2010) as well as the corresponding covering arguments for low-rank and sparse structures, respectively.

By the definition of the objective loss function of matrix sensing, we have

$$\langle \nabla_{\mathbf{X}} \mathcal{L}_n(\mathbf{X}^* + \mathbf{S}) - \nabla_{\mathbf{X}} \mathcal{L}_n(\mathbf{X}^* + \mathbf{S}^*), \mathbf{X} \rangle = \frac{1}{n} \sum_{i=1}^n \langle \mathbf{A}_i, \mathbf{S} - \mathbf{S}^* \rangle \langle \mathbf{A}_i, \mathbf{X} \rangle = \frac{1}{n} \sum_{i=1}^n Y_i,$$

where $Y_i = \langle \mathbf{A}_i, \mathbf{S} - \mathbf{S}^* \rangle \langle \mathbf{A}_i, \mathbf{X} \rangle$. Note that $\langle \mathbf{A}_i, \mathbf{S} - \mathbf{S}^* \rangle$, $\langle \mathbf{A}_i, \mathbf{X} \rangle$ follow i.i.d. normal distribution $N(0, \|\mathbf{S} - \mathbf{S}^*\|_F^2)$ and $N(0, \|\mathbf{X}\|_F^2)$ respectively. Thus $Y_i$ follows i.i.d. chi-square distribution which is also sub-exponential. Besides, $\mathbb{E}(Y_i) = \langle \mathbf{S} - \mathbf{S}^*, \mathbf{X} \rangle$, and we have

$$\|Y_i - \mathbb{E}[Y_i]\|_{\psi_1} \leq 2\|Y_i\|_{\psi_1} \leq 2\|\langle \mathbf{A}_i, \mathbf{S} - \mathbf{S}^* \rangle\|_{\psi_2} \cdot \|\langle \mathbf{A}_i, \mathbf{X} \rangle\|_{\psi_2} \leq 2C^2 \|\mathbf{S} - \mathbf{S}^*\|_F \cdot \|\mathbf{X}\|_F = \lambda,$$

where $C$ is a universal constant. Thus, by applying Proposition 5.16 in Vershynin (2010), for $Y_i - \mathbb{E}[Y_i]$, we obtain

$$\mathbb{P}\left\{\left|\sum_{i=1}^n \frac{1}{n}(Y_i - \mathbb{E}[Y_i])\right| \geq t\right\} \leq 2\exp\left[-c\min\left(\frac{nt^2}{\lambda^2}, \frac{nt}{\lambda}\right)\right].$$

According the covering argument for low-rank matrices Lemma 3.1 in Candes and Plan (2011) and covering number for sparse matrices in Vershynin (2009), we have

$$\mathbb{P}\left\{\sup_{(\mathbf{S}-\mathbf{S}^*)\in\mathcal{N}_\epsilon^{cs},\mathbf{X}\in\mathcal{N}_\epsilon^{3r}}\left|\frac{1}{n}\sum_{i=1}^n\langle\mathbf{A}_i,\mathbf{S}-\mathbf{S}^*\rangle\langle\mathbf{A}_i,\mathbf{X}\rangle-\langle\mathbf{S}-\mathbf{S}^*,\mathbf{X}\rangle\right|\geq t\right\}$$

$$\leq 2|\mathcal{N}_\epsilon^{cs}||\mathcal{N}_\epsilon^{3r}|\exp\left[-c_1\min\left(\frac{nt^2}{\lambda^2},\frac{nt}{\lambda}\right)\right]$$

$$\leq 2\left(\frac{9}{\epsilon}\right)^{(d_1+d_2+1)3r}\cdot\left(\frac{c_2d_1d_2}{cs\epsilon}\right)^{cs}\cdot\exp\left[-c_1\min\left(\frac{nt^2}{\lambda^2},\frac{nt}{\lambda}\right)\right]$$

$$\leq\exp\left[c_3\big(rd\log(1/\epsilon)+s\max\{\log d,\log(1/\epsilon)\}\big)-c_1\min\left(\frac{nt^2}{\lambda^2},\frac{nt}{\lambda}\right)\right]\leq\exp(-c'd), \quad\text{(E.1)}$$

where $c_1,c_2,c_3$ are constants, $\lambda=2C^2$, and the first inequality follows from union bound, the second inequality is due to the covering arguments, and the last inequality holds by setting $t=c_4\sqrt{(rd+s)\log d}/\sqrt{n}$. Besides, note that for any $\mathbf{X}\in\mathcal{M}_{3r}$, $\mathbf{S}\in\mathbf{S}^*+\mathcal{M}_{cs}$, there exists $\mathbf{X}_1\in\mathcal{N}_\epsilon^{3r}$, $\mathbf{S}_1\in\mathbf{S}^*+\mathcal{N}_\epsilon^{cs}$ such that $\|\mathbf{X}-\mathbf{X}_1\|_F\leq\epsilon$ and $\|\mathbf{S}-\mathbf{S}_1\|_F\leq\epsilon$. Thus, we have

$$\left|\frac{1}{n}\sum_{i=1}^n\langle\mathbf{A}_i,\mathbf{S}-\mathbf{S}^*\rangle\langle\mathbf{A}_i,\mathbf{X}\rangle-\frac{1}{n}\sum_{i=1}^n\langle\mathbf{A}_i,\mathbf{S}_1-\mathbf{S}^*\rangle\langle\mathbf{A}_i,\mathbf{X}_1\rangle\right|$$

$$\leq\left|\frac{1}{n}\sum_{i=1}^n\langle\mathbf{A}_i,\mathbf{S}-\mathbf{S}^*\rangle\langle\mathbf{A}_i,\mathbf{X}-\mathbf{X}_1\rangle\right|+\left|\frac{1}{n}\sum_{i=1}^n\langle\mathbf{A}_i,\mathbf{S}-\mathbf{S}_1\rangle\langle\mathbf{A}_i,\mathbf{X}_1\rangle\right|$$

$$\leq\sqrt{L_1L_2}\|\mathbf{S}-\mathbf{S}^*\|_F\cdot\|\mathbf{X}-\mathbf{X}_1\|_F+\sqrt{L_1L_2}\|\mathbf{S}-\mathbf{S}_1\|_F\cdot\|\mathbf{X}_1\|_F\leq 2\epsilon\sqrt{L_1L_2}, \quad\text{(E.2)}$$

where the first inequality holds because of triangle inequality, and the second inequality follows from the restricted strong smoothness condition for both low-rank and sparse structures. Similarly, we have

$$|\langle\mathbf{S}-\mathbf{S}^*,\mathbf{X}\rangle-\langle\mathbf{S}_1-\mathbf{S}^*,\mathbf{X}_1\rangle|\leq\|\mathbf{S}-\mathbf{S}^*\|_F\cdot\|\mathbf{X}-\mathbf{X}_1\|_F+\|\mathbf{S}-\mathbf{S}_1\|_F\cdot\|\mathbf{X}_1\|_F\leq 2\epsilon, \quad\text{(E.3)}$$

Therefore, combining (E.1), (E.2) and (E.3), by triangle inequality, we obtain

$$\sup_{(\mathbf{S}-\mathbf{S}^*)\in\mathcal{M}_{cs},\mathbf{X}\in\mathcal{M}_{3r}}\left|\frac{1}{n}\sum_{i=1}^n\langle\mathbf{A}_i,\mathbf{S}-\mathbf{S}^*\rangle\langle\mathbf{A}_i,\mathbf{X}\rangle-\langle\mathbf{S}-\mathbf{S}^*,\mathbf{X}\rangle\right|\leq t+2\epsilon\sqrt{L_1L_2}+2\epsilon,$$

with probability at least $1-\exp(-c'd)$. We establish the incoherence condition by setting $\epsilon=t/(2\sqrt{L_1L_2}+2)$ in (E.3). By similar techniques, we can prove the second inequality in Lemma D.3. Note that we obtain $K=C\sqrt{(rd+s)\log d/n}$ in Lemma D.3. $\square$

### E.2 Proof of Lemma D.4

*Proof.* The first inequality in Lemma D.4 has been established in Negahban and Wainwright (2011) Lemma 6. We provided the second inequality using Bernstein-type inequality and Union Bound. Recall that, we have

$$\left\|\frac{1}{n}\sum_{i=1}^n\epsilon_i\mathbf{A}_i\right\|_{\infty,\infty}=\max_{j,k}\left|\frac{1}{n}\sum_{i=1}^n\epsilon_iA_{jk}^i\right|=\max_{j,k}\left|\frac{1}{n}\sum_{i=1}^nZ_{jk}^i\right|,$$

where we let $Z_{jk}^i = \epsilon_i A_{jk}^i$. Since $Z_{jk}^i$ are independent centered sub-exponential random variables for $i = 1, \ldots, n$ with $\max_i \|Z_{jk}^i\|_{\psi_1} \leq 2 \max_i \|\epsilon_i\|_{\psi_2} \cdot \|A_{jk}^i\|_{\psi_2} \leq 2\nu$, according to Proposition 5.16 in Vershynin (2010), we have

$$\mathbb{P}\left\{\left|\frac{1}{n}\sum_{i=1}^n Z_{jk}^i\right| \geq t\right\} \leq 2\exp\left(-C'\frac{nt^2}{\nu^2}\right).$$

Thus by union bound, we have

$$\mathbb{P}\left\{\max_{j,k}\left|\frac{1}{n}\sum_{i=1}^n Z_{jk}^i\right| \geq t\right\} \leq 2d_1 d_2 \exp\left(-C'\frac{nt^2}{\nu^2}\right).$$

Let $t = C_2\nu\sqrt{\log d/n}$, we have the following inequality holds with probability at least $1 - C/d$

$$\left\|\frac{1}{n}\sum_{i=1}^n \epsilon_i \mathbf{A}_i\right\|_{\infty,\infty} \leq C_2\nu\sqrt{\frac{\log d}{n}}.$$

□

Thus, we complete the proof.

### E.3 Proof of Lemma D.6

The proof of this lemma is inspired by the proof of Theorem 1 in Negahban and Wainwright (2012), and we extended it to the sparse case. In order to prove Lemma D.6, we only need to prove the inequality (D.3) holds with high probability. Specifically, we consider the following event

$$E = \left\{\exists\, \mathbf{S} \in \mathcal{C}(n) \mid \left|\frac{\|\mathcal{A}(\mathbf{S})\|_2}{\sqrt{n}} - \frac{\|\mathbf{S}\|_F}{\sqrt{d_1 d_2}}\right| \leq \frac{1}{2}\frac{\|\mathbf{S}\|_F}{\sqrt{d_1 d_2}} + \frac{32\|\mathbf{S}\|_{\infty,\infty}}{\sqrt{n}}\right\}.$$

Therefore, we want to establish the probability for event $E$, and we need the following lemmas.

**Lemma E.1.** Consider the robust PCA under observation model in section 4.2, for $\ell = 1, 2, \ldots$, we have

$$\mathbb{P}(E_\ell) \leq \exp(-c_1 n \alpha^{2\ell} \mu^2),$$

where we have

$$E_\ell := \left\{\exists\, \mathbf{S} \in \mathcal{B}'(\alpha^\ell \mu) \mid \left|\frac{\|\mathcal{A}(\mathbf{S})\|_2}{\sqrt{n}} - \frac{\|\mathbf{S}\|_F}{\sqrt{d_1 d_2}}\right| \geq \frac{5}{12}\frac{\alpha^\ell \mu}{\sqrt{d_1 d_2}} + \frac{32\|\mathbf{S}\|_{\infty,\infty}}{\sqrt{n}}\right\},$$

and

$$\mathcal{B}'(\alpha^\ell \mu) = \left\{\mathbf{S} \in \mathcal{C}(n, s) \mid \frac{\|\mathbf{S}\|_F}{\sqrt{d_1 d_2}} \leq \frac{\alpha^\ell \mu}{\sqrt{d_1 d_2}}\right\}.$$

*Proof of Lemma D.6.* The reminder of this proof is to derive the probability of the event $E$. In order to establish the probability of the event $E$, we make use of the peeling argument of the Frobenius norm $\|\mathbf{S}\|_F$. Let $\mu = c\sqrt{\log d/n}$, and $\alpha = 6/5$. For $\ell = 1, 2, \ldots$, we define the sets

$$\mathcal{S}_\ell := \left\{ \mathbf{S} \in \mathcal{C}(n, s) \;\Big|\; \frac{\alpha^{\ell-1}\mu}{\sqrt{d_1 d_2}} \leq \frac{\|\mathbf{S}\|_F}{\sqrt{d_1 d_2}} \leq \frac{\alpha^\ell \mu}{\sqrt{d_1 d_2}} \right\}.$$

Therefore, if the event $E$ holds, there exist a matrix $\mathbf{S}$ that must belongs to $\mathcal{S}_\ell$ for some $\ell = 1, 2, \ldots$ such that

$$\left| \frac{\|\mathcal{A}(\mathbf{S})\|_2}{\sqrt{n}} - \frac{\|\mathbf{S}\|_F}{\sqrt{d_1 d_2}} \right| \geq \frac{1}{2} \frac{\|\mathbf{S}\|_F}{\sqrt{d_1 d_2}} + \frac{32 \|\mathbf{S}\|_{\infty,\infty}}{\sqrt{n}} \geq \frac{1}{2} \frac{\alpha^{\ell-1}\mu}{\sqrt{d_1 d_2}} + \frac{32\|\mathbf{S}\|_{\infty,\infty}}{\sqrt{n}} = \frac{5}{12} \frac{\alpha^\ell \mu}{\sqrt{d_1 d_2}} + \frac{32\|\mathbf{S}\|_{\infty,\infty}}{\sqrt{n}},$$

where the last equality is due to the fact that $\alpha = 6/5$.

Next, consider following events $E_\ell$, for $\ell = 1, 2, \ldots$

$$E_\ell := \left\{ \exists\, \mathbf{S} \in \mathcal{B}'(\alpha^\ell \mu) \;\Big|\; \left| \frac{\|\mathcal{A}(\mathbf{S})\|_2}{\sqrt{n}} - \frac{\|\mathbf{S}\|_F}{\sqrt{d_1 d_2}} \right| \geq \frac{5}{12} \frac{\alpha^\ell \mu}{\sqrt{d_1 d_2}} + \frac{32\|\mathbf{S}\|_{\infty,\infty}}{\sqrt{n}} \right\},$$

where we have the constraint set

$$\mathcal{B}'(\alpha^\ell \mu) = \left\{ \mathbf{S} \in \mathcal{C}(n, s) \;\Big|\; \frac{\|\mathbf{S}\|_F}{\sqrt{d_1 d_2}} \leq \frac{\alpha^\ell \mu}{\sqrt{d_1 d_2}} \right\}.$$

Since $\mathbf{S} \in \mathcal{S}_\ell$ implies that $\mathbf{S} \in \mathcal{B}(\alpha^\ell \mu)$, we can get $E \subset \bigcup_{\ell=1}^\infty E_\ell$. Therefore, we only need to upper bound the probability $\mathbb{P}(\bigcup_{\ell=1}^\infty E_\ell)$. In order to do so, we need upper bound the probability $\mathbb{P}(E_\ell)$. According to Lemma E.1, we have $\mathbb{P}(E_\ell) \leq \exp(-c_1 n \alpha^{2\ell} \mu^2)$. Therefore, we can obtain

$$\mathbb{P}(E) \leq \mathbb{P}\left( \bigcup_{\ell=1}^\infty E_\ell \right) \leq \sum_{\ell=1}^\infty \mathbb{P}(E_\ell) \leq \sum_{\ell=1}^\infty \exp(-c_1 n \alpha^{2\ell} \mu^2).$$

Thus according to the inequality $a \leq e^a$, we can obtain

$$\mathbb{P}(E) \leq \sum_{\ell=1}^\infty \exp\left( -2\ell c_1 n \mu^2 \log \alpha \right) \leq \frac{\exp\left( -2c_1 n \mu^2 \log \alpha \right)}{1 - \exp\left( -2c_1 n \mu^2 \log \alpha \right)} = \frac{\exp(-c_2 \log d)}{1 - \exp(-c_2 \log d)},$$

where the last equality comes from the definition $\mu = c\sqrt{\log d/n}$, and this implies $\mathbb{P}(E) \leq c_3 \exp(-c_2 \log d)$. □

### E.4 Proof of Lemma D.7

*Proof.* The proof of this Lemma is similar to the proof of Lemma D.3, using Proposition 5.16 in Vershynin (2010) and covering number argument, with probability at least $1 - \exp(-c_1 d)$, we can obtain the restricted Lipschitz gradient condition in Condition 4.4 with parameter $K = c_2 \sqrt{(rd + s) \log d/n}$. □

## E.5 Proof of Lemma D.8

*Proof.* For the first inequality in Lemma D.8, it has been established in Negahban and Wainwright (2012) Proposition 1. For the second inequality in Lemma D.8, we use the similar proof as in the proof of Lemma D.4. By proposition 5.16 in Vershynin (2010) and union bound, with probability at least $1 - C/d$, we can obtain the required inequality. $\square$

## F Proof of Auxiliary Lemmas in Appendix E

In order to prove Lemma E.1, we need the following lemmas.

**Lemma F.1.** We have the following holds with probability at least $1 - C\exp(-C_1 nD^2)$

$$\max_{k=1,\ldots,N(D/8)} \left| \frac{\|\mathcal{A}(\mathbf{S}^k)\|_2}{\sqrt{n}} - \frac{\|\mathbf{S}^k\|_F}{\sqrt{d_1 d_2}} \right| \leq \frac{D}{8\sqrt{d_1 d_2}} + \frac{32\|\mathbf{S}\|_{\infty,\infty}}{\sqrt{n}}.$$

**Lemma F.2.** We have the following holds

$$\sup_{\mathbf{\Delta} \in \mathcal{D}(\delta)} \frac{\|\mathcal{A}(\mathbf{\Delta})\|_2}{\sqrt{n}} \leq \frac{D}{2\sqrt{d_1 d_2}},$$

where we have

$$\mathcal{D}(\delta) := \{\mathbf{\Delta} \in \mathbb{R}^{d_1 \times d_2} \mid \|\mathbf{\Delta}\|_F \leq \delta, \ \|\mathbf{\Delta}\|_{1,1} \leq 2\rho(D), \ \|\mathbf{\Delta}\|_0 \leq 2\widetilde{s}\},$$

and $\rho(D) \leq D^2/(c\sqrt{\log d/n})$.

*Proof of Lemma E.1.* The proof of this lemma is inspired by the proof of Lemma 3 in Negahban and Wainwright (2012). Note that since the definition of the constraint set $\mathcal{C}(n)$ and $E$ is invariant to rescaling of $\mathbf{S}$, we can assume w.l.o.g. that $\|\mathbf{S}\|_{\infty,\infty} = 1/\sqrt{d_1 d_2}$. Therefore, it is equivalent to consider following events

$$E_\ell := \left\{ \exists \mathbf{S} \in \mathcal{B}(\alpha^\ell \mu) \mid \left| \frac{\|\mathcal{A}(\mathbf{S})\|_2}{\sqrt{n}} - \frac{\|\mathbf{S}\|_F}{\sqrt{d_1 d_2}} \right| \geq \frac{3\alpha^\ell \mu}{4\sqrt{d_1 d_2}} + \frac{32}{\sqrt{n d_1 d_2}} \right\}$$

where we have the constraint set

$$\mathcal{B}(\alpha^\ell \mu) = \left\{ \mathbf{S} \in \mathcal{C}(n,s) \mid \|\mathbf{S}\|_{\infty,\infty} \leq \frac{1}{\sqrt{d_1 d_2}}, \ \frac{\|\mathbf{S}\|_F}{\sqrt{d_1 d_2}} \leq \frac{\alpha^\ell \mu}{\sqrt{d_1 d_2}}, \ \|\mathbf{S}\|_{1,1} \leq \rho(\alpha^\ell \mu) \right\},$$

where $\rho(\alpha^\ell \mu) \leq (\alpha^\ell \mu)^2/(c\sqrt{\log d/n})$. Define

$$Z_n(\alpha^\ell \mu) := \sup_{\mathbf{S} \in \mathcal{B}(\alpha^\ell \mu)} \left| \frac{\|\mathcal{A}(\mathbf{S})\|_2}{\sqrt{n}} - \frac{\|\mathbf{S}\|_F}{\sqrt{d_1 d_2}} \right|.$$

For simplicity, we use $D$ to denote $\alpha^\ell \mu$ in the following discussion. Therefore, we just need to prove the following probability bound

$$\mathbb{P}\left( Z_n(D) \geq \frac{3D}{4\sqrt{d_1 d_2}} + \frac{32}{\sqrt{n d_1 d_2}} \right) \leq c_3 \exp(-c_4 n D^2).$$

Suppose $\mathbf{S}^1, \ldots, \mathbf{S}^{N(\delta)}$ are a $\delta$-covering of $\mathcal{B}(D)$ in terms of Frobenius norm. Therefore, for any $\mathbf{S} \in \mathcal{B}(D)$, there exist a matrix $\mathbf{\Delta} \in \mathbb{R}^{d_1 \times d_2}$ and some index $k \in \{1, \ldots, N(\delta)\}$ satisfying $\mathbf{S} = \mathbf{S}^k + \mathbf{\Delta}$, where $\|\mathbf{\Delta}\|_F \leq \delta$. Thus we can obtain

$$\frac{\|\mathcal{A}(\mathbf{S})\|_2}{\sqrt{n}} - \frac{\|\mathbf{S}\|_F}{\sqrt{d_1 d_2}} = \frac{\|\mathcal{A}(\mathbf{S}^k + \mathbf{\Delta})\|_2}{\sqrt{n}} - \frac{\|\mathbf{S}^k + \mathbf{\Delta}\|_F}{\sqrt{d_1 d_2}}$$

$$\leq \frac{\|\mathcal{A}(\mathbf{S}^k)\|_2}{\sqrt{n}} + \frac{\|\mathcal{A}(\mathbf{\Delta})\|_2}{\sqrt{n}} - \frac{\|\mathbf{S}^k\|_F}{\sqrt{d_1 d_2}} + \frac{\|\mathbf{\Delta}\|_F}{\sqrt{d_1 d_2}}$$

$$\leq \left| \frac{\|\mathcal{A}(\mathbf{S}^k)\|_2}{\sqrt{n}} - \frac{\|\mathbf{S}^k\|_F}{\sqrt{d_1 d_2}} \right| + \frac{\|\mathcal{A}(\mathbf{\Delta})\|_2}{\sqrt{n}} + \frac{\delta}{\sqrt{d_1 d_2}}.$$

In addition we can get

$$\left| \frac{\|\mathcal{A}(\mathbf{S})\|_2}{\sqrt{n}} - \frac{\|\mathbf{S}\|_F}{\sqrt{d_1 d_2}} \right| \leq \left| \frac{\|\mathcal{A}(\mathbf{S}^k)\|_2}{\sqrt{n}} - \frac{\|\mathbf{S}^k\|_F}{\sqrt{d_1 d_2}} \right| + \frac{\|\mathcal{A}(\mathbf{\Delta})\|_2}{\sqrt{n}} + \frac{\delta}{\sqrt{d_1 d_2}}.$$

Therefore, we have

$$Z_n(D) \leq \frac{\delta}{\sqrt{d_1 d_2}} + \max_{k=1\ldots,N(\delta)} \left| \frac{\|\mathcal{A}(\mathbf{S}^k)\|_2}{\sqrt{n}} - \frac{\|\mathbf{S}^k\|_F}{\sqrt{d_1 d_2}} \right| + \sup_{\mathbf{\Delta} \in \mathcal{D}(\delta)} \frac{\|\mathcal{A}(\mathbf{\Delta})\|_2}{\sqrt{n}}, \tag{F.1}$$

where we have $\mathcal{D}(\delta) := \{\mathbf{\Delta} \in \mathbb{R}^{d_1 \times d_2} \mid \|\mathbf{\Delta}\|_F \leq \delta, \|\mathbf{\Delta}\|_{1,1} \leq 2\rho(D), \|\mathbf{\Delta}\|_0 \leq 2cs\}$. We establish the high probability bound of (F.1) with $\delta = D/8$. First, according to Lemma F.1, we have

$$\max_{k=1\ldots,N(D/8)} \left| \frac{\|\mathcal{A}(\mathbf{S}^k)\|_2}{\sqrt{n}} - \frac{\|\mathbf{S}^k\|_F}{\sqrt{d_1 d_2}} \right| \leq \frac{D}{8\sqrt{d_1 d_2}} + \frac{32\|\mathbf{S}\|_{\infty,\infty}}{\sqrt{n}}, \tag{F.2}$$

holds with probability at least $1 - c\exp(-c_1 n D^2)$.

Next, according to Lemma F.2, we have

$$\sup_{\mathbf{\Delta} \in \mathcal{D}(\delta)} \frac{\|\mathcal{A}(\mathbf{\Delta})\|_2}{\sqrt{n}} \leq \frac{D}{2\sqrt{d_1 d_2}}, \tag{F.3}$$

holds with probability at least $1 - c_2 \exp(-c_3 n D^2)$.

Therefore, combining (F.2) and (F.3), we can get

$$Z_n(D) \leq \frac{D}{8\sqrt{d_1 d_2}} + \frac{D}{8\sqrt{d_1 d_2}} + \frac{D}{2\sqrt{d_1 d_2}} + \frac{32\|\mathbf{S}\|_{\infty,\infty}}{\sqrt{n}} \leq \frac{3D}{4\sqrt{d_1 d_2}} + \frac{32}{\sqrt{n d_1 d_2}},$$

holds with probability at least $1 - c_3 \exp(-c_4 n D^2)$, and the last inequality comes from that $\|\mathbf{S}\|_{\infty,\infty} \leq 1/\sqrt{d_1 d_2}$. $\square$

# G  Proofs of Auxiliary Lemmas in Appendix F

## G.1  Proof of Lemma F.1

*Proof.* First, we prove that for a fixed matrix $\mathbf{S}$, we have the following inequality holds

$$\mathbb{P}\left( \left| \frac{\|\mathcal{A}(\mathbf{S})\|_2}{\sqrt{n}} - \frac{\|\mathbf{S}\|_F}{\sqrt{d_1 d_2}} \right| \geq \frac{\delta}{\sqrt{d_1 d_2}} + \frac{32\|\mathbf{S}\|_{\infty,\infty}}{\sqrt{n}} \right) \leq C \exp(-C_1 n \delta^2).$$

Since we have
$$\frac{\|\mathcal{A}(\mathbf{S})\|_2}{\sqrt{n}} = \frac{1}{\sqrt{n}}\sqrt{\sum_{j,k\in\Omega}\langle\mathbf{A}_{jk},\mathbf{S}\rangle^2} = \frac{1}{\sqrt{n}}\sup_{\|\mathbf{w}\|_2=1}\sum_{j,k\in\Omega}w_i\langle\mathbf{A}_{jk},\mathbf{S}\rangle,$$

we consider
$$\sqrt{d_1d_2}\frac{\|\mathcal{A}(\mathbf{S})\|_2}{\sqrt{n}} = \frac{\sqrt{d_1d_2}}{\sqrt{n}}\sqrt{\sum_{j,k\in\Omega}\langle\mathbf{A}_{jk},\mathbf{S}\rangle^2} = \frac{1}{\sqrt{n}}\sup_{\|\mathbf{w}\|_2=1}\sum_{j,k\in\Omega}w_{jk}\langle\sqrt{d_1d_2}\mathbf{A}_{jk},\mathbf{S}\rangle$$
$$= \frac{1}{\sqrt{n}}\sup_{\|\mathbf{w}\|_2=1}\sum_{j,k\in\Omega}w_{jk}Y_{jk},$$

where we have the random variables $Y_{jk}$ satisfying $|Y_{jk}| = |\langle\sqrt{d_1d_2}\mathbf{A}_{jk},\mathbf{S}\rangle| \leq \sqrt{d_1d_2}\|\mathbf{S}\|_{\infty,\infty} = 1$. Therefore, according to lemma H.1, we have

$$\mathbb{P}\left(\left|\frac{\|\mathcal{A}(\mathbf{S})\|_2}{\sqrt{n}} - \mathbb{E}\left[\frac{\|\mathcal{A}(\mathbf{S})\|_2}{\sqrt{n}}\right]\right| \geq \frac{\delta}{\sqrt{d_1d_2}} + \frac{16}{\sqrt{nd_1d_2}}\right) \leq C\exp(-C_1n\delta^2). \tag{G.1}$$

In addition, we have

$$\left|\frac{\|\mathbf{S}\|_F}{\sqrt{d_1d_2}} - \mathbb{E}\left[\frac{\|\mathcal{A}(\mathbf{S})\|_2}{\sqrt{n}}\right]\right| = \left|\sqrt{\mathbb{E}\left[\frac{\|\mathcal{A}(\mathbf{S})\|_2^2}{n}\right]} - \mathbb{E}\left[\frac{\|\mathcal{A}(\mathbf{S})\|_2}{\sqrt{n}}\right]\right|$$
$$\leq \sqrt{\mathbb{E}\left[\frac{\|\mathcal{A}(\mathbf{S})\|_2^2}{n}\right] - \mathbb{E}\left(\left[\frac{\|\mathcal{A}(\mathbf{S})\|_2}{\sqrt{n}}\right]\right)^2} \leq \frac{16}{\sqrt{nd_1d_2}}. \tag{G.2}$$

Therefore, combining (G.1) and (G.2), we can obtain
$$\mathbb{P}\left(\left|\frac{\|\mathcal{A}(\mathbf{S})\|_2}{\sqrt{n}} - \frac{\|\mathbf{S}\|_F}{\sqrt{d_1d_2}}\right| \geq \frac{\delta}{d_1d_2} + \frac{32}{\sqrt{nd_1d_2}}\right) \leq C\exp(-C_1n\delta^2).$$

Next, according to Lemma 4 in Negahban and Wainwright (2012), there exists a $\delta$-covering of $\mathcal{B}(D)$ such that

$$\log N(\delta) \leq C_3(\rho(D)/\delta)^2\log d.$$

Therefore, we can get
$$\mathbb{P}\left[\max_{k=1,\ldots,N(D/8)}\left|\frac{\|\mathcal{A}(\mathbf{S}^k)\|_2}{\sqrt{n}} - \frac{\|\mathbf{S}^k\|_F}{\sqrt{d_1d_2}}\right| \geq \frac{\delta}{\sqrt{d_1d_2}} + \frac{32}{\sqrt{nd_1d_2}}\right] \leq C\exp(-C_1n\delta^2 + \log N(\delta))$$
$$\leq C\exp(-C_1n\delta^2 + C_3(\rho(D)/\delta)^2\log d).$$

Since we have $\delta = D/8$ and $\rho(D) = C_4D^2/\sqrt{\log d/n}$, we can obtain

$$\mathbb{P}\left[\max_{k=1,\ldots,N(D/8)}\left|\frac{\|\mathcal{A}(\mathbf{S}^k)\|_2}{\sqrt{n}} - \frac{\|\mathbf{S}^k\|_F}{\sqrt{d_1d_2}}\right| \geq \frac{\delta}{\sqrt{d_1d_2}} + \frac{32\|\mathbf{S}\|_{\infty,\infty}}{\sqrt{n}}\right] \leq C\exp(-C_2n\delta^2),$$

which complete the proof. $\square$

## G.2 Proof of Lemma F.2

*Proof.* According to Lemma 5 in Negahban and Wainwright (2012), we have following results

$$\mathbb{P}\left[\left|\sup_{\boldsymbol{\Delta}\in\mathcal{D}(\delta)}\sqrt{d_1 d_2}\frac{\|\mathcal{A}(\boldsymbol{\Delta})\|_2}{\sqrt{n}} - \mathbb{E}\left[\sup_{\boldsymbol{\Delta}\in\mathcal{D}(\delta)}\sqrt{d_1 d_2}\frac{\|\mathcal{A}(\boldsymbol{\Delta})\|_2}{\sqrt{n}}\right]\right| \geq \delta\right] \leq C\exp(-C_1 n\delta^2), \quad (G.3)$$

and

$$\left(\mathbb{E}\left[\sup_{\boldsymbol{\Delta}\in\mathcal{D}(\delta)}\sqrt{d_1 d_2}\frac{\|\mathcal{A}(\boldsymbol{\Delta})\|_2}{\sqrt{n}}\right]\right)^2 \leq 16\sqrt{d_1 d_2}\|\boldsymbol{\Delta}\|_\infty \mathbb{E}\left[\sup_{\boldsymbol{\Delta}\in\mathcal{D}(\delta)}\frac{1}{n}\sum_{j,k\in\Omega}\xi_{jk}\langle\mathbf{A}_{jk},\boldsymbol{\Delta}\rangle\right] + \delta^2,$$

where $\xi_{jk}$ are independent Rademacher variables. Furthermore, by the duality between norms, we can obtain

$$\frac{1}{n}\sum_{j,k\in\Omega}\xi_{jk}\langle\mathbf{A}_{jk},\boldsymbol{\Delta}\rangle \leq \left\|\frac{1}{n}\sum_{j,k\in\Omega}\xi_{jk}\mathbf{A}_{jk}\right\|_\infty \cdot \|\boldsymbol{\Delta}\|_{1,1} \leq \rho(D)\left\|\frac{1}{n}\sum_{j,k\in\Omega}\xi_{jk}\mathbf{A}_{jk}\right\|_{\infty,\infty},$$

where the last inequality is due to the fact that $\boldsymbol{\Delta}\in\mathcal{D}(\delta)$. Finally, we have

$$\left\|\frac{1}{n}\sum_{j,k\in\Omega}\xi_{jk}\mathbf{A}_{jk}\right\|_{\infty,\infty} \leq C\sqrt{\frac{\log d}{n}}. \quad (G.4)$$

To prove this, we use Hoeffding's inequality and Union Bound. By the definition of $\mathbf{A}_i$, we can obtain

$$\left\|\frac{1}{n}\sum_{i=1}^n \xi_i \mathbf{A}_i\right\|_{\infty,\infty} = \max_{j,k}\left|\frac{1}{n}\sum_{i=1}^n \xi_i \mathbf{e}_j^i \mathbf{e}_k^i\right| = \max_{j,k}\left|\frac{1}{n}\sum_{i=1}^n Z_{jk}^i\right|,$$

where we have $Z_{jk}^i = \xi_i A_{jk}$. Thus we can get $|Z_{jk}^i| \leq |\xi_i| = 1$, and we conclude that $Z_{jk}^i$ are independent centered sub-Gaussian random variables for $i = 1,\ldots,n$. Therefore, following the same procedure as in the proof of Lemma D.4, we can obtain inequality (G.4). Therefore, we can obtain

$$\left(\mathbb{E}\left[\sup_{\boldsymbol{\Delta}\in\mathcal{D}(\delta)}\frac{\|\mathcal{A}(\boldsymbol{\Delta})\|_2}{\sqrt{n}}\right]\right)^2 \leq C\frac{\|\boldsymbol{\Delta}\|_{\infty,\infty}}{\sqrt{d_1 d_2}}\rho(D)\sqrt{\frac{\log d}{n}} + \frac{\delta^2}{d_1 d_2} \leq C'\frac{D^2}{d_1 d_2},$$

where the last inequality comes from the definition of $\rho(D)$, $\delta$ and $\|\boldsymbol{\Delta}\|_{\infty,\infty} \leq 2/\sqrt{d_1 d_2}$. It implies that

$$\mathbb{E}\left[\sup_{\boldsymbol{\Delta}\in\mathcal{D}(\delta)}\frac{\|\mathcal{A}(\boldsymbol{\Delta})\|_2}{\sqrt{n}}\right] \leq C''\frac{D}{\sqrt{d_1 d_2}}. \quad (G.5)$$

Thus combining (G.3) and (G.5), we have

$$\sup_{\boldsymbol{\Delta}\in\mathcal{D}(\delta)}\frac{\|\mathcal{A}(\boldsymbol{\Delta})\|_2}{\sqrt{n}} \leq \frac{D}{2\sqrt{d_1 d_2}}$$

holds with probability at least $1 - C\exp(-C_1 nD^2)$. $\square$

# H Other Auxiliary Lemmas

**Lemma H.1.** (Ledoux, 2005) Consider independent random variables $Y_1, \ldots, Y_n$ such that $a_i \leq Y_i \leq b_i$ for $i = 1, \ldots, n$. Let

$$Z := \sup_{\mathbf{t} \in \mathcal{T}} \sum_{i=1}^{n} t_i Y_i,$$

where $\mathcal{T}$ is a family of vectors $\mathbf{t} \in \mathbb{R}^n$ such that $\sigma = \sup_{t \in \mathcal{T}} \left( \sum_{i=1}^{n} t_i^2 (b_i - a_i)^2 \right)^{1/2} \leq \infty$. Then, for any $r \geq 0$, we have

$$\mathbb{P}\big(|Z - m_Z| \geq r\big) \leq 4 \exp\Big( -\frac{r^2}{4\sigma^2} \Big),$$

where $m_Z$ is a median of $Z$. Furthermore, we have

$$|\mathbb{E}(Z) - m_Z| \leq 4\sqrt{\pi}\sigma \quad \text{and} \quad \text{Var}(Z) \leq 16\sigma^2.$$